\documentclass{article}

\usepackage[nonatbib,preprint]{neurips_2022}

\usepackage[utf8]{inputenc} 
\usepackage[T1]{fontenc}    
\usepackage{hyperref}       
\usepackage{url}            
\usepackage{booktabs}       
\usepackage{amsfonts}       
\usepackage{nicefrac}       
\usepackage{microtype}      
\usepackage{xcolor}         
\usepackage{tabto}    
\usepackage{setspace}  
\usepackage{titlesec}

\usepackage{subfig}
\usepackage{wrapfig}
\usepackage{tikz}
\usepackage{tikz-cd}
\usetikzlibrary{cd}
\usepackage{microtype}
\usepackage{graphicx}
\usepackage{booktabs} 
\usepackage{pifont}       
\usepackage{bbding}       %
\usepackage{fontawesome}  
\usepackage{jabbrv}
\usepackage{mhchem} 

\usepackage{amsmath}
\usepackage{amssymb}
\usepackage{mathtools}
\usepackage{amsthm}
\usepackage{arydshln}
\usepackage[capitalize,noabbrev]{cleveref}

\usepackage{mathtools} 

\usepackage{amsmath,amsfonts,bm}

\def\1{\bm{1}}

\DeclareMathAlphabet{\mathsfit}{\encodingdefault}{\sfdefault}{m}{sl}
\SetMathAlphabet{\mathsfit}{bold}{\encodingdefault}{\sfdefault}{bx}{n}

\usepackage{listings}

\definecolor{codegreen}{rgb}{0,0.6,0}
\definecolor{codegray}{rgb}{0.5,0.5,0.5}
\definecolor{codepurple}{rgb}{0.58,0,0.82}
\definecolor{backcolour}{rgb}{0.95,0.95,0.92}

\makeatletter
\renewcommand{\fnum@figure}{\textbf{Fig. \thefigure} | }
\renewcommand{\fnum@table}{\textbf{Table \thetable \ |}}

\usepackage{hyperref}
\usepackage{url}
\usepackage{minted} 
\usepackage{adjustbox} 
\setminted[python]{breaklines}
\usepackage{multicol}

\makeatother
\usepackage[labelsep=space]{caption}
\usepackage[
singlelinecheck=false 
]{caption}

\usepackage{authblk}

\usepackage[
  backend=biber,
  style=nature,
  sorting=none,
  defernumbers=true,
  doi=true,
  url=false,
  isbn=false
]{biblatex}

\DeclareFieldFormat{doi}{%
  \mkbibacro{DOI}\addcolon\space
  \href{https://doi.org/#1}{#1}%
}

\DeclareBibliographyCategory{maincited}
\newif\ifmainsegment
\mainsegmentfalse

\AtEveryCitekey{%
  \ifmainsegment
    \addtocategory{maincited}{\thefield{entrykey}}%
  \fi
}

\title{
Evaluating Large Language Models in Scientific Discovery
}

\author[1, $\dag$, $\ddag$]{Zhangde Song}
\author[1, $\dag$]{Jieyu Lu}
\author[2, $\dag$]{Yuanqi Du}
\author[3, $\dag$]{Botao Yu}
\author[4, $\dag$]{Thomas M. Pruyn}
\author[5, $\dag$]{Yue Huang}
\author[5, $\dag$]{Kehan Guo}
\author[6, $\dag$]{Xiuzhe Luo}
\author[7, $\dag$]{Yuanhao Qu}
\author[8, $\ddag$]{Yi Qu}
\author[9, $\ddag$]{Yinkai Wang}
\author[10, $\ddag$]{Haorui Wang}
\author[11, $\ddag$]{Jeff Guo}
\author[12, $\ddag$]{Jingru Gan}
\author[13, $\ddag$]{Parshin Shojaee}
\author[14, 15, $\ddag$]{Di Luo}
\author[7, $\ddag$]{Steven Dillmann}
\author[11]{Andres M Bran}
\author[16]{Gen Li}
\author[1]{Qiyuan Zhao}
\author[17]{Shao-Xiong Lennon Luo}
\author[18, 19, 20]{Yuxuan Zhang}
\author[19]{Xiang Zou}
\author[21]{Wanru Zhao}
\author[22]{Yifan F. Zhang}
\author[23]{Wucheng Zhang}
\author[24]{Shunan Zheng}
\author[25]{Saiyang Zhang}
\author[4]{Sartaaj Takrim Khan}
\author[4]{Mahyar Rajabi-Kochi}
\author[26]{Samantha Paradi-Maropakis}
\author[27]{Tony Baltoiu}
\author[28]{Fengyu Xie}
\author[29]{Tianyang Chen}
\author[7]{Kexin Huang}
\author[30, 31]{Weiliang Luo}
\author[32]{Meijing Fang}
\author[30]{Xin Yang}
\author[33]{Lixue Cheng}
\author[34]{Jiajun He}
\author[9]{Soha Hassoun}
\author[5]{Xiangliang Zhang}
\author[12]{Wei Wang}
\author[13]{Chandan K. Reddy}
\author[10]{Chao Zhang}
\author[35]{Zhiling Zheng}
\author[22]{Mengdi Wang}
\author[7]{Le Cong}
\author[2]{Carla P. Gomes}
\author[32]{Chang-Yu Hsieh}
\author[36]{Aditya Nandy}
\author[11]{Philippe Schwaller}
\author[30, 31]{Heather J. Kulik}
\author[1, *]{Haojun Jia}
\author[3, *]{Huan Sun}
\author[4, 18, *]{Seyed Mohamad Moosavi}
\author[1, $\ddag$, *]{Chenru Duan}

\affil[1]{Deep Principle, Hangzhou, China}
\affil[2]{Department of Computer Science, Cornell University, Ithaca, NY, USA}
\affil[3]{Department of Computer Science and Engineering, The Ohio State University, Columbus, OH, USA}
\affil[4]{Department of Chemical Engineering \& Applied Chemistry, University of Toronto, Toronto, ON, Canada}
\affil[5]{Department of Computer Science and Engineering, University of Notre Dame, Notre Dame, IN, USA}
\affil[6]{QuEra Computing Inc., Boston, MA, USA}
\affil[7]{Department of Pathology, Department of Genetics, Cancer Biology Program, Stanford University School of Medicine, Stanford, CA, USA}
\affil[8]{Harvard Law School, Cambridge, MA, USA}
\affil[9]{Department of Computer Science, Tufts University, Medford, MA, USA}
\affil[10]{School of Computational Science and Engineering, Georgia Institute of Technology, Atlanta, GA, USA}
\affil[11]{Laboratory of Artificial Chemical Intelligence, Ecole Polytechnique Federale de Lausanne, Lausanne, Switzerland}
\affil[12]{Department of Computer Science, University of California, Los Angeles, Los Angeles, CA, USA}
\affil[13]{Department of Computer Science, Virginia Tech, Arlington, VA, USA}
\affil[14]{Department of Physics, Tsinghua University, Beijing, China}
\affil[15]{Institute for Advanced Study, Tsinghua University, Beijing, China}
\affil[16]{Department of Chemistry, Princeton University, Princeton, NJ, USA}
\affil[17]{School of Engineering and Applied Sciences, Harvard University, Cambridge, MA, USA}
\affil[18]{Vector Institute for Artificial Intelligence, Toronto, ON, Canada}
\affil[19]{Department of Physics, University of Toronto, Toronto, ON, Canada}
\affil[20]{Institute of Physics, Ecole Polytechnique Federale de Lausanne, Lausanne, Switzerland}
\affil[21]{Department of Computer Science and Technology, University of Cambridge, Cambridge, United Kingdom}
\affil[22]{Department of Electrical and Computer Engineering, Princeton University, Princeton, NJ, USA}
\affil[23]{Department of Physics, Princeton University, Princeton, NJ, USA}
\affil[24]{Department of Mechanical Engineering, The University of Texas at Austin, Austin, TX, USA}
\affil[25]{Department of Physics, The University of Texas at Austin, Austin, TX, USA}
\affil[26]{Department of Biomedical Engineering, University of Toronto, Toronto, ON, Canada}
\affil[27]{Department of Mechanical Engineering, McGill University, Montreal, QC, Canada}
\affil[28]{College of Artificial Intelligence and Data Science, Suzhou Institute of Advanced Research, University of Science and Technology of China, Suzhou, Jiangsu, China}
\affil[29]{School of Science and Engineering, The Chinese University of Hong Kong (Shenzhen), Shenzhen, Guangdong, China}
\affil[30]{Department of Chemistry, Massachusetts Institute of Technology, Cambridge, MA, USA}
\affil[31]{Department of Chemical Engineering, Massachusetts Institute of Technology, Cambridge, MA, USA}
\affil[32]{College of Pharmaceutical Sciences, Zhejiang University, Hangzhou, Zhejiang, China}
\affil[33]{Department of Chemistry, The Hong Kong University of Science and Technology, Clear Water Bay, Kowloon, Hong Kong SAR, China}
\affil[34]{Department of Engineering, University of Cambridge, Cambridge, United Kingdom}
\affil[35]{Department of Chemistry, Washington University in St. Louis, St. Louis, MO, USA}
\affil[36]{Department of Chemical and Biomolecular Engineering, University of California, Los Angeles, CA, USA}

\affil[$\dag$]{These authors contribute equally}
\affil[$\ddag$]{Project contributor}
\affil[*]{Correspondence to: haojunjia@deepprinciple.com, sun.397@osu.edu, mohamad.moosavi@utoronto.ca, duanchenru@gmail.com}

\begin{document}

\maketitle
\begin{refsection}
\mainsegmenttrue
\begin{refsegment}

\begin{abstract}
Large language models (LLMs) are increasingly applied to scientific research, yet prevailing science benchmarks probe decontextualized knowledge and overlook the iterative reasoning, hypothesis generation, and observation interpretation that drive scientific discovery.
Here, we introduce a scenario-grounded benchmark that evaluates LLMs across biology, chemistry, material science, and physics. For this benchmark, domain experts define research projects of genuine value and interest, and decompose them into modular research scenarios from which vetted questions are sampled. 
The framework assesses models at two levels: (i) question-level accuracy on scenario-tied items and (ii) project-level performance, where models must propose testable hypotheses, execute simulations or experiments, and interpret results.
Applying this two-phase scientific discovery evaluation (\texttt{SDE}) framework to state-of-the-art LLMs reveals a consistent performance gap relative to general science benchmarks, diminishing return of scaling up model sizes and reasoning, and systematic weaknesses shared across top-tier models from different providers.
Large performance variation in research scenarios leads to changing choices of the best performing model on scientific discovery projects evaluated, suggesting all current LLMs are distant to general scientific ``superintelligence''.
Nevertheless, LLMs already demonstrate promise in a great variety of scientific discovery projects, including cases where constituent scenario scores are low, highlighting the role of guided exploration and serendipity in discovery.
This \texttt{SDE} framework offers a reproducible benchmark for discovery-relevant evaluation of LLMs and charts practical paths to advance their development toward scientific innovation.
\end{abstract}
\setstretch{1.8}

\section*{Introduction}
Large language models (LLMs) are beginning to accelerate core stages of scientific discovery, from literature triage and hypothesis generation to computational simulation, code synthesis, and even autonomous experimentation\cite{vaswani2017attention,brown2020language,skarlinski2024paperqa2,kaplan2020scaling,yao2022react,Rapp2024,Cooper2024,wang2023scientific,scienceagentbench,autosdt}. 
Starting as surrogates for structure-property prediction and simple question-answering\cite{jablonka2024leveraging,zheng2025llm4sd,gelman2025metl,hayes2025esm3,pruyn2025mof}, LLMs, especially with recent reasoning capability emerged from reinforcement learning and test-time compute, further extend their roles in scientific discovery by having the potential to provide intuitions and insights\cite{wei2022chain,wang2023selfconsistency,o1systemcard2024,guo2025deepseek,hayes2025esm3,textgrad}.
Illustrative successes include chemistry tool-use agents\cite{bran2024chemcrow,yu2024chemtoolagent}, autonomous ``co‑scientists'' \cite{boiko2023coscientist,google2025aicoscientist,aiscientist_v2}, and the Virtual Lab for nanobody design\cite{swanson2025virtuallab} that have begun to plan, execute, and interpret experiments by coupling language reasoning to domain tools, laboratory automation, and even embodied systems (e.g., LabOS\cite{cong2025labos}).
Together, these examples suggest that LLMs can already assist scientists in a ``human-in-the-loop'' scientific discovery\cite{du2024generative,lu2024aiscientist,tom2024selfdriving,xin2025towards,agentsurvey,crisprgpt,scitoolagent,tooluniverse,chatMOF,reddy2025towards,kosmos,huang2025biomni,qiu2025_physicssupernova,Mandal2025}. 

\begin{figure*}[t!]
    \centering
    \includegraphics[width=0.95\textwidth]{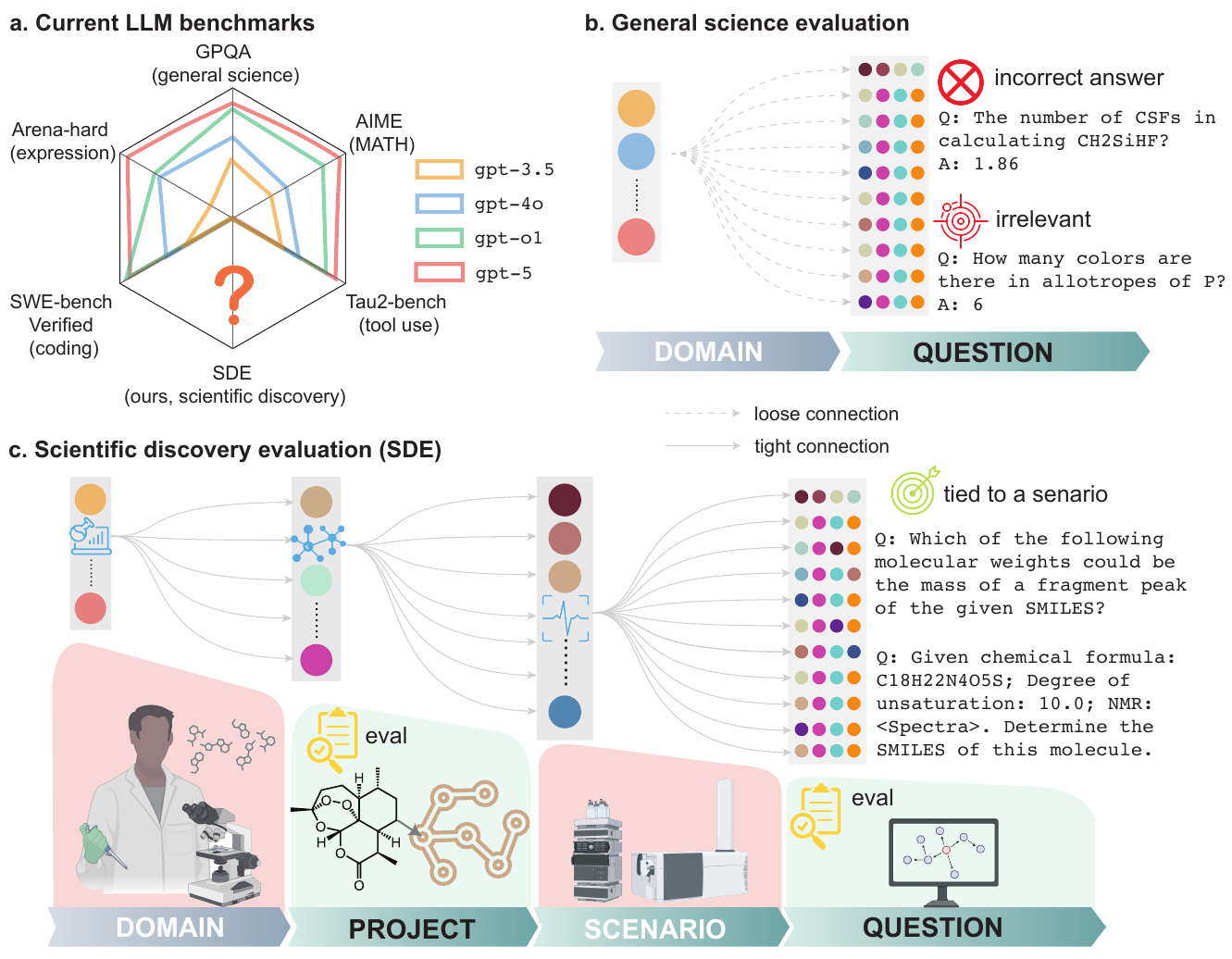}
    \caption{
    \textbf{From evaluating LLMs on general-science quizzes to scenario-grounded scientific discovery.}
    \textbf{a.} Schematic comparison of representative LLM benchmarks. GPQA, AIME, Arena-hard, SWE-bench verified, Tau2-bench, alongside our scientific discovery evaluation (\texttt{SDE}) are shown. Shaded polygons indicate relative performance of four models (\texttt{gpt-3.5}, \texttt{gpt-4o}, \texttt{gpt-o1}, \texttt{gpt-5}) across benchmarks. Only GPT series are shown as representatives to show their performance improve with time.
    \textbf{b.} Limitations of general-science Q$\&$A. Existing benchmarks often contain questions that are less relevant to scientific discovery or incorrect answers as ground-truth.
    \textbf{c.} The \texttt{SDE} framework anchors assessment to projects and realistic research scenarios within each scientific domain, producing tightly coupled questions, enabling more faithful evaluation of LLMs for scientific discovery. LLMs are evaluated on both question and project levels. A project of discovering new pathways for artemisinin synthesis is shown as an example, which comprises multiple scenarios, such as forward reaction prediction and structure elucidation from nuclear magnetic resonance (NMR) spectra, where the question sets are finally collected.
}
    \label{fig:overview}
\end{figure*}

\qquad In contrast, evaluation has lagged behind this end‑to‑end reality in scientific discovery\cite{SciArena2025,scienceagentbench,DiscoveryBench}.
Benchmarks in coding (e.g., SWE‑bench verified\cite{swebench_verified2024}), mathematics (e.g., AIME\cite{aime2025}), writing and expression (e.g., Arena-hard\cite{li2025arena_hard}), and tool use (e.g., Tau2-bench\cite{yao2024tau_bench}) have matured into comparatively stable tests with clear ground truth and strong predictive validity for capability gains (Fig. \ref{fig:overview}a). 
Widely used science benchmarks (e.g., GPQA\cite{rein2023gpqa}, ScienceQA\cite{lu2022scienceqa}, MMMU\cite{yue2023mmmu}, SuperGPQA\cite{SuperGPQA}, Humanity's Last Exam\cite{hle2025}), however, remain largely decontextualized, perception‑heavy question and answering (Q$\&$A), with items loosely connected to specific research domains and susceptible to label noise (Fig. \ref{fig:overview}b).
\emph{Mastery of static, decontextualized questions, even if perfect, does not guarantee readiness to discovery, just as earning straight A's in coursework does not indicate a great researcher}\cite{zhang2025exploring,mirza2025chembench,yin2025genomebench}.
As LLMs become more deeply integrated into scientific research and discovery workflows, proper evaluation must measure a model’s abilities of understanding the specific context of research, reasoning under imperfect evidence, and iteratively refining hypotheses, not just answering isolated questions\cite{alampara2025macbench}.

\qquad We introduce a systematic evaluation of LLMs grounded in real-world research scenarios for scientific discovery (named Scientific Discovery Evaluation, or \texttt{SDE}, Fig. \ref{fig:overview}c). 
Across four domains (biology, chemistry, materials, and physics), we start with concrete research \textbf{projects} of genuine interest to domain experts and decompose each into modular research \textbf{scenarios}, which are scientifically grounded and reusable across multiple applications. 
Within each scenario, we construct expert-vetted \textbf{questions}, formatted in line with conventional LLM benchmarks (multiple choice or exact match), such that their evaluation constitutes measurable progress toward in-context scientific discovery.
This tight connection among \textbf{questions, scenarios, and projects} built in \texttt{SDE} reveals the true capability of LLMs in scientific discovery.
Beyond per-question evaluation as in conventional science benchmarks, we also evaluate LLMs’ performance at the level of open-ended scientific discovery projects. 
In this setting, LLMs are put into the loop of scientific discovery, where they are required to autonomously propose testable hypotheses, obtain feedback from simulations or experiments, and interpret the results to refine their original hypotheses, imitating an end-to-end scientific discovery process.
There, discovery-oriented outcomes (e.g., polarizability of proposed transition metal complexes) are finally evaluated. 
This project-level evaluation reveals capability gaps and failure modes across different research workflows. 
Applying this multi-level evaluation framework to state-of-the-art LLMs released over time yields a longitudinal, fine-grained benchmark that reveals where current models succeed, where they fail, and why. 
The resulting analysis suggests actionable avenues, spanning targeted training on problem formulation, diversifying data sources, baking in computational tool use in training, and designing reinforcement learning strategies in scientific reasoning, for steering LLM development toward scientific discovery.

\section*{Results}
\subsection*{Question-level evaluations}
\begin{figure*}[!h]
    \centering
    \includegraphics[width=0.98\textwidth]{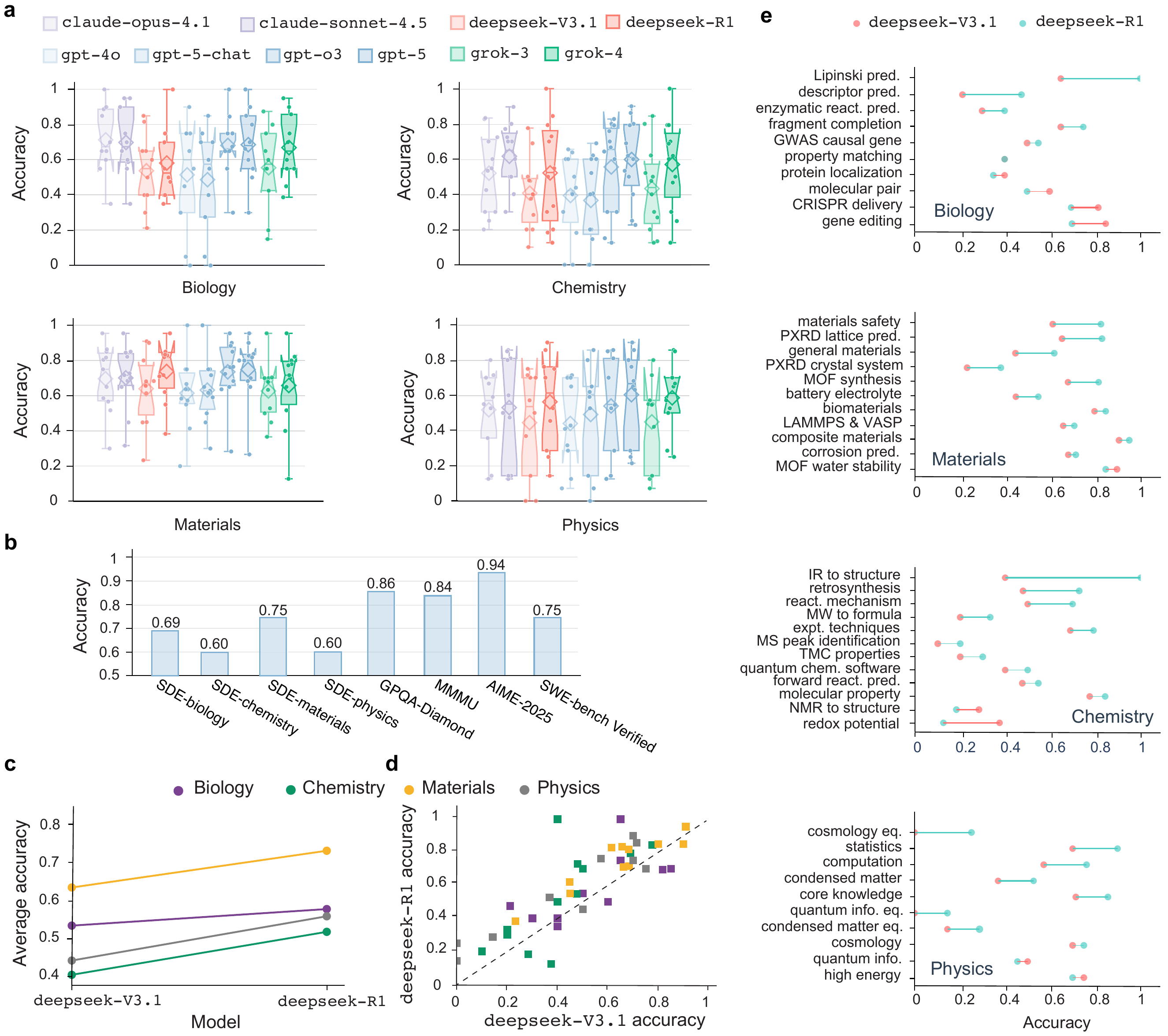}
    \caption{
    \textbf{Comparative performance of frontier language models across scientific domains.}
    \textbf{a.} Distribution of per-domain accuracies for ten models on biology, chemistry, materials and physics. Box plots summaries aggregate scenario-level performance, where each scenario is represented as a dot. Mean and median accuracy are shown by diamond and solid line, respectively. The models are colored as the following: light purple for \texttt{claude-opus-4.1} and \texttt{claude-sonnet-4.5}, coral red for \texttt{deepseek-V3.1} and \texttt{deepseek-R1}, light blue for \texttt{gpt-4o}, \texttt{gpt-5-chat}, \texttt{gpt-o3}, and \texttt{gpt-5}, teal green for \texttt{grok-3} and \texttt{grok-4}), with higher opacity for more recent release. 
    \textbf{b.} Mean accuracy of \texttt{gpt-5} on four domains of questions in \texttt{SDE} in comparison to select conventional benchmarks  (GPQA-Diamond, MMMU, AIME-2025, SWE-bench Verified).
    \textbf{c.} Domain-averaged accuracy for \texttt{deepseek-V3.1} and \texttt{deepseek-R1} with biology in purple, chemistry in green, materials in orange, and physics in gray.
    \textbf{d.} Scenario-wise comparison of \texttt{deepseek-R1} (y-axis) versus \texttt{deepseek-V3.1} (x-axis). The dashed diagonal line denotes parity, with points above the line indicating scenarios where \texttt{deepseek-R1} outperforms \texttt{deepseek-V3.1}.
    \textbf{e.} Accuracies for \texttt{deepseek-V3.1} (red) and \texttt{deepseek-R1} (indego) categorized by domains and scenarios. The horizontal line is colored as indigo when \texttt{deepseek-R1} outperforms \texttt{deepseek-V3.1}, otherwise as red.
}
    \label{fig:performance}
\end{figure*}

\paragraph{Performance gap in quiz- and discovery-type questions.}
To go beyond the conventional science Q$\&$A benchmark where questions are sometimes assembled opportunistically, questions in \texttt{SDE} are collected in a completely different routine (Fig. \ref{fig:overview}c).
In each domain, a multi-member expert panel defined roughly ten common research scenarios where LLMs could plausibly help their ongoing projects. 
These scenarios span a broad spectrum, from those human experts are proficient (e.g., making decisions from specific experimental observations) to those effectively intractable to human experts without the assistance of tools (e.g., inferring oxidation and spin states solely from a transition metal complex structure). 
When feasible, questions were generated semi-automatically by sampling and templating from open datasets\cite{mirza2025chembench}, with NMR spectra to molecular structure mapping as an example. 
Otherwise, especially for experiment-related scenarios, questions were drafted manually by an expert. Every question underwent panel review, with inclusion contingent on consensus about the validity and correctness, resulting in 1,125 questions in the \texttt{SDE} benchmark (see Methods section, \textit{\nameref{question_collection}}).
This design ties every question to a research scenario, ensuring that its correctness reflects progress on a practical scientific discovery project rather than decontextualized trivia, which also allows comparisons across LLMs at the same level of granularity.
With the goal of understanding how the performance of popular coding, math, and expression benchmarks translates to scientific discovery, top-tier models from various providers (i.e., OpenAI, Anthropic, Grok, and DeepSeek) are evaluated through an adapted version of \texttt{lm-evaluation-harness} framework\cite{forkedLMHarness}, which supports flexible evaluation through API on various task types\cite{lm_eval_harness} (see Methods section, \textit{\nameref{eval}}).
Among all LLMs, only \texttt{deepseek-V3.1} and \texttt{deepseek-R1} are fully open-weight\cite{guo2025deepseek}.

\qquad Scores at each scenario, defined as percentages of questions that a model answered correctly, are aggregated per domain for all models evaluated (Fig. \ref{fig:performance}a).
The performance varies drastically across different models, while in all domains with the latest flagship LLM from a commercial provider ranks the highest (Supplementary Fig. \ref{Supp:model_average_performance}).
To situate these results, we compare model performance on our discovery-grounded questions with widely used general-science Q$\&$A benchmarks. 
On our \texttt{SDE} benchmark, state-of-the-art models reach a score of 0.71 in biology (\texttt{claude-4.1-opus}), 0.60 in chemistry (\texttt{claude-4.5-sonnet}), 0.75 in materials (\texttt{gpt-5}), and 0.60 in physics (\texttt{gpt-5}). 
By contrast, the same class of models attains 0.84 on MMMU-Pro and 0.86 on GPQA-Diamond (\texttt{gpt-5}), illustrating a consistent gap between decontextualized Q$\&$A and scenario-grounded scientific discovery questions (Fig. \ref{fig:performance}b).
In spite of the corpus-language effect that recent scientific literature is predominantly written in English, we find that \texttt{deepseek-R1}, as the representative of the strongest open-weight models, starts to approach the performance of top-tier closed-source LLMs, narrowing gaps that were pronounced only a few releases ago. 
This observation underscores the pace of community catching up on iterative improvement of training data, methodology, and infrastructure, thanks to the efforts in open source\cite{guo2025deepseek,gpt_oss}.

\qquad The performance of a model varies significantly across research scenarios (Fig. \ref{fig:performance}a, Supplementary Fig. \ref{Supp:task_agreement_analysis}).
For example, \texttt{gpt-5} achieves impressive performance in retrosynthesis planning (score of 0.85) while struggling with NMR structure elucidation (score of 0.23). 
This observation, as exemplified by the wide spectrum of accuracy in each domain, holds for all LLMs evaluated, reinforcing the fact that conventional science benchmarks that only categorize questions into domains or subdomains are insufficient to detail the fields of mastery and improvement for LLMs.
This finer-grained assessment is important, as scientific discovery is often blocked by misinformation and incorrect decisions rooted in the weakest scenario.
With the \texttt{SDE} benchmark, we establish a look-up table that assesses LLMs' capability in specific research scenarios when people consider applying LLMs in their research workflows.

\paragraph{Reasoning and scaling plateau.}
On established coding and mathematics benchmarks, state-of-the-art performance typically progresses with model releases. 
Reasoning is a major driver of those gains, which matters no less in scientific discovery.\cite{yue2025RLReasoning,karan2025reasoningsamplingbasemodel}
In the head-to-head comparisons of otherwise comparable models, variants with explicit test-time reasoning consistently outperform their non-reasoning counterparts on the \texttt{SDE} problems, best exemplified by the enhanced performance of \texttt{deepseek-R1} compared to \texttt{deepseek-V3.1}, both sharing the same base model\cite{guo2025deepseek} (Fig. \ref{fig:performance}c).
The effect holds across biology, chemistry, materials, and physics and across most of the scenarios, indicating that improvements in reasoning corresponding to multi-step derivation and evidence integration translate directly into higher accuracy in discovery-oriented settings (Fig. \ref{fig:performance}d).
One salient example is to let LLMs judge whether an organic molecule satisfies Lipinski's rule of five, a famous guideline for predicting the oral bioavailability of a drug candidate, where reasoning is expected to be vital (Fig. \ref{fig:performance}e).
There, the accuracy boosts from 0.65 to 1.00 by turning on reasoning capability in DeepSeek models.

\begin{figure*}[!h]
    \centering
    \includegraphics[width=0.98\textwidth]{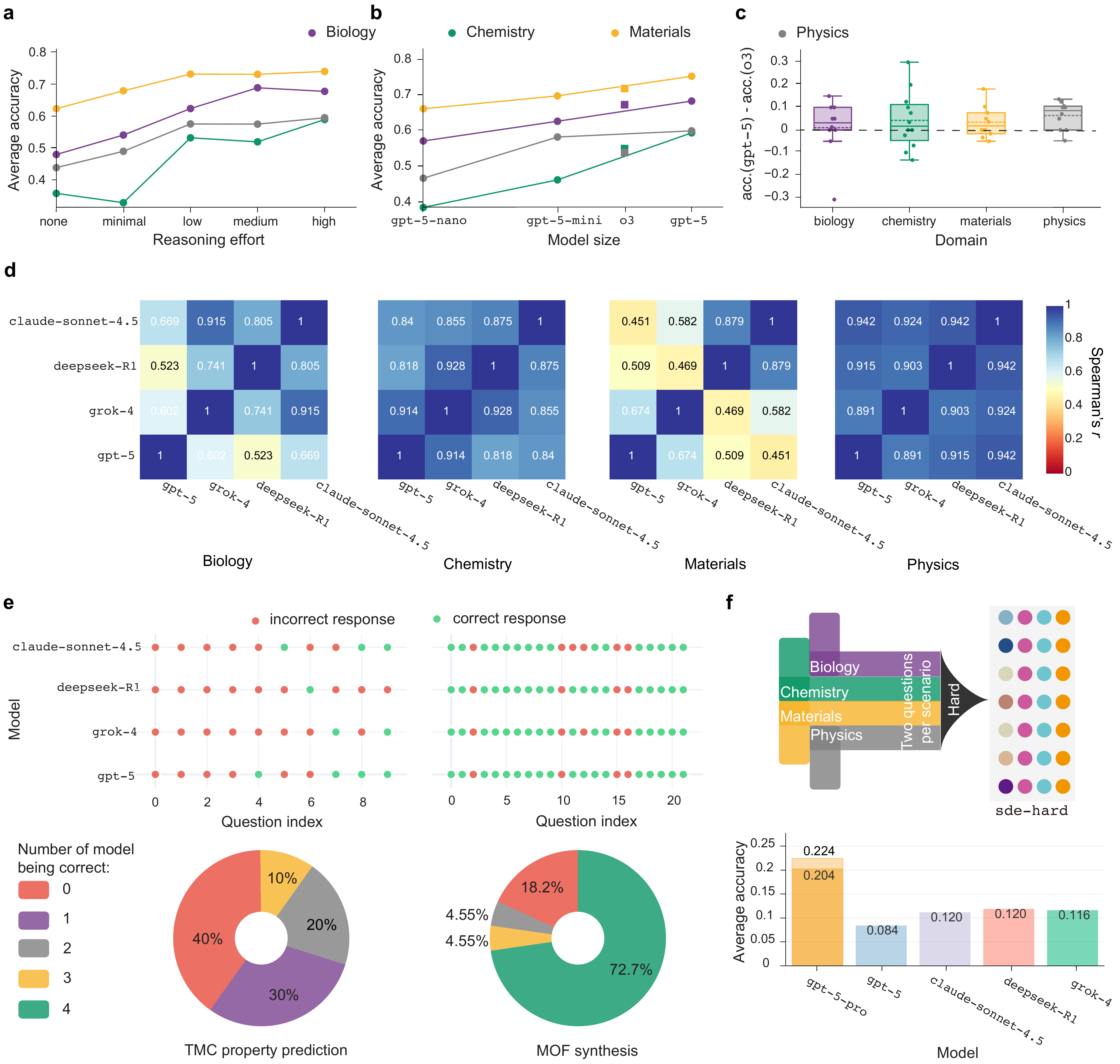}
    \caption{
    \textbf{Scaling, reasoning, and cross-model patterns on scientific discovery questions.}
    \textbf{a.} Average accuracy as a function of reasoning effort (from none to high) across four domains for \texttt{gpt-5} model series. Biology is colored in purple, chemistry in green, materials in orange, and physics in gray.
    \textbf{b.} Average accuracy versus model size (\texttt{gpt-5-nano}, \texttt{gpt-5-mini},  \texttt{gpt-5}), showing scaling gains in all four domains. Performance of \texttt{o3} is shown in between of \texttt{gpt-5-mini} and \texttt{gpt-5} as an estimate. All models are evaluated at the reasoning effort of high.
    \textbf{c.} Per-domain distribution of accuracy difference between \texttt{gpt-5} and \texttt{o3}. Box plot summaries variability, with each dot showing a specific scenario and the dashed line marking parity. 
    \textbf{d.} Cross-model rank correlation by domain (Spearman’s \textit{r}) for the top-performing models from each provider, \texttt{gpt-5}, \texttt{grok-4}, \texttt{deepseek-R1}, and \texttt{claude-sonnet-4.5}.
    \textbf{e.} Question-level performance correlation among four models and two scenarios, TMC property predictions (left) and MOF synthesis (right). Each question is marked by its correctness (green dots for correct and red dots for incorrect), together with a doughnut plot for analysis of model consensus (bottom).
    \textbf{f.} Construction of \texttt{sde-hard} (top) and its corresponding model performance (bottom). For \texttt{gpt-5-pro}, there is a consistent timeout behavior on eight questions, resulting ``no response''. The accuracy that considers those questions with ``no response'' as incorrect is shown in solid and as correct in transparent.
}
    \label{fig:scaling}
\end{figure*}

\qquad Yet, despite the clear benefits of reasoning, overall performance starts to saturate on our \texttt{SDE} benchmark when tracked across various reasoning efforts for \texttt{gpt-5}, where the gains become modest and often fall within statistically negligible margins, even when the corresponding models set new records on coding or math (Fig. \ref{fig:scaling}a, Supplementary Fig. \ref{Supp:o3_vs_gpt5_performance_diff} and Fig. \ref{Supp:gpt5_reasoning_distributions}). 
For example, the accuracy barely improves between reasoning efforts of medium and high (0.70 vs. 0.69 in biology, 0.53 vs. 0.60 in chemistry, 0.74 vs 0.75 in materials, and 0.58 vs 0.60 in physics), indicating diminishing returns from the prevailing roadmap of increasing test-time compute for the purpose of scientific discovery (Supplementary Fig. \ref{Supp:medium_vs_high_reasoning_comparison}).
Besides reasoning, scaling up model sizes is considered as a huge contribution in the current success of LLMs (Supplementary Fig. \ref{Supp:reasoning_bench} and Fig. \ref{Supp:size_bench}).
We indeed observe monotonic improvement in model accuracy as \texttt{gpt-5} scales from \texttt{nano} to \texttt{mini} and to its default large size (Fig. \ref{fig:scaling}b). 
However, the scaling effect may also have slowed down during the past year, as indicated by the marginal performance gain of \texttt{gpt-5} over \texttt{o3}, even with 8 scenarios having significantly (i.e., with >0.075 accuracy difference) worse performance (Fig. \ref{fig:scaling}c). 
Similarly, when the factor of reasoning being isolated, the performance improvement from \texttt{gpt-4o} to \texttt{gpt-5-chat} is also negligible, which indicates a seemingly converged behavior in discovery tasks for pretrained base foundation LLMs in the past 18 months.
The implication of reasoning and scaling analysis is not that progress has stalled, but that scientific discovery stresses different competencies than generic scientific Q$\&$A, such as problem formulation, hypothesis refinement, and interpretation of imperfect evidence. 

\paragraph{Shared failure modes among top-performing LLMs.}
When comparing the top performers across different providers (i.e., \texttt{gpt-5}, \texttt{grok-4}, \texttt{deepseek-R1}, and \texttt{claude-sonnet-4.5}), we observe that their accuracy profiles are highly correlated, which tend to rise and fall on the same scenarios (Fig. \ref{fig:scaling}d, Supplementary Fig. \ref{Supp:model_scatter_matrix}). 
This correlation is most prominent in chemistry and physics, where all pairwise Spearman's \textit{r} and Pearson's \textit{r} among the four top-performing models are greater than 0.8 (Supplementary Fig. \ref{Supp:pearson_heatmaps_by_domain}).
In materials, the correlation is the worst, where \texttt{deepseek-R1} and \texttt{claude-sonnet-4.5} have the highest Spearman's \textit{r} of 0.88.
For cases where model correlations are low, we find outstanding contributions from several scenarios of drastically different accuracy by different models, such as PXRD lattice prediction (Supplementary Fig. \ref{Supp:model_scatter_matrix}).
Moreover, top-performing LLMs frequently converge on the same incorrect set of most difficult questions, even when their overall accuracies differ (Fig. \ref{fig:scaling}e, Supplementary Fig. \ref{Supp:ranking_consistency}). 
For example, despite a relatively high accuracy on MOF synthesis questions, the four models make the same mistake on four out of 22 total questions.
This alignment of errors indicates that frontier LLMs mostly share common strengths as well as common systematic weaknesses, plausibly inherited from similar pre-training data and objectives rather than from their distinctive architecture and implementation details\cite{stresstestllms2025}.
Practically, this means that naive ensemble strategies (e.g., majority voting across providers) may deliver limited improvement on scenarios and questions that are inherently difficult to current LLMs (Supplementary Fig. \ref{Supp:task_agreement_analysis} and Fig. \ref{Supp:std_dev_distribution}). 
Our scenario-grounded design makes these correlations visible and reproducible, which not only reveals where models overall succeed, but also in a finer-grained where and why they fail on discovery-oriented tasks, exposing shared failure modes across research pipelines (Supplementary Fig. \ref{Supp:std_vs_mean}).

\qquad Seeing this consensus failing behavior on most difficult questions, we further collected 86 questions, 2 in each research scenario where the top-performing LLMs make most mistakes on, as a subset called \texttt{SDE-hard} (Fig. \ref{fig:scaling}f).
All LLMs score less than 0.12 on these most difficult scientific discovery questions (Supplementary Fig. \ref{Supp:question_performance_scatter} and Fig. \ref{Supp:sde_hard_agreement}).
Surprisingly, \texttt{gpt-5-pro} improves by a significant margin compared to \texttt{gpt-5} and flagship models from other providers.
Despite its impeding (i.e., 12x higher) cost, \texttt{gpt-5-pro} gives correct response on 9 questions where all other models are incorrect (Supplementary Fig. \ref{Supp:sde_hard_pro_correct}).
This observation suggests its competitive advantage on most difficult questions that require extended reasoning, which is characteristic in scientific discovery. 
This accuracy, however, still leaves much room to improve, which makes \texttt{SDE-hard} a great test suite for LLMs with high inference costs that would be released in the future.

\begin{figure*}[!h]
    \centering
    \includegraphics[width=0.98\textwidth]{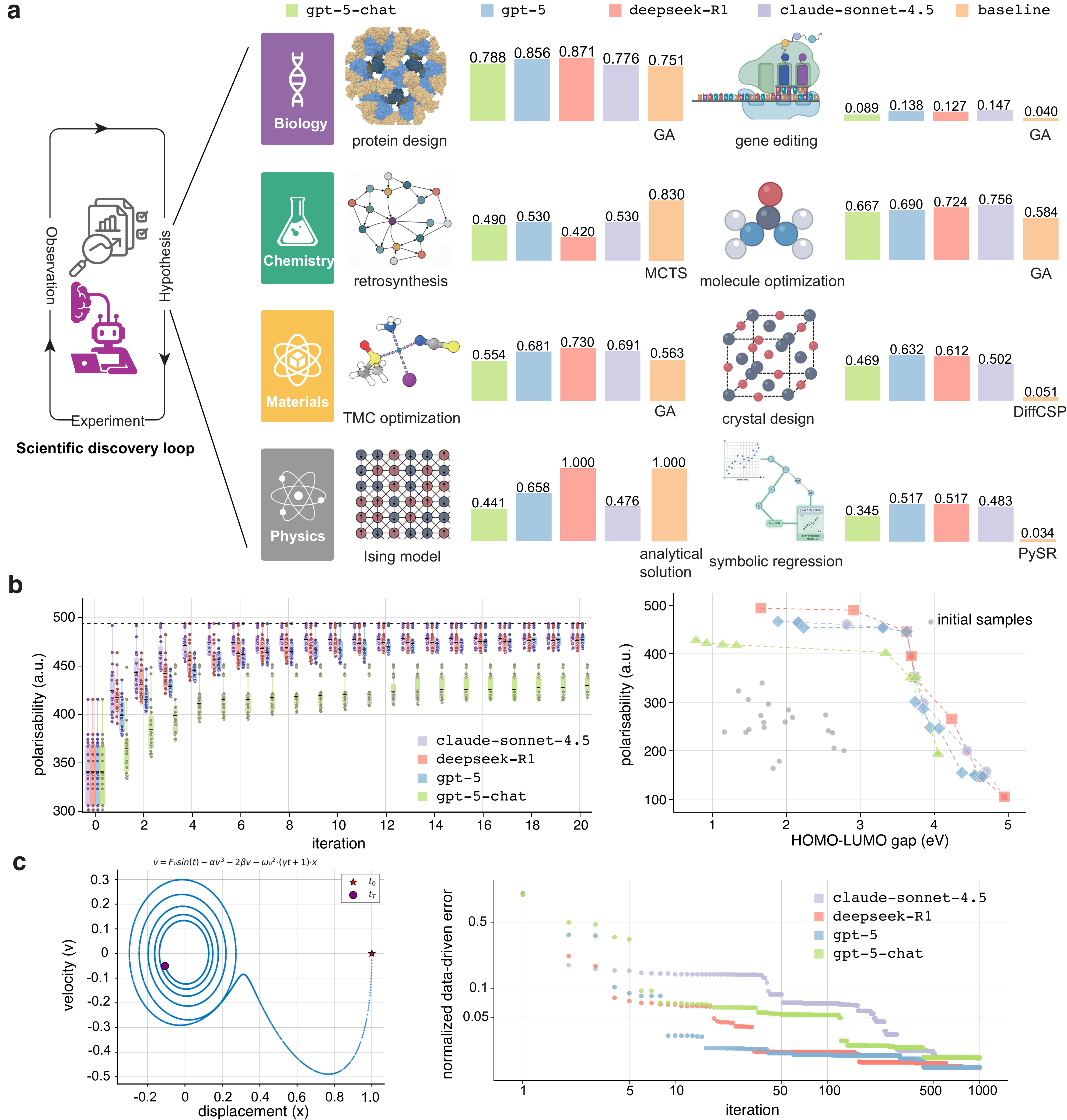}
    \caption{
    \textbf{Evaluating LLMs on scientific discovery projects.}
    \textbf{a.} Schematic for evaluating LLMs as hypothesis generator in the scientific discovery loop and eight projects that span four domains, biology, chemistry, materials, and physics. For each project, a bar plot shows a normalized single-metric performance of four LLMs, \texttt{gpt-5-chat-latest} in light green, \texttt{gpt-5} in light blue, \texttt{deepseek-R1} in coral red, \texttt{claude-sonnet-4.5} in light purple, and baseline in orange. 
    \textbf{b.} Performance of various LLMs on TMC optimization project. (left) Distribution of top-10 TMCs with highest polarisability versus increasing number of iterations, with the theoretical maximum shown by the dashed line for the 1.37M TMC space. (right) Pareto frontier of TMCs for various models after 20 iterations and their initial samples (gray). 
    \textbf{c.} Symbolic regression results on nonlinear dynamical systems. (left) Representative example of phase-space trajectories and (right) discovery curves of the best equation found over iterations, measured by normalized error (lower is better), highlighting differences in convergence behavior and final accuracy across different LLMs.
    Both x and y axis are shown in log scale for visibility.
}
    \label{fig:projects}
\end{figure*}

\subsection*{Project-level evaluations}
\paragraph{Establishing LLM evaluation on the scientific discovery loop.} 
Conventional Q$\&$A benchmarks typically evaluate models via single-turn interactions, scoring isolated responses to static queries.
Scientific discovery, by contrast, advances through iterative cycles of hypothesis proposal, testing, interpretation, and refinement \cite{wang2023scientific}. 
To mirror this process, we introduce \texttt{sde-harness}, a modular framework that formalizes the closed discovery loop of hypothesis, experiment, and observation, wherein the hypothesis is generated by an LLM rather than a human investigator (Fig. \ref{fig:projects}a, see Methods section, \textit{\nameref{project_collection}}). 
Moving beyond per-question accuracy, this framework enables project-level assessment, requiring models to formulate testable hypotheses, execute analyses or simulations, and interpret outcomes to approximate an end-to-end discovery workflow. 
Consequently, \texttt{sde-harness} isolates capabilities that static Q$\&$A tests fail to capture, such as maintaining state across multiple assessment rounds, integrating intermediate evidence, and strategically deciding when to branch or abandon a line of inquiry.
We instantiated eight projects spanning biology, chemistry, materials, and physics, each aligned with a set of specific research scenarios in the \texttt{SDE} Q$\&$A benchmark (Supplementary Table \ref{Supp:correspondence}). 
Each project defines: (i) a hypothesis space (e.g., retrosynthetic routes, metal–ligand complexes with target electronic properties, or symbolic expressions of mathematical relations); (ii) computational oracles or simulators that map hypotheses to observations; and (iii) a selection rule that propagates promising hypotheses across iterations.
Concretely, \texttt{sde-harness} orchestrates iterative optimization to emulate the authentic cycle of scientific discovery. 
This transparent update mechanism reveals how LLMs refine their hypotheses over time, distinguishing iterative reasoning from mere one-shot response generation.

\paragraph{LLMs are  driving scientific discovery.} 
Projects characterized by abundant, well-structured open-source data and codified knowledge, such as protein design, transition metal complex (TMC) optimization, organic molecule optimization, crystal design, and symbolic regression, exhibit the most significant gains from LLM integration (Fig. \ref{fig:projects}a and Supplementary Text \ref{Supp:projects_section}). In symbolic regression, for example, we evaluate LLMs on their ability to iteratively discover governing equations of nonlinear dynamical systems from data, a setting that requires both structured exploration of the hypothesis space and progressive refinement of symbolic forms.
Across different LLMs, reasoning models exhibit more effective discovery dynamics (Fig. ~\ref{fig:projects}c ). In particular, \texttt{deepseek-R1} and \texttt{gpt-5} demonstrate faster convergence and consistently reach lower final errors than \texttt{claude-sonnet-4.5} and \texttt{gpt-5-chat-latest}. 
These models are able to make early progress in reducing error and continue to refine candidate equations over hundreds of iterations, indicating more reliable exploration–exploitation trade-offs in the symbolic hypothesis space (Supplementary Table~\ref{tab:sr}). 
Although \texttt{claude-sonnet-4.5} performs reasonably in-distribution, it exhibits slower convergence and higher residual errors, particularly in earlier stages of discovery.
By comparison with PySR~\cite{cranmer2023interpretable}, a widely used state-of-the-art baseline for symbolic regression, we observe a significant performance gap from LLM based approaches, where PySR achieves substantially lower accuracy and significantly higher NMSE, especially in the OOD regime (Supplementary Table~\ref{tab:sr}). 
These results reflect LLM's great capability in scenarios such as computation and statistics, and  highlight a key advantage of LLM-guided discovery: the ability to propose based on knowledge, revise, and recombine symbolic structures in a globally informed and knowledgeable manner, rather than relying solely on pure local search over operators.

\qquad In the context of TMC optimization, \texttt{gpt-5}, \texttt{deepseek-R1}, and \texttt{claude-sonnet-4.5} all demonstrate rapid convergence when asked to identify candidates with maximized polarisability.
These models locate the optimal solution within 100 recommendations (fewer than 10 iterations) within a search space of 1.37M TMCs (Fig. \ref{fig:projects}b). 
Notably, \texttt{claude-sonnet-4.5} exhibits superior convergence rates and robustness across varying initialization sets  (Supplementary Text \ref{Supp:tmc_section} and Figure \ref{Supp:tmc_sp}).
Regarding the exploration of the Pareto frontier defined by polarisability and the HOMO-LUMO gap, \texttt{deepseek-R1} yields the most extensive and balanced distribution, effectively covering both the small-gap/high-polarisability and large-gap/low-polarisability regimes  (Fig. \ref{fig:projects}b). 
In contrast, \texttt{claude-sonnet-4.5} is significantly sensitive to the initial population, restricting its exploration primarily to the large-gap/high-polarisability region  (Supplementary Fig. \ref{Supp:tmc_pf}).
In both scenarios, the non-reasoning model, \texttt{gpt-5-chat-latest}, exhibits suboptimal performance compared to its reasoning-enhanced counterparts, underscoring the critical role of derivation and multi-step inference in TMC optimization.

\subsection*{Connecting question- and project-level performance}
\paragraph{Performance on scenarios does not always translate to projects.}
A distinguishing feature of the \texttt{SDE} framework is its ability to bridge question- and project-level evaluations through well-defined research scenarios, enabling direct analysis of error propagation from Q$\&$A to downstream discovery (Fig. \ref{fig:overview}c). 
Top-performing LLMs (e.g., \texttt{gpt-5}) excel at molecular property prediction, SMILES and gene manipulation, protein localization, and algebra. 
Consequently, they demonstrate strong performance in corresponding projects, including organic molecule optimization, gene editing, symbolic regression, and protein design (Fig. \ref{fig:projects}a, Supplementary Fig. \ref{Supp:task_agreement_analysis} and Text \ref{Supp:projects_section}). 
Although the ability of LLMs to generate three-dimensional crystal structures might be questioned given their lack of intrinsic SE(3)-equivariant architecture, we find that top-tier reasoning LLMs generate stable, unique, and novel materials that outperform many state-of-the-art diffusion models. 
This success mirrors their proficiency in related materials scenarios, such as PXRD lattice prediction (Supplementary Table \ref{tab:matllm_results}). 
Conversely, unsatisfactory results across all models in quantum information and condensed matter theory translate directly to the project level: in solving the all-to-all Ising model, most models (with the exception of \texttt{deepseek-R1}) fail to surpass the evolutionary algorithm baseline (Supplementary Fig. \ref{Supp:ising}).

\paragraph{Serendipity in LLM-based optimizations.}
\qquad Interestingly, we observe striking exceptions to the positive correlation between question- and project-level performance. 
For instance, while no model demonstrates high proficiency in TMC-related scenarios (e.g., predicting oxidation states, spin states, and redox potentials), \texttt{gpt-5}, \texttt{deepseek-R1}, and \texttt{claude-sonnet-4.5} all yield excellent efficiency in proposing TMCs with high polarisability and exploring the Pareto frontier within a 1.37M TMC space (Fig. \ref{fig:projects}b). 
This suggests that rigorous knowledge of explicit structure-property relationships is not a strict prerequisite for LLM-driven discovery. 
Rather, the capacity to discern optimization directions and facilitate serendipitous exploration appears more critical. 
Conversely, although top-performing LLMs score highly on questions regarding retrosynthesis, reaction mechanisms, and forward reaction prediction, they struggle to generate valid multi-step synthesis routes. 
Due to frequent failures in molecul·e or reaction validity checks, these models fail to outperform traditional retrosynthesis models on established benchmarks (Supplementary Table \ref{tab:llm_synplanner_results}). 
Notably, \texttt{gpt-4o}, a relatively older model without test-time reasoning, achieves the best results in this project, surpassing both its direct successor (\texttt{gpt-5-chat}) and the reasoning-enhanced variant (\texttt{gpt-5}).
There, \texttt{gpt-5} and \texttt{claude-sonnet-4-5} propose routes that have more synthetic steps when compared to other LLMs. Furthermore, \texttt{gpt-5}, \texttt{gpt-5-chat}, and \texttt{deepseek-R1} generate more invalid SMILES, which is the major culprit in the decreased performance when compared to \texttt{gpt-4o}. However, this does not mean that they are poor at SMILES generation, but rather, worse at generating valid SMILES in the context of the synthesis route (Supplementary Table \ref{tab:llm_synplanner_failure_analysis}).

\paragraph{No single model wins on all projects.}
Across the eight projects, we observe no definitive hierarchy in model performance, where leadership rotates, with models excelling in certain projects while underperforming in others (Fig. \ref{fig:projects}a). 
This variability reflects the composite nature of scientific discovery, which integrates multiple interdependent research scenarios.
Consequently, obtaining outstanding project-level performance requires, at least, proficiency across all constituent scenarios, as a deficit in any single component introduces compounding uncertainty. 
Moreover, the anticipated benefits of strong reasoning enhancements were notably absent in certain projects (such as retrosynthesis and protein design), where such capabilities were expected to be critical (Supplementary Text \ref{Supp:projects_section}).
This suggests that tailored post-training strategies are required to drive further improvements. Notably, the advantage of pre-training corpora appears less decisive in discovery projects than in static question-level evaluation. 
For instance, \texttt{deepseek-R1}, despite showing slightly weaker performance on question-level benchmarks, ranks within the top two across nearly all projects where reasoning is advantageous. 
Ultimately, all contemporary models remain distant from true scientific ``superintelligence'' as no single model excels in all eight (yet limited set of) projects on different themes of scientific discovery.
To effectively orchestrate the loop of scientific discovery, future developments that prioritize balanced knowledge and learning capabilities across diverse scenarios over narrow specialization is desired.

\section*{Discussion}
The integration of large language models (LLMs) into scientific discovery necessitates an evaluation paradigm that transcends static knowledge retrieval. 
While conventional benchmarks have successfully tracked progress in answering general science questions, our results demonstrate that they are insufficient proxies for scientific discovery, which relies on iterative reasoning, hypothesis generation, and evidence interpretation. 
In the scientific discovery evaluation (\texttt{SDE}) framework, we bridge this gap by establishing a tight connection between all questions collected in the benchmark to modular research scenarios, which constitute building blocks in projects aimed for scientific discovery.
There, models are not only evaluated on their ability to answer isolated questions, but also on their capacity to orchestrate the end-to-end research project. 
This dual-layered approach reveals critical insights into the readiness of current foundation LLMs for autonomous scientific inquiry.

\qquad Our question-level evaluation reveals that top-tier models, despite achieving high accuracy on decontextualized benchmarks (e.g., GPQA-Diamond), consistently score lower on \texttt{SDE} questions rooted in active research projects. 
This divergence underscores that proficiency in standard examinations does not guarantee mastery of the nuanced, context-dependent reasoning required for scientific discovery. 
We observe that the gains from scaling model size and test-time compute, strategies that have driven recent breakthroughs in coding and mathematics, exhibit diminishing returns within the domain of scientific discovery. Furthermore, top-performing models from diverse providers exhibit high error correlations, frequently converging on identical incorrect answers for the most challenging questions. 
This shared failure mode suggests that current frontier models are approaching a performance plateau likely imposed by similar pre-training data distributions rather than distinct architectural limitations, thereby motivating the development of discovery-specific objectives and curated domain datasets. 
Project-level evaluation indicates that question-level patterns only partially predict discovery performance and that a model’s capacity to drive a research project relies on factors more complex than a simple linear correlation with its Q$\&$A accuracy.
This implies that precise knowledge of structure-property relationships may be less critical than the ability to navigate a hypothesis space effectively.
Specifically, discerning optimization directions and facilitating serendipitous exploration can compensate for imperfect granular knowledge. 
However, this capability is non-uniform: while LLMs excel at 
optimizing objectives involving well-structured data (e.g., TMC optimization), they struggle with endeavors requiring rigorous, long-horizon planning and strict validity checks, such as retrosynthesis. 
Collectively, these findings highlight the distinct competencies assessed at each evaluation level, underscoring the necessity of comprehensive, multi-scale benchmarking.
Concurrent work from OpenAI\cite{wang2025frontierscience} also indicates solving more challenging Olympic questions, even as comparable to top-tier students\cite{superchem}, do not guarantee success in science research, echoing part of our conclusion.

\qquad Based on these findings, we identify several directions for advancing the utility of LLMs in scientific discovery. 
First, shifting focus from indiscriminate scaling to targeted training on problem formulation and hypothesis generation could bridge current gaps in scientific methodology. 
Second, pronounced cross-model error correlations underscore the urgent need to diversify pre-training data sources and explore novel inductive biases to mitigate shared failure modes.
Third, the integration of robust tool use in fine-tuning is essential, as many of the most challenging research scenarios necessitate a tight coupling between linguistic reasoning and domain-specific simulators, structure builders, and computational libraries. 
Consequently, training and evaluation paradigms must expand beyond textual accuracy to prioritize executable actions—specifically, the capacity to invoke tools, debug execution failures, and iteratively refine protocols in response to noisy feedback.
Finally, given that reasoning enhancements optimized for coding and mathematics yielded negligible gains in many discovery-type projects, developing reinforcement learning strategies tailored specifically for scientific reasoning represents a promising frontier.

\qquad Current \texttt{SDE} encompasses four domains, eight research projects, and 43 scenarios curated by a finite cohort of experts.
Consequently, the benchmark inherently reflects the specific research interests, geographic distributions, and methodological preferences of its contributors.
While disciplines such as earth sciences, social sciences, and engineering are currently unrepresented, the modular architecture of our framework allows for their seamless integration.
Furthermore, reliance on commercial API endpoints introduces unavoidable performance fluctuations due to provider-side A/B testing.
To mitigate this reproducibility challenge, the only solution would be local deployment of open-source models as a critical baseline, enabling independent replication and rigorous ablation free from access constraints. 
Additionally, high computational costs limited our project-level evaluation to a subset of frontier models, assessed using a single evolutionary search strategy and prompting protocol. 
Future research should expand this scope to include alternative optimization algorithms and agentic frameworks, particularly as domain-specific reasoning and tool use are integrated into reinforcement fine-tuning pipelines.
Lastly, we shall not overlook the safety risks posed by increasingly capable biological AI systems. 
Recent efforts, such as built-in safeguard proposals, broader biosecurity roadmaps, jailbreak/red-teaming/watermark techniques and analyses, highlight early steps toward understanding misuse pathways\cite{wang2025call_biosecurity}. 
Despite these constraints, \texttt{SDE} delivers the first integrated assessment of LLM performance across the scientific discovery workflows, providing a robust scaffold upon which the community can build increasingly complex and realistic evaluations.

\section*{Methods}
\phantomsection
\paragraph{Research scenario and question collection.} 
\label{question_collection}
We organized the collection of research scenarios and corresponding questions through a structured, hierarchical collaboration across four scientific domains: biology, chemistry, materials, and physics. 
Each domain was led by a designated group lead with expertise in both scientific field and LLM-based benchmarking (see \hyperref[author_contrib]{\textit{Author Contribution}} section). 
Contributors were grouped by domain according to their research background. 

\qquad Each domain group first identified research scenarios that capture recurring and foundational reasoning patterns in realistic scientific discovery workflows. These scenarios were drawn from ongoing or past research projects and reflect active scientific interests rather than textbook exercises. A ``scenario'' is defined as a modular, self-contained scientific reasoning unit (e.g., forward reaction prediction in chemistry) that can contribute toward solving one or more research projects. Once the domain coverage and key scenarios were defined, contributors were assigned to specific topics based on their expertise to develop concrete question sets under each scenario. 

\qquad Question generation followed a hybrid strategy combining semi-automated and manual curation. When feasible, questions were derived semi-automatically by sampling from existing benchmark datasets (e.g., GPQA) or open-access datasets (e.g., NIST)  and converting structured entries into natural-language question-answer pairs using template scripts. In some cases, domain-specific computational pipelines were used to obtain reference answers. For instance, some molecular descriptors are computed with RDKit\cite{Landrum2025RDKit}. For scenarios lacking structured public records, such as experimental techniques, questions were manually written by domain experts using unified templates to ensure consistency with semi-automated questions. They were subsequently reviewed by the group leads for clarity and relevance.

\qquad To mitigate random variance, each scenario contained at least five validated questions. Question formats included multiple-choice and short-answer types, evaluated through exact-match accuracy, threshold-based tolerance, or similarity scoring to ensure compatibility with automated evaluation pipelines. In this way, ambiguity in scoring the final answers from LLMs is avoided. 

\qquad The resulting dataset spans four domains with 43 distinct scenarios and 1,125 questions, as summarized below (the number of questions in each scenarios is in parenthesis):
\begin{itemize}
    \item Chemistry (276): includes forward reaction prediction (42), retrosynthesis (48), molecular property estimation (58), experimental techniques (29), quantum chemistry software usage (10), NMR-based structure elucidation (31), IR-based structure elucidation (5), MS peak identification (10), reaction mechanism reasoning (10), transition-metal complex property prediction (10), redox potential estimation (8), and mass-to-formula conversion (15).
    \item Materials (486): covers corrosion prediction (60), materials safety classification (140), PXRD crystal system determination (60) and lattice parameter prediction (60), MOF water stability (20) and synthesis (22), battery electrolyte (20), biomaterials (20), composite materials (22), general materials science knowledge (29), and LAMMPS/VASP computational workflows (33).
    \item Biology (200): includes enzymatic reaction prediction (20), protein localization (20), GWAS causal gene identification (20), gene editing design (20), CRISPR delivery strategy (20), drug-likeness/Lipinski assessment (20), descriptor prediction (20), fragment completion (20), matched molecular pair analysis (20), and property-based compound matching (20).
    \item Physics (163): includes astrophysics and cosmology (28), quantum information science (36), condensed matter physics (26), high-energy physics (20), probability and statistics (25), computational physics (21), and core physics knowledge (7).
\end{itemize}

Detailed documentation of dataset sources, question templates, prompt formats and evaluation protocols for all scenarios are accessible in \hyperref[data_avail]{\textit{Data Availability}} section. Detailed curation procedures and representative example questions are provided in the Supplementary Information.

\phantomsection
\paragraph{Research project collection.}
\label{project_collection}
We curated eight research projects across biology, chemistry, materials, and physics, each involving multiple modular research scenarios  (Supplementary Table \ref{Supp:correspondence}). For example, a project for retrosynthesis path design would naturally involves scenarios of single-step retrosynthesis, reaction mechanism analysis, and forward reaction prediction, among many others. Each research project was formulated as a search or optimization problem following the scientific discovery loop, using LLMs as proposals over a hypothesis space (e.g., the space of all possible molecular structures, symbolic equations). These hypotheses were then examined by computational oracles to access the fitness, which were then fed into LLMs to refine their proposals. Without loss of generality, we chose evolutionary optimization as a simple yet efficient search approach. The evolutionary optimization for each project followed a general workflow: (1) initialization: the process was initialized with a set of hypotheses (cold-start generation from LLMs or warm-up from a predefined set), (2) mutation, crossover, and \textit{de novo} proposal: LLMs were prompted to generate offspring based on parent hypotheses sampled from the pool, and (3) selection: after each generation of offspring was sampled, selection was made by keeping top-ranked hypotheses from the evaluated oracle scores in the parent and offspring hypotheses. The step (2) and (3) were repeated until the convergence of the search process or exceeding the maximum number of oracle calls. In practice, the implementation of each problem was flexible to incorporate task-specific descriptions and adaptations following the establishment of those projects from previous literature. We now detail the descriptions for each project below:
\begin{itemize}
\item \textit{(chemistry) Retrosynthesis pathway design.-} Retrosynthesis tackles the planning problem to find a reaction pathway to synthesize molecules. Given a target molecule, it aims to decompose the structure into commercially available precursors (i.e. building blocks), often over many reaction steps in a process known as multi-step retrosynthesis. In this project, each decomposition step must abide by an available reaction template, which encodes a specific chemical transformation, thus grounding the LLM's proposed decompositions to fixed rules. 
This process defines a planning problem as the LLM must decide the \textit{strategy} in which it decomposes target molecules (e.g. which part of the molecules to decompose first and how). Reference molecules and their associated synthesis routes are used as context to the LLM and extracted from Chen et al.\cite{chen2020retro}, which in turn is based on the reaction data from the United States Patent and Trademark Office (USPTO)\cite{Lowe2017}.  
The evaluation follows the protocol of the authors' original work\cite{wang2025llm}.
\item \textit{(chemistry) Molecule optimization.-} The discovery of novel molecules with desired properties is important in molecular science such as drug discovery. In this project, LLMs are used to search over the vast chemical space to find molecular structures with optimal properties. The evaluation follows the protocol of the authors' original work\cite{wang2024efficient}.
\item \textit{(materials) Transition metal complex (TMC) optimization.-} Designing functional TMCs with combinatorial explosion from the choices of ligands. This project pushes LLMs to generate candidate TMCs with desired HOMO-LUMO gap and polarisability under an evolutionary optimization loop, showcasing LLMs’ deep understanding of transition metal chemistry. The evaluation follows the protocol of the authors' original work\cite{tmcopt}.
\item \textit{(materials) Crystal structure discovery.-} Discovering novel crystal structures computationally is challenging, as candidate structures must simultaneously satisfy multiple physical constraints, including three-dimensional periodicity, chemically valid atomic coordination, charge neutrality, and thermodynamic stability. In this project, LLMs are used to perform implicit crossover and mutation on reference parent structures under an evolutionary framework, generating novel crystal structures with low energy above the hull. The evaluation follows the protocol of the authors' original work\cite{gan2025large}. 
\item \textit{(biology) Protein sequence optimization.-} Protein engineering aims to develop novel protein sequences with improved functions. The search space consisted of protein sequences containing 4-250 mutation sites depending on the dataset, with 20 possible amino acid types per site. In this project, each objective is defined by an oracle function that maps a sequence to a scalar fitness value, where LLMs are used to optimize protein sequence to reach the optimal fitness.
The evaluation follows the protocol of the authors' original work\cite{wang2025large}.

\item \textit{(biology) Gene editing.-} Genetic perturbation experiments aims to find subsets out of many possible genes that result in a specific phenotype when they are perturbed. In this project, LLMs are pushed to design new experiments for proposing perturbation for finding new phenotypes. The evaluation follows the protocol of Ref. \cite{gene_editing}. 
\item \textit{(physics) Symbolic regression.-} Discovering mathematical models governing scientific observations presents significant challenges and prevents understanding natural phenomena in physics. This project aims to find symbolic equations that recover the experimental observations measured by the errors in the simulated observations with LLMs. The evaluation follows the protocol of authors' original work\cite{shojaee2025llmsr,shojaee2025llmsrbench}. 
\item \textit{(physics) Solving Ising model.-} Discovering the best spin configurations that minimize the Ising model energy presents significant challenges due to vast combinatorial configuration spaces. In this project, LLMs are used to mimic the discovery process of human scientists for inferring the optimal configuration that minimizes the Ising model’s Hamiltonian, leveraging LLMs to accelerate the search over exponentially large configuration spaces. 
\end{itemize}

\phantomsection
\paragraph{Model evaluation.}
\label{eval}
\textit{Question-level.-} All evaluations for questions in \texttt{SDE} were performed using a customized fork of \texttt{lm-evaluation-harness}\cite{lm_eval_harness}. Each scenario is specified by a YAML configuration that loads its corresponding Hugging Face dataset. During evaluation, deterministic decoding ($\mathrm{temperature} = 0, \mathrm{do\_sample=false}$) was used unless models requires other parameter setting explicitly (for example, \texttt{gpt-5} only accepts $\mathrm{temperature} = 1$).
Across domain, for most scenarios, standardized prompt and output formats were used to enable LLMs to present their final response within an XML-style tag (e.g.: <answer>...</answer>). It would be captured by regrex filter and stripped before scoring. Unless otherwise noted, metrics follow exact-match accuracy, case- and punctuation-insensitive.

\qquad Most biology, chemistry, materials, and physics scenarios share this evaluation mode, with domain-specific utilities handling numeric and special outputs.
For chemistry, molecular structure outputs (e.g., structure elucidation via spectra) are canonicalized with RDKit and scored by Tanimoto similarity, while numeric predictions (e.g., redox potentials) are evaluated by checking whether the prediction falls within a scenario-defined tolerance window around the reference value.
In materials, classification scenarios (e.g., corrosion prediction) use exact match, and lattice-parameter regression grant partial credit per correctly predicted axis within 3 Å.
Biology scenarios extend exact match to structured descriptors (e.g., HBD, MW, LogP) with numeric tolerances, weighted partial score (CRISPR delivery prediction), and RDKit canonicalized molecular structures.
In physics, algebraic responses are parsed through a symbolic verifier (\textit{math-verify} package) that grants credit for mathematically equivalent expressions.
Across scenarios, metrics are bounded in $[0,1]$ and higher values indicate better performance. Scenario-level scores (typically exact match, but occasionally similarity, tolerance-based accuracy, or MAE) are obtained as the average across all questions in that scenario. Domain scores are then aggregated by simple mean across topics to form the question-level component of the \texttt{SDE} benchmark.

\textit{Project-level.-} All evaluations for research projects in \texttt{SDE} were performed using \texttt{sde-harness}\cite{sdeharness}. We aggregated the performance for each project into a single score, normalizing the scale of each sub-objectives, and averaged the performance across sub-objectives to obtain a single score (Fig. \ref{fig:projects}a). Considering the cost of evaluating projects is much higher than that of questions, all projects are only evaluated on \texttt{gpt-5-chat-latest}, \texttt{gpt-5}, \texttt{claude-sonnet-4.5}, and \texttt{deepseek-R1}, with both best non-reasoning and reasoning models tested.  Details for each project are described in Supplementary Sec. \ref{Supp:projects_section}.

\phantomsection
\section*{Data Availability}
\label{data_avail}
All datasets used in this study are publicly available. The complete collection of question–answer pairs, associated metadata, configurations, and scientific discovery projects that constitute the \texttt{SDE} benchmark is hosted under the \textit{deep-principle} organization.

\begin{itemize}
\item \textbf{Question-level resources}: 
    \begin{itemize}
        \item \textbf{Datasets}. Question–answer datasets are organized by scientific domain (science\_{\text{chemistry}, \text{materials}, \text{biology}, \text{physics}}) and are available at \url{https://huggingface.co/deep-principle/datasets}.
        \item \textbf{Code}. All code and utilities required to reproduce the question-level results—including YAML configurations, prompt templates, and evaluation scripts are available at \url{https://github.com/deepprinciple/lm-evaluation-harness/tree/main}
    \end{itemize}

\item \textbf{Project-level datasets and oracles}: \url{https://github.com/HowieHwong/sde-harness}

\end{itemize}

\section*{Acknowledgment}
\phantomsection
Z.S., J.L., Q.Z., H.J., and C.D. would like to thank our entire team from Deep Principle for helpful discussions and support.
C.D. thanks Wenhao Gao, Ben Blaiszik, Miles Cranmer,  Peichen Zhong for helpful discussions. Y.D. acknowledges the support of Cornell University. J.H. acknowledges support from the University of Cambridge Harding Distinguished Post-
graduate Scholars Programme.
C.P.G. acknowledges the support of an AI2050 Senior Fellowship, a Schmidt Sciences program, the National Science Foundation (NSF), the National Institute of Food and Agriculture (USDA/NIFA), the Air Force Office of Scientific Research (AFOSR), and Cornell University. A.N. gratefully acknowledges support from the Eric and Wendy Schmidt AI in Science Postdoctoral Fellowship, a Schmidt Sciences, LLC program, and startup funding from UCLA.
S.M.M. gratefully acknowledges support from  the University of Toronto’s Acceleration Consortium and the Data Science Institute, and the Natural Sciences and Engineering Research Council of Canada (NSERC) for his research program, including support for T.M.P., S.T.K., and M.R.K.

\section*{Author Contributions}
\phantomsection
\label{author_contrib}
Coordination lead and writing of original draft: Zhangde Song and Jieyu Lu;
Project collection and evaluation lead and writing of original draft: Yuanqi Du;
Coding lead: Botao Yu and Yue Huang;
Materials question collection and evaluation lead: Thomas M. Pruyn;
Chemistry question collection and evaluation lead: Kehan Guo;
Physics question collection and evaluation lead: Xiuzhe Luo;
Biology question collection and evaluation lead: Yuanhao Qu;
Protein design project implementation and evaluation: Yinkai Wang;
Gene editing project implementation and evaluation: Yi Qu and Chenru Duan;
Retrosynthesis project implementation and evaluation: Jeff Guo;
Molecule optimization project implementation and evaluation: Haorui Wang;
TMC optimization project implementation and evaluation: Zhangde Song and Chenru Duan;
Crystal design project implementation and evaluation: Jingru Gan;
Symbolic regression project implementation and evaluation: Parshin Shojaee;
Ising model project implementation and evaluation: Di Luo;
Chemistry question collection: Jieyu Lu, Yi Qu, Jeff Guo, Andres M. Bran, Gen Li, Qiyuan Zhao, and Shao-Xiong Lennon Luo;
Physics question collection: Yuxuan Zhang, Xiang Zou, Wanru Zhao, Yifan Zhang, Wucheng Zhang, Shunan Zheng, and Saiyang Zhang;
Materials question collection: Sartaaj Takrim Khan, Mahyar Rajabi, Samantha Paradi-Maropakis, Tony Baltoiu, Fengyu Xie, and Tianyang Chen;
Biology question collection: Kexin Huang, Yinkai Wang, Weiliang Luo, and Meijing Fang;
Visualization: Xin Yang and Lixue Cheng;
Supervision: Jiajun He, Soha Hassoun, Xiangliang Zhang, Chandan K. Reddy, Chao Zhang, Zhiling Zheng,  Mengdi Wang, Le Cong, Carla P. Gomes, Chang-Yu Hsieh, Aditya Nandy, Philippe Schwaller, and Heather J. Kulik, and Haojun Jia;
Supervision, conceptualization, and methodology: Huan Sun and Seyed Mohamad Moosavi;
Supervision, conceptualization, methodology, and writing of original draft: Chenru Duan

\section*{Competing interests}
The authors declare that they have no competing financial interests at this time.

\end{refsegment}
\mainsegmentfalse
\printbibliography[title={References}, segment=1]

\clearpage
\setstretch{1}

\begin{refsegment}
\appendix

\title{\textit{Supplementary Information} for "Evaluating LLMs in Scientific Discovery"}

\maketitle

\renewcommand{\thesection}{\arabic{section}}  
\renewcommand{\thetable}{\arabic{table}}  
\renewcommand{\thefigure}{\arabic{figure}}
\setcounter{figure}{0}
\setcounter{table}{0}

\makeatletter
\renewcommand{\fnum@figure}{\textbf{Figure \thefigure}. }
\renewcommand{\fnum@table}{\textbf{Table \thetable. }}

\addcontentsline{toc}{section}{Abbreviation}
\section*{Abbreviation}
The following is the list of abbreviations utilized in the main paper and Supplementary Information.
\begin{itemize}
    \item LLM: Large Language Model
    \item SDE: Scientific Discovery Evaluation
    \item Q\&A: Question and Answer
    \item API: Application Programming Interface
    \item RL: Reinforcement Learning
    \item MSE: Mean Squared Error
    \item NMSE: Normalized Mean Squared Error
    \item AUC: Area Under the Curve
    \item AUC$_{\text{top-}k}$: Area Under the Curve of Top-$k$ Metric
    \item XML: Extensible Markup Language

    \item AIME: American Invitational Mathematics Examination
    \item MMMU: Multidiscipline Multimodal Benchmark for Universality
    \item GPQA: Graduate-level Google-Proof Scientific Q\&A
    \item SWE-bench: Software Engineering Benchmark
    \item $\tau$-bench: Tool–Agent–User Interaction Benchmark
    \item HLE: Humanity’s Last Exam

    \item NMR: Nuclear Magnetic Resonance
    \item IR: Infrared Spectroscopy
    \item MS: Mass Spectrometry
    \item TMC: Transition Metal Complex
    \item MOF: Metal Organic Framework
    \item PXRD: Powder X-Ray Diffraction
    \item VASP: Vienna Ab-initio Simulation Package
    \item LAMMPS: Large-scale Atomic/Molecular Massively Parallel Simulator
    \item GFN2-xTB: Geometry-optimized eXtended Tight Binding Method
    \item SC: Synthetic Complexity (small molecule synthesizability metric)
    \item SA: Synthetic Accessibility (small molecule synthesizability metric)
    \item USPTO: United States Patent and Trademark Office
    \item MCTS: Monte Carlo Tree Search
    \item SMILES: Simplified Molecular Input Line Entry System
    \item HOMO-LUMO gap: Highest Occupied Molecular Orbital -- Lowest Unoccupied Molecular Orbital gap
    \item $E_d$: Energy above the convex hull
    \item SUN: Stable, Unique, Novel (crystal structure metric)
    \item CHGNet: Crystal Hamiltonian Graph Neural Network
    \item CDVAE: Crystal Diffusion Variational Autoencoder
    \item DiffCSP: Diffusion Model for Crystal Structure Prediction
    \item GA: Genetic Algorithm

    \item GWAS: Genome-Wide Association Study
    \item CRISPR: Clustered Regularly Interspaced Short Palindromic Repeats
    \item IFNG: Interferon-gamma
    \item AAV: Adeno-Associated Virus
    \item GFP: Green Fluorescent Protein

    \item ID: In-Domain
    \item OOD: Out-of-Domain

    \item RDKit: Cheminformatics Software Toolkit
    \item ZINC: Small Molecule Database
    \item molSimplify: Transition Metal Complex Toolkit
    \item PySR: Python Symbolic Regression Package
    \item StructureMatcher: Pymatgen Structural Comparator
    \item MatBench: Materials Benchmark Dataset
    \item MatBench-bandgap: MatBench Bandgap Prediction Dataset
\end{itemize}

\begin{figure*}[!htbp]
    \includegraphics[width=1.0\textwidth]{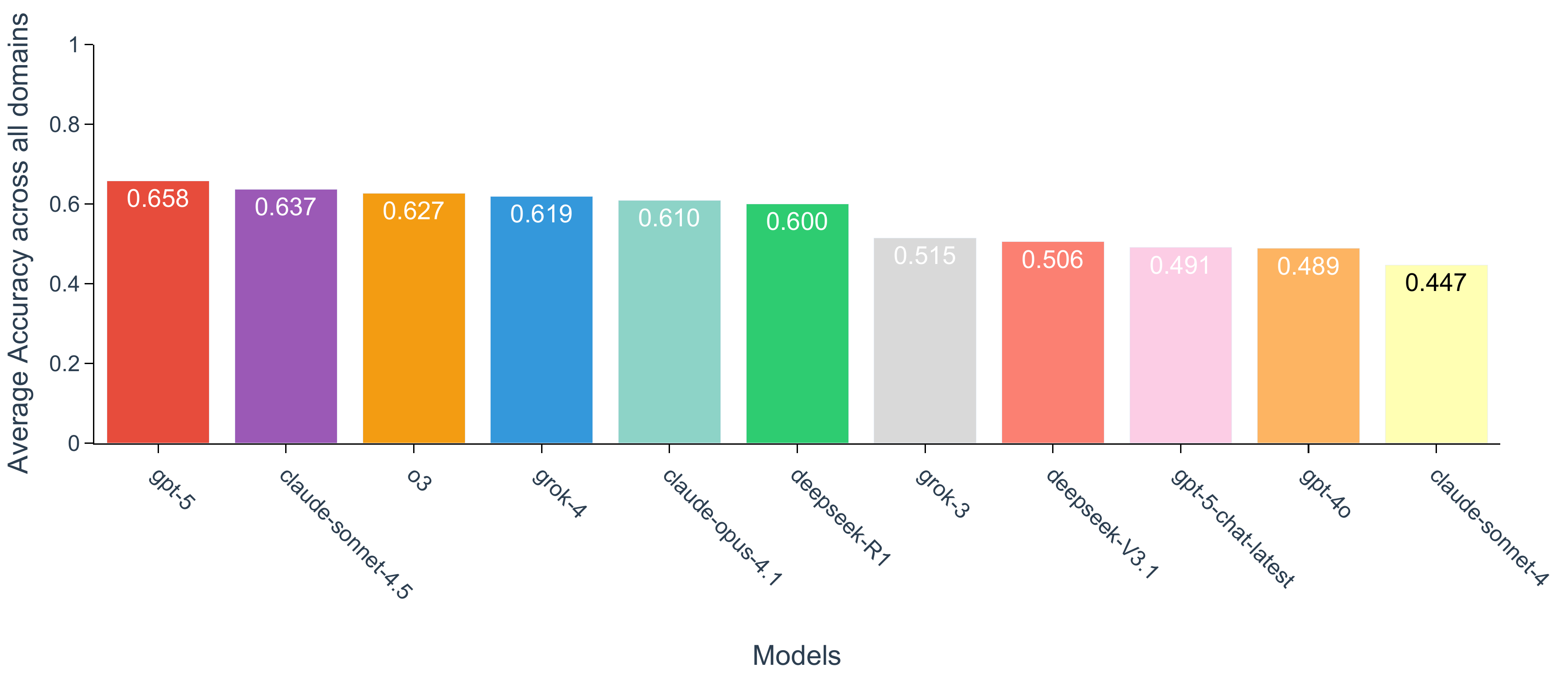}
    \caption{\textbf{Average model accuracy across all 43 research scenarios.} The models are ranked by the average accuracy.
    }
    \label{Supp:model_average_performance}
\end{figure*}

\begin{figure*}[!htbp]
    \includegraphics[width=1.0\textwidth]{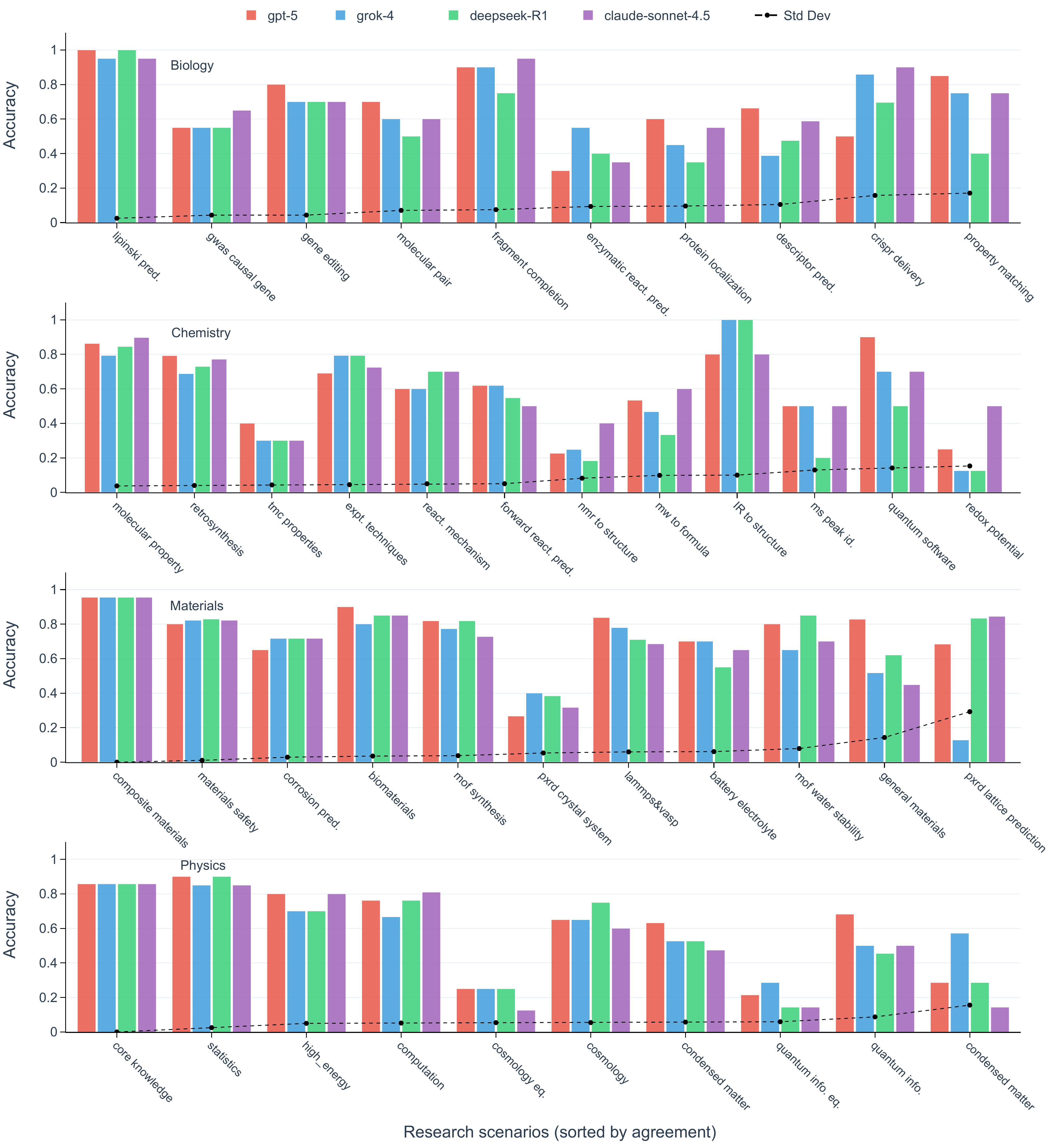}
    \caption{\textbf{Per-scenario accuracy for top-performing models at four domains.} \texttt{gpt-5} is colored in red, \texttt{grok-4} in blue, \texttt{deepseek-R1} in green, and \texttt{claude-sonnet-4.5} in purple. Research scenarios are ranked with increasing standard deviations of the four model accuracies for each domain, which are shown as the black dashed lines.
    }
    \label{Supp:task_agreement_analysis}
\end{figure*}

\begin{figure*}[!htbp]
    \includegraphics[width=0.8\textwidth]{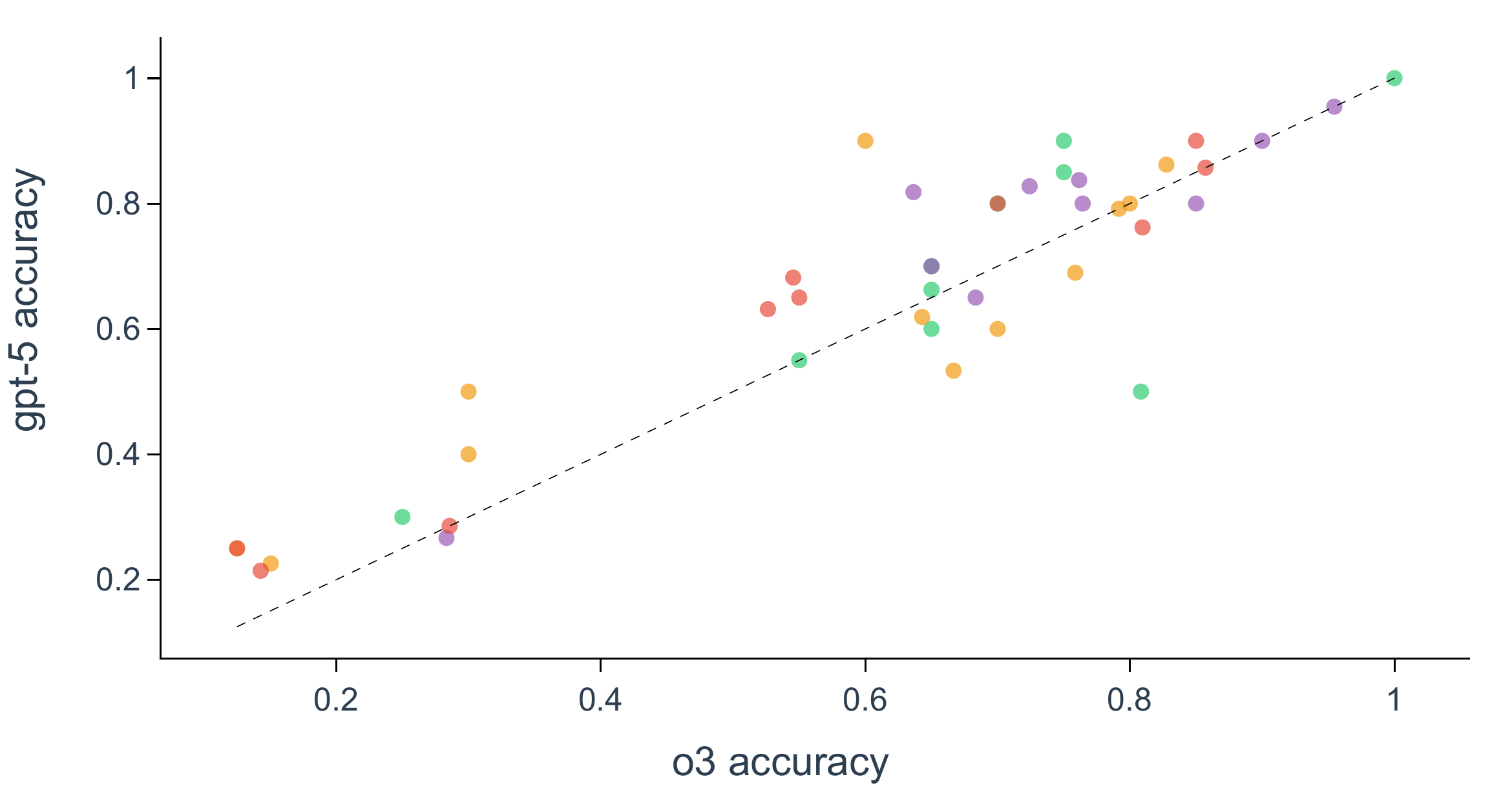}
    \caption{\textbf{Per-scenario accuracy for \texttt{gpt-5} and \texttt{o3}.} Scenarios in biology are colored in green, chemistry in orange, materials in purple, and physics in red. Parity is shown with a black dashed line.
    }
    \label{Supp:o3_vs_gpt5_performance_diff}
\end{figure*}

\begin{figure*}[!htbp]
    \includegraphics[width=1.0\textwidth]{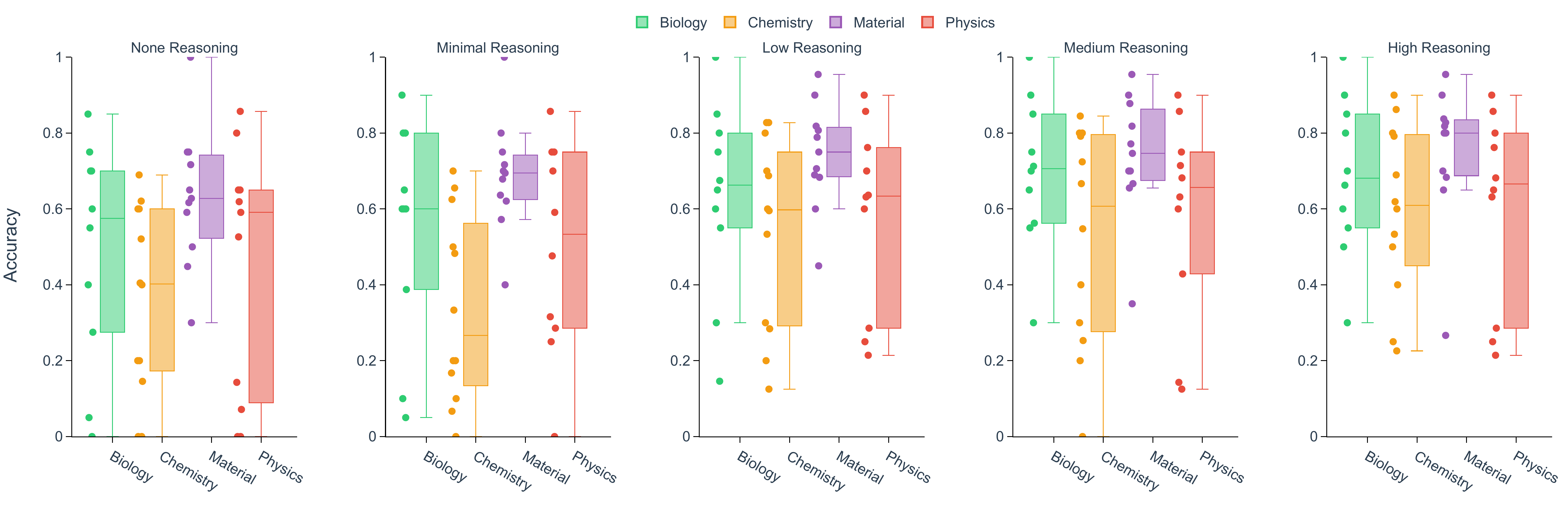}
    \caption{\textbf{Accuracy of \texttt{gpt-5} at various reasoning levels.} Scenarios in biology are colored in green, chemistry in orange, materials in purple, and physics in red. A box plot is shown for the distribution where all points are explicitly added. 
    }
    \label{Supp:gpt5_reasoning_distributions}
\end{figure*}

\begin{figure*}[!htbp]
    \includegraphics[width=1.0\textwidth]{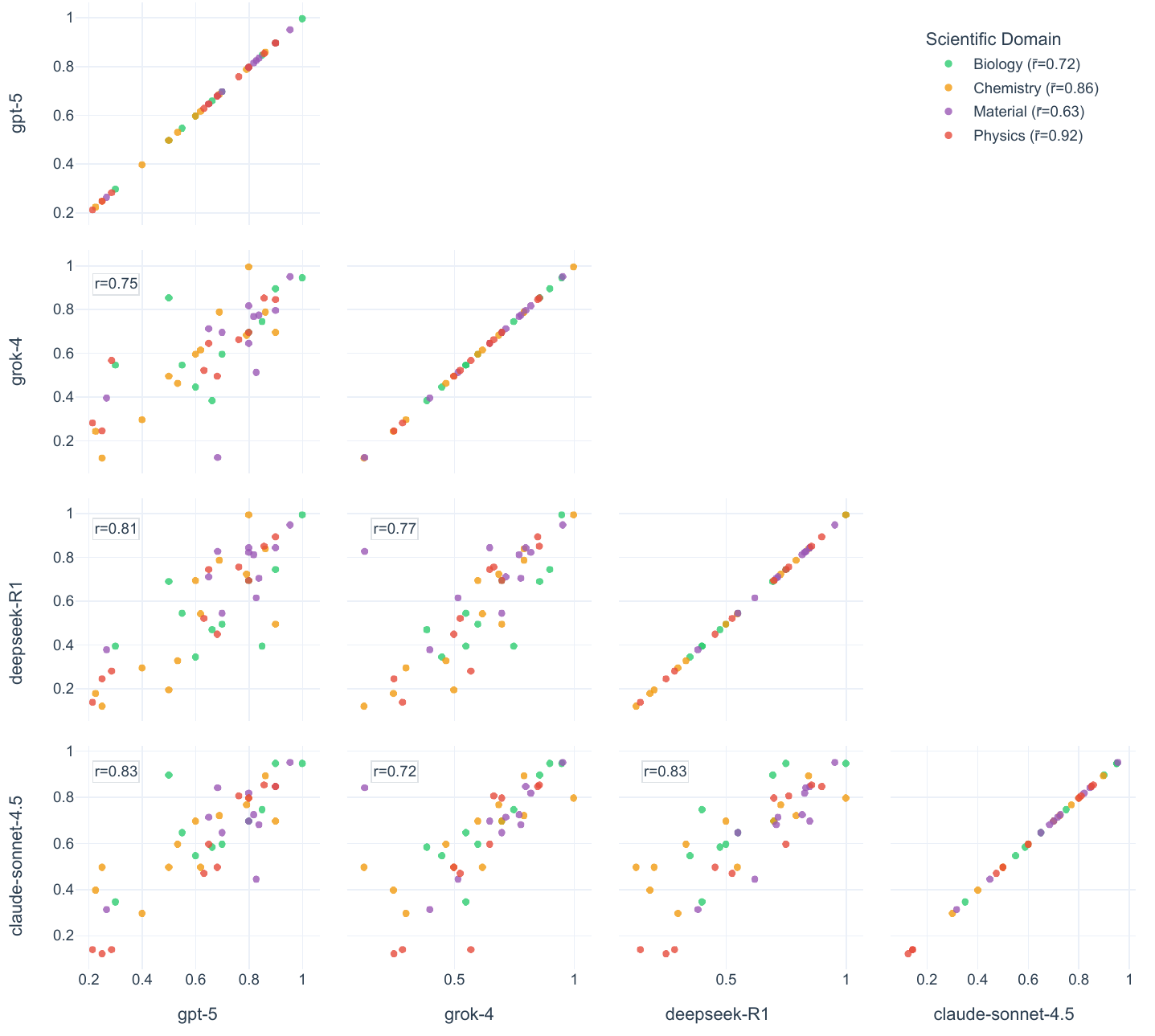}
    \caption{\textbf{Pairwise per-scenario accuracy for top-performing models.} Scenarios in biology are colored in green, chemistry in orange, materials in purple, and physics in red. Pearson's r is shown for each pairwise plot, where the overall Pearson's r for each domain is displayed on the legend.
    }
    \label{Supp:model_scatter_matrix}
\end{figure*}

\begin{figure*}[!htbp]
    \includegraphics[width=1.0\textwidth]{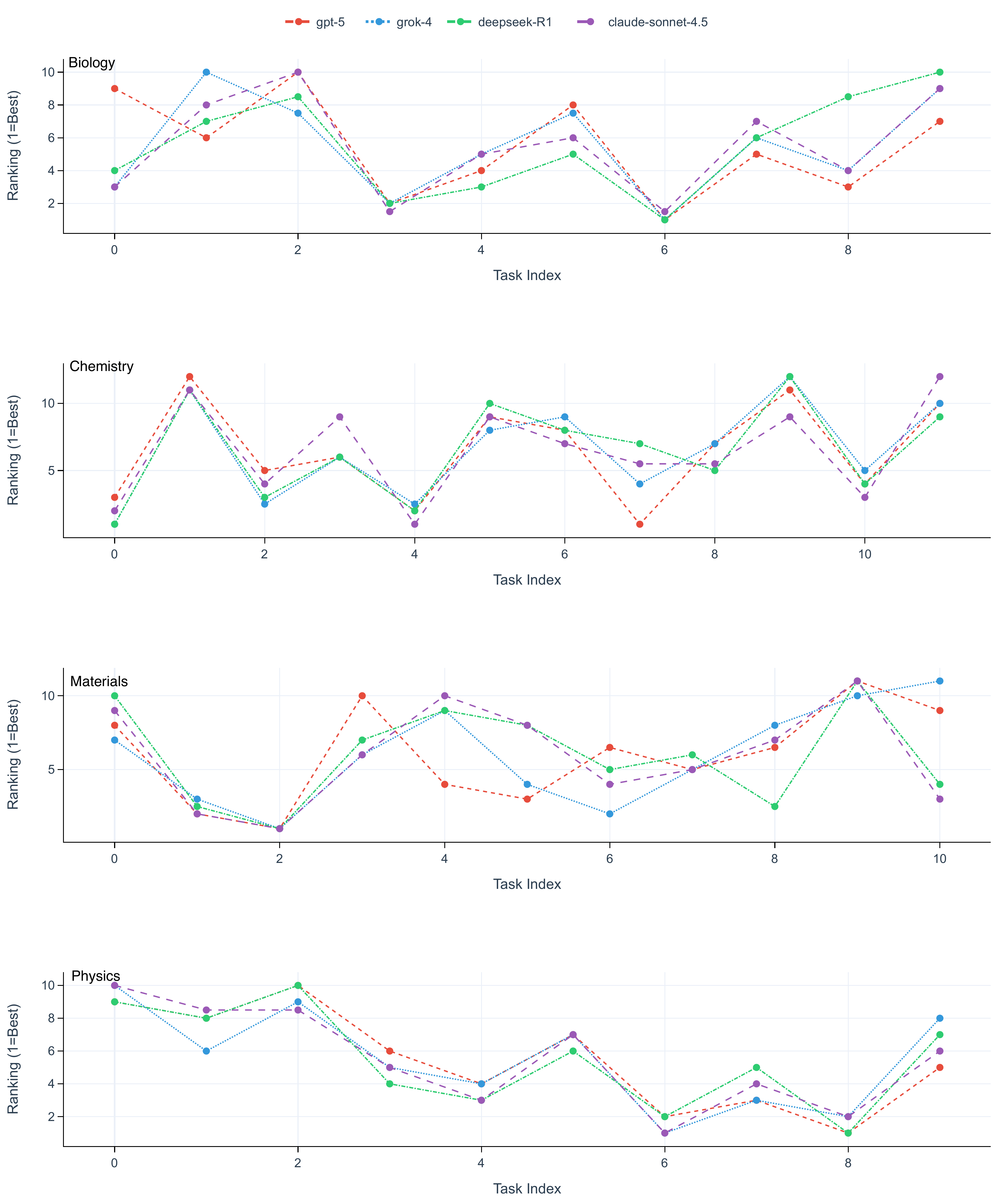}
    \caption{\textbf{Ranking vs. research scenario for top-performing models.} \texttt{gpt-5} is colored in red, \texttt{grok-4} in blue, \texttt{deepseek-R1} in green, and \texttt{claude-sonnet-4.5} in purple. 
    }
    \label{Supp:ranking_consistency}
\end{figure*}

\begin{figure*}[!htbp]
    \includegraphics[width=1.0\textwidth]{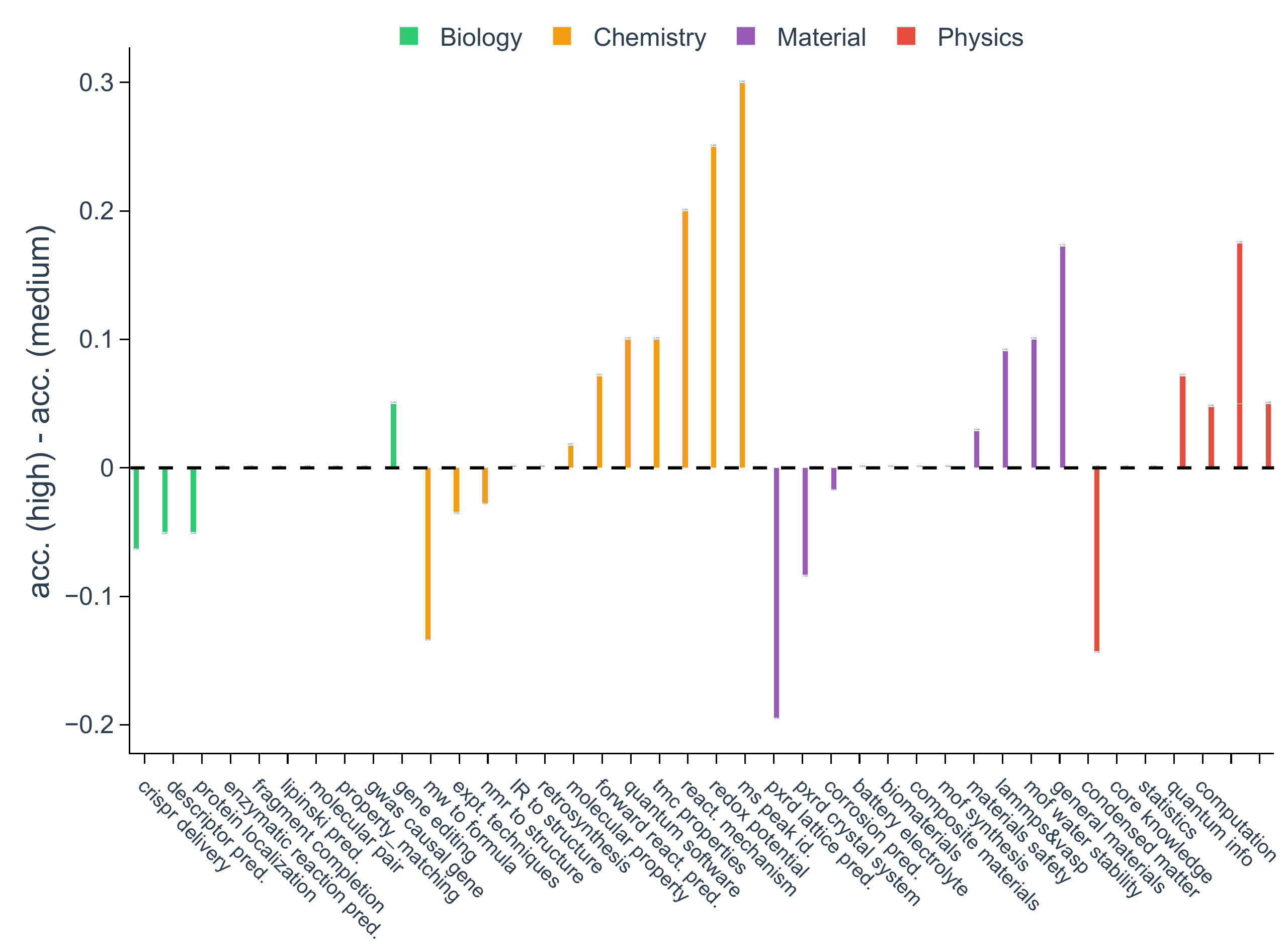}
    \caption{\textbf{Performance difference between \texttt{gpt-5} with high and medium reasoning efforts.} Scenarios in biology are colored in green, chemistry in orange, materials in purple, and physics in red. A dashed line is shown for no accuracy difference.
    }
    \label{Supp:medium_vs_high_reasoning_comparison}
\end{figure*}

\clearpage

\begin{figure*}[!htbp]
    \includegraphics[width=0.7\textwidth]{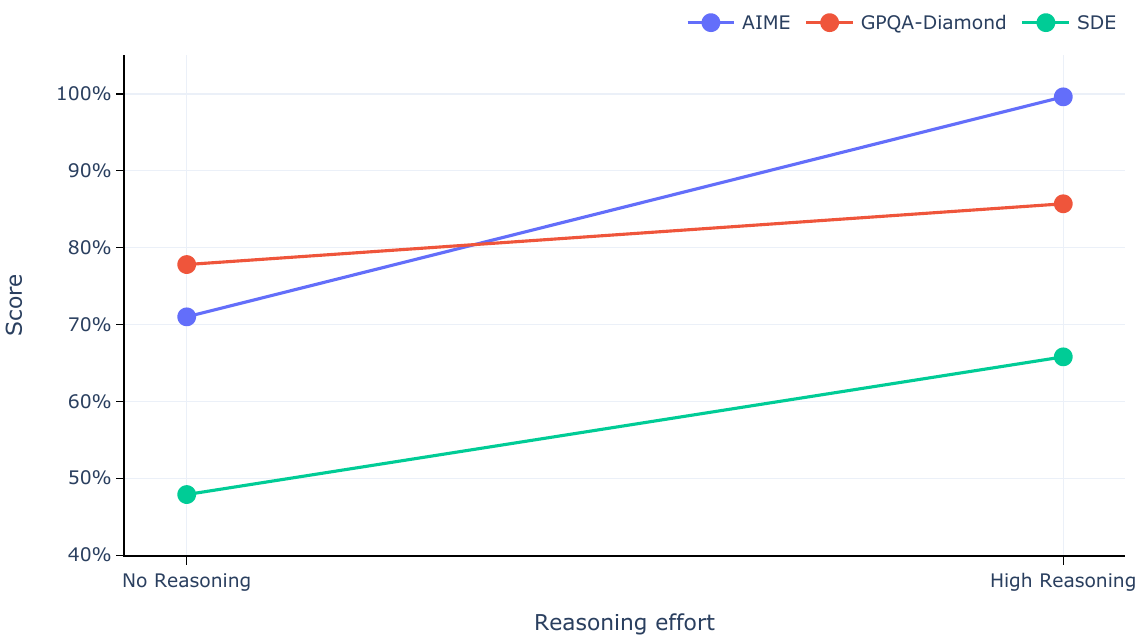}
    \caption{\textbf{Reasoning versus no-reasoning performance on three benchmarks for \texttt{gpt-5}.} AIME 2025 is colored in blue, GPQA-Diamond in red, and SDE question-level evaluation in green. 
    }
    \label{Supp:reasoning_bench}
\end{figure*}

\begin{figure*}[!htbp]
    \includegraphics[width=0.7\textwidth]{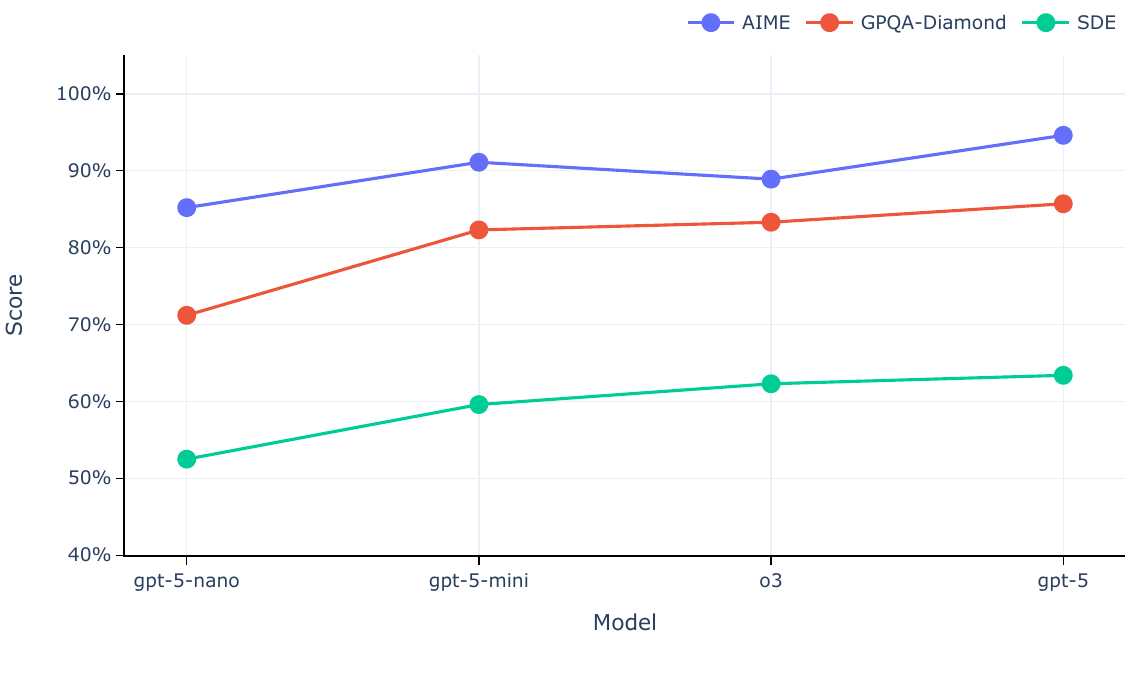}
    \caption{\textbf{Model size effects on three benchmarks illustrated by \texttt{gpt-5-nano}, \texttt{gpt-5-mini}, \texttt{gpt-5}, and \texttt{gpt-o3}.} AIME 2025 is colored in blue, GPQA-Diamond in red, and SDE question-level evaluation in green. 
    }
    \label{Supp:size_bench}
\end{figure*}

\begin{figure*}[!htbp]
    \includegraphics[width=1.0\textwidth]{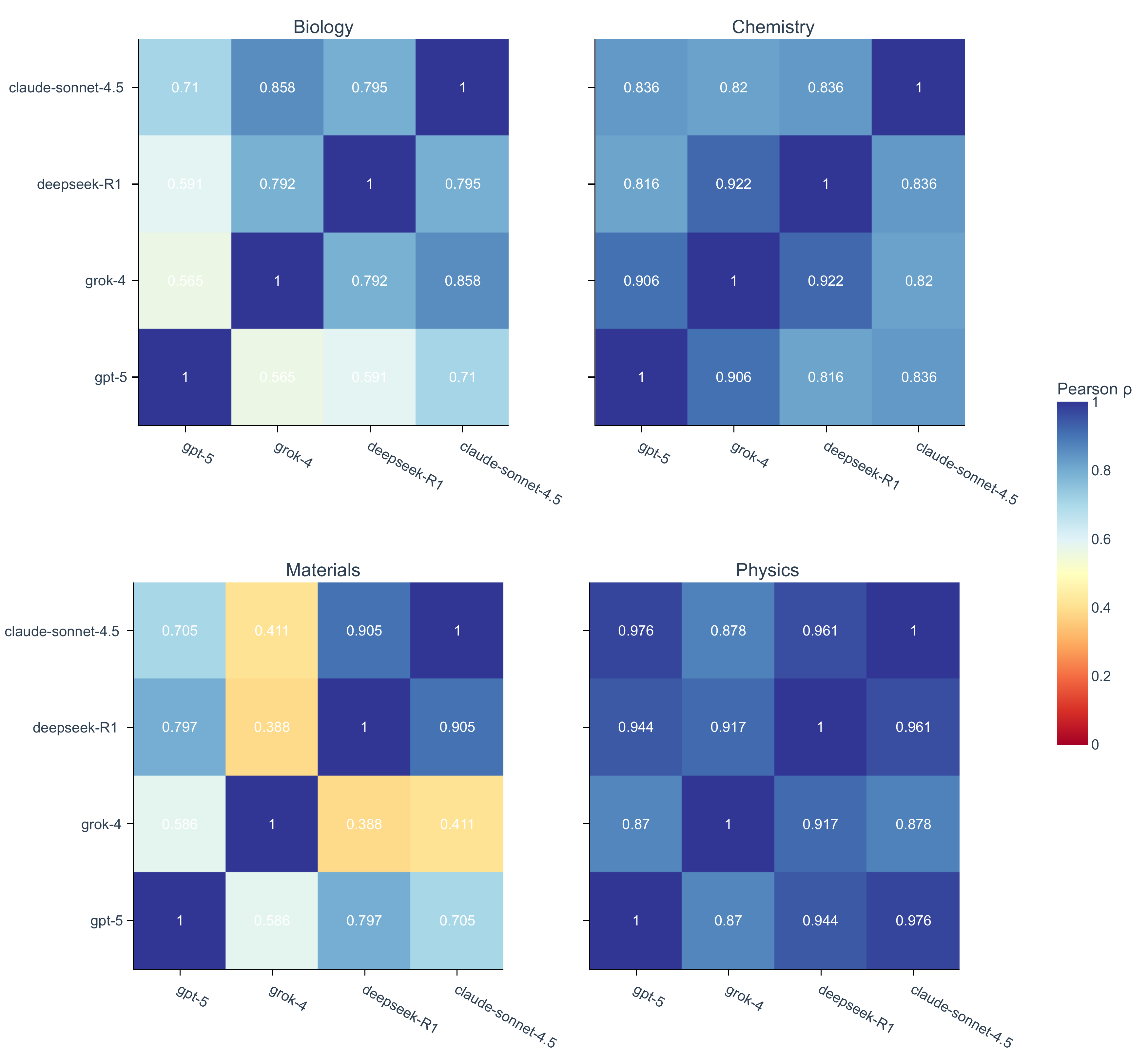}
    \caption{\textbf{Pearson correlation heatmaps for top-performing models.} The results are shown by domains.
    }
    \label{Supp:pearson_heatmaps_by_domain}
\end{figure*}

\begin{figure*}[!htbp]
    \includegraphics[width=0.6\textwidth]{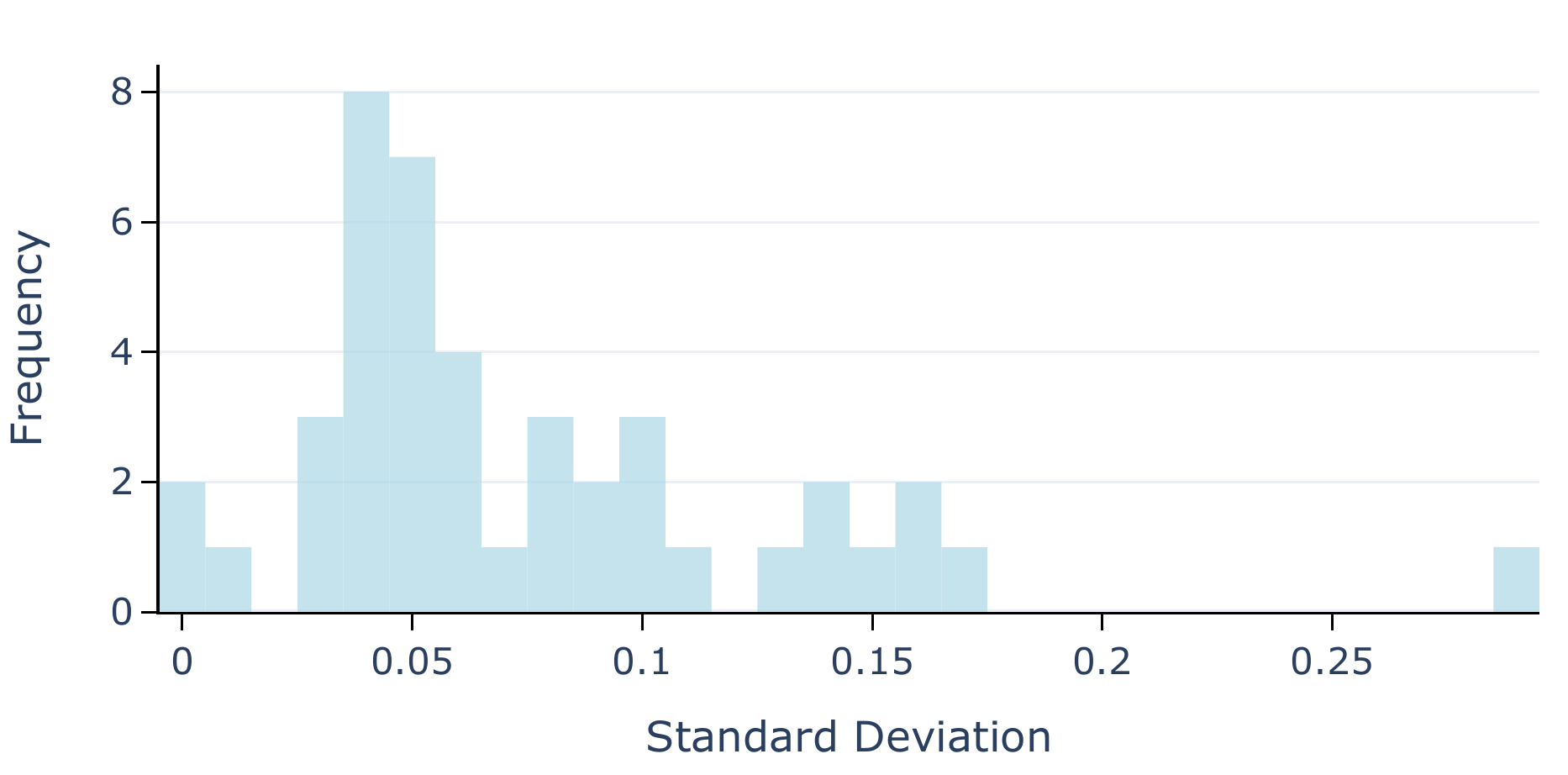}
    \caption{\textbf{Distribution of standard deviation for four top-performing models on 46 tasks.} The four top-performing models are \texttt{gpt-5}, \texttt{grok-4}, \texttt{deepseek-R1}, and \texttt{claude-sonnet-4.5}.
    }
    \label{Supp:std_dev_distribution}
\end{figure*}

\begin{figure*}[!htbp]
    \includegraphics[width=0.7\textwidth]{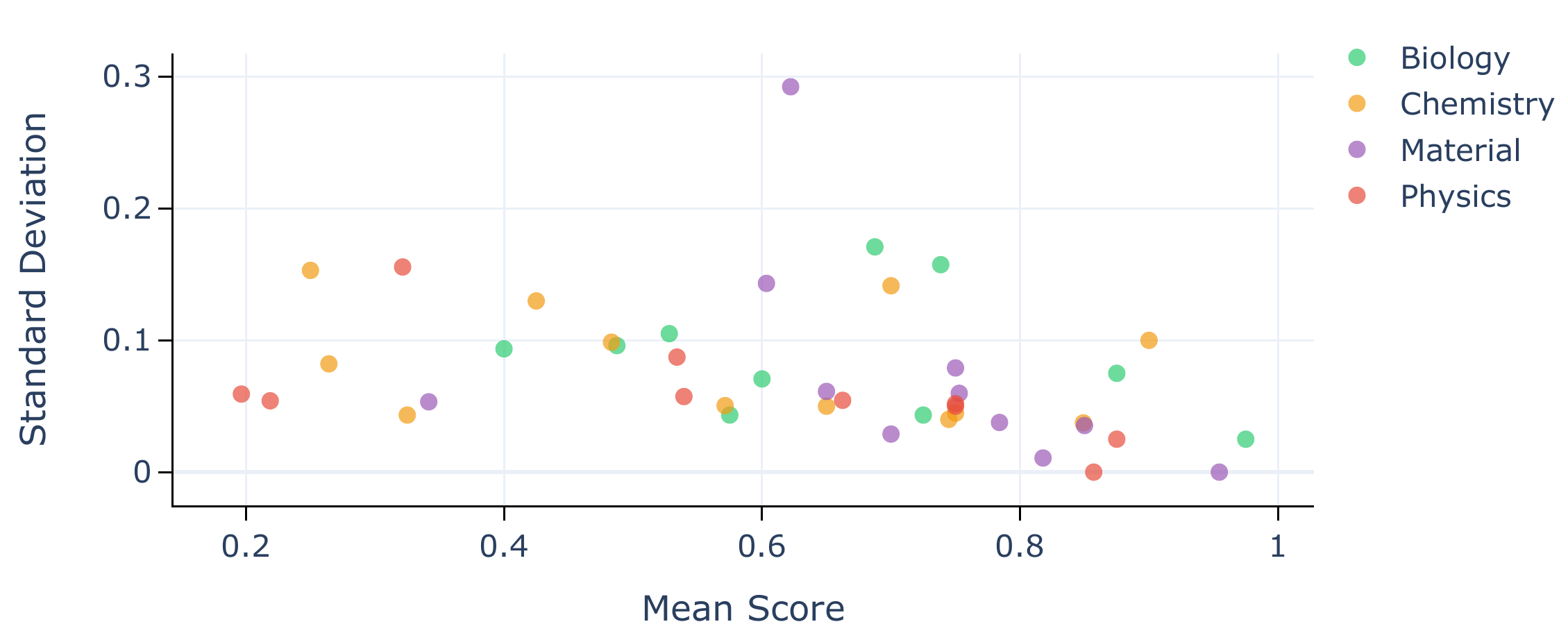}
    \caption{\textbf{Standard deviation vs. mean value on accuracy for four top-performing models on 46 tasks.} The four top-performing models are \texttt{gpt-5}, \texttt{grok-4}, \texttt{deepseek-R1}, and \texttt{claude-sonnet-4.5}. Scenarios in biology are colored in green, chemistry in orange, materials in purple, and physics in red.
    }
    \label{Supp:std_vs_mean}
\end{figure*}

\begin{figure*}[!htbp]
    \includegraphics[width=1.0\textwidth]{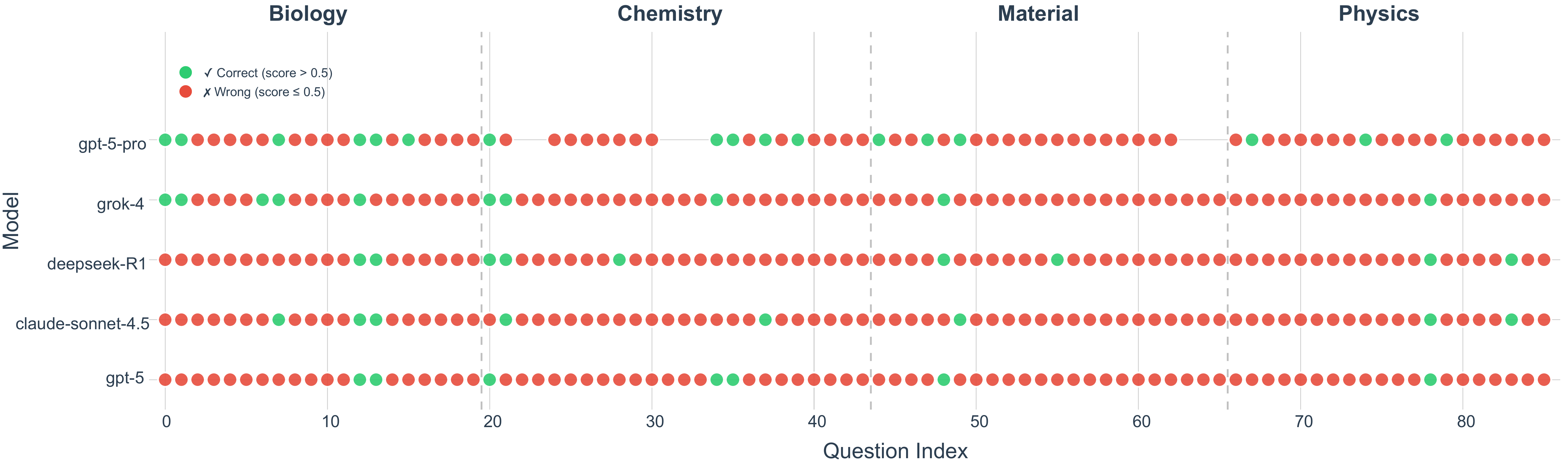}
    \caption{\textbf{Question-level performance correlation among five models on \textsc{SDE-hard}}. Each question is marked by its correctness, green for correct and red for incorrect. 
    }
    \label{Supp:question_performance_scatter}
\end{figure*}

\begin{figure*}[!htbp]
    \includegraphics[width=0.5\textwidth]{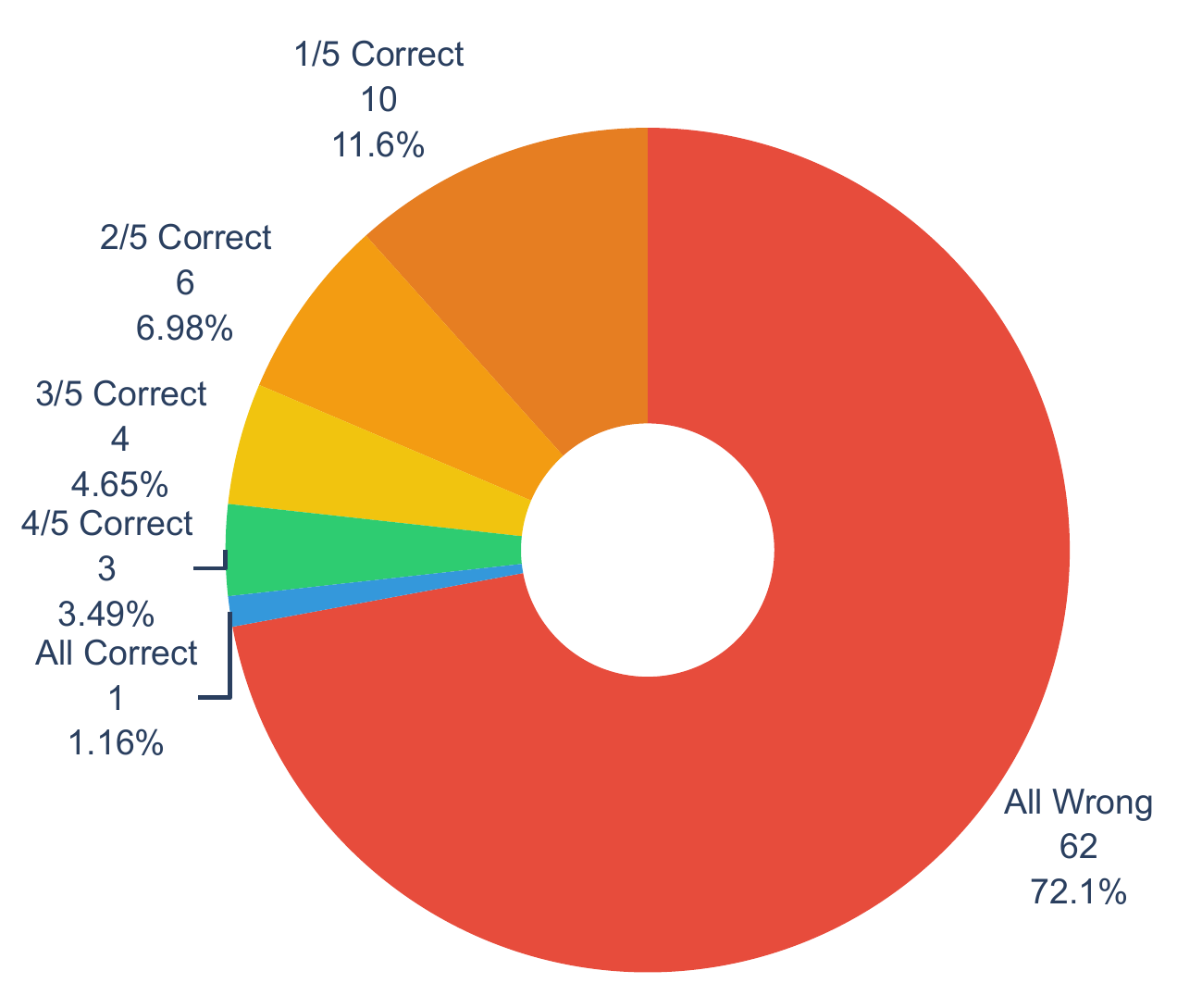}
    \caption{\textbf{Doughnut plot for analysis of model consensus on \textsc{SDE-hard}} The five models are \texttt{gpt-5-pro}, \texttt{gpt-5}, \texttt{grok-4}, \texttt{deepseek-R1}, and \texttt{claude-sonnet-4.5}.
    }
    \label{Supp:sde_hard_agreement}
\end{figure*}

\begin{figure*}[!htbp]
    \includegraphics[width=0.5\textwidth]{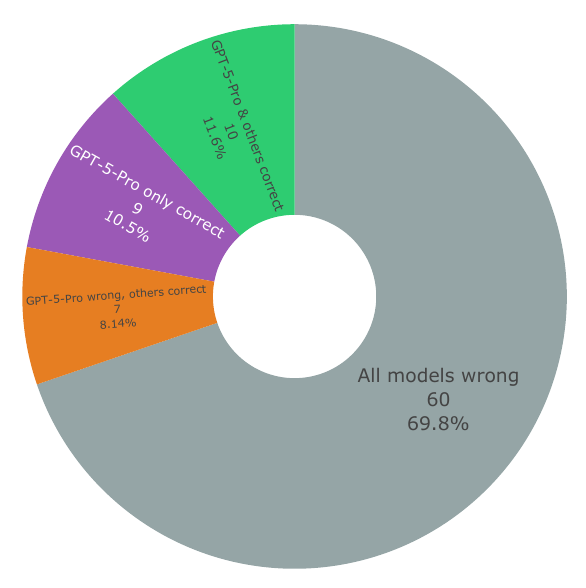}
    \caption{\textbf{Doughnut plot for analysis of \texttt{gpt-5-pro} performance over other four top-performing models}. The four models are \texttt{gpt-5}, \texttt{grok-4}, \texttt{deepseek-R1}, and \texttt{claude-sonnet-4.5}.
    }
    \label{Supp:sde_hard_pro_correct}
\end{figure*}

\clearpage
\section{Question curation details}
\subsection{Question curation for chemistry tasks}
This section describes the task-specific curation procedures for the chemistry domain in the SDE benchmark. The chemistry subset comprises 276 questions spanning twelve distinct task types, designed to reflect recurring reasoning patterns in both wet-lab and dry-lab chemical research. Question construction followed a hybrid strategy combining semi-automated generation from public resources with manual expert curation for scenarios lacking structured records. Representative example questions illustrating these task formats are provided in \nameref{example_questions}.
\paragraph{Forward reaction prediction (42)}
Questions for forward reaction prediction were sourced from two parts. A subset was adapted from existing benchmark-style questions (e.g., GPQA), while others were sampled from the USPTO reaction dataset covering diverse reaction classes. Reaction entries were filtered to retain single-step transformations with unambiguous reactant–product mappings. Structured reaction records were then converted into natural-language multiple-choice questions using standardized templates. Additional questions were manually curated based on real-world wet-lab scenarios to capture practical reaction reasoning beyond dataset distributions.
\paragraph{Retrosynthesis (48)}
Retrosynthesis questions were derived from both benchmark-style sources and template-based sampling from USPTO (USPTO\_TPL\cite{schwaller2021mapping}). 
For the latter, we used known reaction templates as a reference for what constitutes a chemically reasonable retrosynthetic step. Given a target molecule, we identified which types of bond disconnections are consistent with known reaction pattern. We then formulated multiple-choice questions in which chemically reasonable disconnections served as correct answers, while implausible or poorly motivated disconnections—drawn from unrelated reaction patterns were used as distractors.
\paragraph{Molecular property estimation (58)}
Molecular property questions were constructed using a mixture of benchmark-derived examples and molecules sampled from the ZINC database\cite{zinc2012}. Molecules were represented using SMILES or IUPAC names. Reference properties—including logP, topological polar surface area (TPSA), number of rotatable bonds, ring counts, molecular weight, and Tanimoto similarity—were computed using RDKit\cite{Landrum2025RDKit}. Questions were framed to require comparative or ordering-based reasoning (e.g., ranking molecules by a given property).
\paragraph{NMR-based structure elucidation (31)}
NMR structure elucidation questions were manually curated from real experimental scenarios and published supplementary information. Each question provides molecular formulae and multi-modal spectroscopic data (¹H NMR, ¹³C NMR, and occasionally MS), requiring models to infer the complete molecular structure. Reference answers are represented as SMILES strings. During evaluation, predicted and reference structures are canonicalized using RDKit and scored via Tanimoto similarity between Morgan fingerprints, allowing partial credit for near-correct structures.
\paragraph{Redox potential estimation (8)}
Redox potential questions were curated from published electrochemical studies. Given the three-dimensional structure of a photocatalyst or metal complex, models are asked to predict the reduction potential relative to a specified reference electrode. Answers are numeric and evaluated using a task-specific tolerance band (0.25) around the reference value, reflecting experimental uncertainty.
\paragraph{Experimental techniques (29)/Quantum chemistry software usage (10)/Reaction mechanism reasoning (10)/MS peak identification (10)/IR-based structure elucidation (5)/Transition-metal complex property prediction (10)/Mass-to-formula conversion (15)}
Questions were entirely manually curated by domain experts. These questions were inspired by real world wet-lab/dry-lab experimental procedures or literature. All questions are formulated as either short answers or multiple-choice problems, and answers were evaluated with exact match accuracy.

\subsection{Question curation for biology tasks}
This section describes the task-specific curation procedures for the biology domain in the SDE benchmark. 
The biology subset contains 200 questions spanning 10 task types, designed to capture recurring reasoning patterns in biochemistry, cell biology, genetics and molecular biology. 
Most questions are derived from established biological databases (e.g., ChEMBL, Therapeutics Data Commons, UniProt, GWAS Catalog), supplemented by expert-curated scenarios where structured records are unavailable.
Representative example questions are provided in \nameref{example_questions}.

\paragraph{Molecular descriptor prediction (20)}
Descriptor prediction questions assess a model’s ability to infer fundamental physicochemical properties directly from molecular structure.
Drug-like molecules were sampled with reference labels obtained from CycPeptMPDB and ChEMBL records.
Reported values for hydrogen bond donors (HBD), hydrogen bond acceptors (HBA), molecular weight (MW), and logP were cross-checked for consistency with RDKit calculations.
Questions are formulated as structured short-answer tasks requiring numerical prediction of multiple descriptors from a given SMILES representation. During evaluation, exact matches are required for integer-valued descriptors (HBD, HBA), while MW and logP are scored using a predefined numerical tolerance to account for minor computational variability.

\paragraph{Lipinski rule assessment (20)}
Lipinski assessment questions evaluate integrative reasoning over multiple molecular descriptors.
Molecules and labels were sampled from CycPeptMPDB and ChEMBL.
Given a SMILES string, models are required to determine whether the molecule satisfies Lipinski’s Rule of Five by implicitly estimating HBD, HBA, MW, and logP and verifying all threshold conditions.
Questions are framed as binary multiple-choice classification tasks and evaluated via exact match accuracy.

\paragraph{Enzymatic reaction prediction (20)}
Enzymatic reaction prediction questions probe biochemical transformation reasoning under enzyme-specific constraints.
Reaction records were sourced from the ECREACT 1.0 dataset, supplemented with authoritative enzyme annotations from IUBMB resources.
Each question provides reactant structures, an enzyme EC number, and curated supporting information describing enzyme specificity and cofactors.
Models are asked to predict the transformed product in canonical SMILES form, accounting for reaction mechanism and physiological protonation state.
Evaluation is performed by exact match comparison between canonicalized predicted and reference SMILES.

\paragraph{Fragment-based molecular completion (20)}
Fragment completion questions evaluate structure--property reasoning under partial structural information.
Masked molecular fragments and associated target property values (e.g., LD50) were sampled from the Therapeutics Data Commons.
Models are required to identify which candidate molecule, when completing the masked fragment, best matches the specified property value.
Questions are formulated as multiple-choice selection tasks, requiring both structural compatibility and property inference.

\paragraph{Matched molecular pair comparison (20)}
Matched molecular pair questions test comparative property reasoning between closely related molecules.
Property values (LD50 or Vdss) were obtained from the Therapeutics Data Commons.
Given two structurally similar molecules, models must predict whether the comparison molecule exhibits a higher or lower property value relative to a reference molecule.
This task emphasizes sensitivity to localized structural modifications. Questions are framed as binary classification tasks and evaluated via exact match accuracy.

\paragraph{Property-based molecular matching (20)}
Property-based matching questions assess similarity-based reasoning across structurally diverse candidates.
Given a target molecule with a known property value and a set of candidate molecules, models are asked to identify the candidate whose property value is numerically closest to the target. Property values (LD50 or Vdss) were obtained from the Therapeutics Data Commons.
Questions are evaluated via exact match over multiple-choice selections.

\paragraph{Protein subcellular localization prediction (20)}
Protein localization questions evaluate sequence-based biological inference.
Protein sequences and reference localization annotations were derived from UniProtKB/Swiss-Prot, focusing on well-annotated proteins with experimentally supported localization labels.
Given an amino acid sequence, models are asked to predict the most likely subcellular compartment (e.g., nucleus, cytoplasm, mitochondrion, secreted), based on known localization signals and sequence features.
Questions are formulated as multiple-choice classification tasks and evaluated via exact match accuracy.

\paragraph{GWAS causal gene identification (20)}
GWAS causal gene questions probe integrative genetic reasoning across phenotype, gene function, and locus context.
GWAS loci and associated phenotypes were curated from the NHGRI-EBI GWAS Catalog.
For each question, a phenotype and a list of genes within the associated genomic locus are provided.
Models are required to identify the gene most likely to be causally related to the phenotype, based on known biological functions, pathways, and disease associations.
Answers are evaluated via exact match against expert-curated reference genes.

\paragraph{CRISPR delivery strategy selection (20)}
CRISPR delivery questions evaluate experimental design reasoning in gene-editing workflows.
Scenarios were manually curated by domain experts based on common experimental constraints reported in the CRISPR literature and community best practices.
Given a biological context (e.g., editing primary T cells), models are asked to select the most appropriate and second-most appropriate delivery methods from a predefined list (e.g., viral vectors, electroporation, lipid nanoparticles).
Evaluation is performed using a weighted scoring scheme that assigns partial credit based on the relative suitability of the selected delivery methods.

\paragraph{Gene-editing protocol reasoning (20)}
These questions assess practical reasoning about CRISPR screening and gene-editing protocols.
Scenarios were inspired by widely used experimental workflows reported in the functional genomics and CRISPR screening literature.
Models are asked to select the most scientifically sound option in situations such as maintaining sgRNA library diversity or designing screening logistics.
Questions are framed as multiple-choice tasks and evaluated via exact match accuracy.

\subsection{Question curation for physics \& math tasks}
This section describes the task-specific curation procedures for the physics domain in the SDE benchmark.
The physics subset contains 163 questions with 8 task categories, covering astrophysics, quantum information, condensed matter physics, high-energy physics, probability and statistics, computational physics, and core physics knowledge.
Unlike domains dominated by large curated datasets, many physics problems involve symbolic manipulation, mathematical equivalence, and conceptual abstraction that resist straightforward data-driven generation, and are primarily manually curated by domain experts, reflecting the fact that many physics reasoning tasks—particularly those involving analytical derivations or conceptual arguments—lack large-scale structured public benchmarks.

Tasks include both multiple-choice and short-answer formats. For short-answer physics questions involving mathematical expressions, symbolic relationships, or derived formulas, answers are evaluated using a symbolic equivalence verification framework.
Predicted and reference answers are parsed as mathematical expressions and verified for equivalence up to algebraic transformations with \textit{math\_verify} package \cite{Kydlicek_Math-Verify_Math_Verification}.
This approach ensures that mathematically correct but syntactically different expressions are treated as equivalent, while incorrect derivations are penalized.
When symbolic verification is not applicable, a normalized exact-match fallback is used.
Representative example questions are provided in \nameref{example_questions}.

\paragraph{Astrophysics and cosmology (28)}
Astrophysics and cosmology questions probe conceptual reasoning and domain knowledge related to galaxy classification, large-scale structure, and cosmological models.
Questions were manually written by domain experts and include both multiple-choice conceptual questions and short-answer reasoning tasks.

\paragraph{Quantum information science (36)}
Quantum information questions assess reasoning about quantum states, entanglement measures, quantum channels, and information-theoretic quantities.
Questions are manually curated to reflect reasoning steps commonly encountered in quantum information research rather than rote formula recall.
This task category includes both multiple-choice questions and short-answer problems.

\paragraph{Condensed matter physics (26)}
Condensed matter questions cover statistical mechanics, solid-state physics and quantum many-body systems.
Tasks include conceptual multiple-choice questions as well as short-answer derivations.

\paragraph{High-energy physics (20)}
High-energy physics questions focus on conceptual understanding of quantum field theory and particle physics.
These questions are formulated as multiple-choice problems that require recognizing theoretical correspondences rather than recalling isolated facts.
All questions in this category are multiple-choice problems.

\paragraph{Probability and statistics (25)}
Probability and statistics questions assess mathematical reasoning relevant to statistical physics and data analysis.
The task set includes both multiple-choice questions and short-answer derivation problems.

\paragraph{Computational physics (21)}
Computational physics questions evaluate understanding of numerical methods and algorithmic error analysis.
Questions are formulated as multiple-choice problems.

\paragraph{Core physics knowledge (7)}
Core physics questions test foundational operator algebra and elementary physical principles that frequently serve as sub-steps in more complex reasoning tasks.
These questions are framed as multiple-choice problems and evaluated via exact match accuracy.

\subsection{Question curation for materials tasks}
This section describes the task-specific curation procedures for the materials domain in the SDE benchmark. The materials subset contains 540 questions, spanning thirteen distinct task types. These tasks are organized around three key use-cases for LLMs in materials discovery applications: Question-Answering (Q\&A), Materials Property Prediction, and Text-Mining. 

Q\&A-style questions were manually curated by experts and are meant to simulate the types of questions researchers encounter during routine materials discovery workflows. These questions are designed to assess the breadth and coverage of an LLM’s foundational materials science knowledge. Property Prediction questions test an LLM’s ability to reasonably predict material properties given various textual descriptions and representations of materials as inputs. Lastly, Text-Mining questions evaluate an LLM’s ability not only to extract structured materials data from unstructured text, but also to draw conclusions from text with ambiguous or implicitly stated meanings, commonly referred to as argument mining.

\paragraph{General materials science Q\&A (29)}
Basic materials science questions were sourced from expert domain knowledge and first-principle theories and equations. Questions were framed to represent a variety of lab experimental design and result scenarios, with distractor responses formulated based on common misconceptions.

\paragraph{Composite materials Q\&A (22)}
Questions relating to composite materials knowledge were compiled from fundamental mechanics theories and domain knowledge. Lab-scenario and more broad composite material knowledge was constructed into natural-language multiple-choice questions utilizing standard templates, with implausible mechanics of materials and modelling responses as distractors.

\paragraph{Biomaterials Q\&A (20)}
Biomaterials questions were constructed by adapting exam-style scenario-based questions, derived from educational and experimental experience, and supplemented by relevant textbooks and literature where applicable \cite{park2002biomaterials, wozniak2021influence, corliss2016macrophages}. 12 questions were case-study style, in which the correct biomaterial phenomenon to explain the observations, or the correct conclusion based on the biomaterial phenomenon observed, must be selected. 5 questions were design style, in which the correct design choice to meet certain criteria must be selected. 3 questions were basic knowledge recall style questions. All questions were formatted as multiple choice, with questions and answers in natural language, often using acronyms or common names for compounds or phenomena.

\paragraph{Metal-organic framework synthesis Q\&A (20)}
These questions were manually authored by a domain expert and formatted as multiple-choice questions. They cover a range of materials synthesis, characterization, and analysis topics for metal-organic framweworks (MOFs), with incorrect options designed to reflect common misconceptions or plausible alternatives.

\paragraph{Battery electrolyte Q\&A (20)}
These questions were manually authored by a domain expert and formatted as multiple-choice questions focusing on liquid electrolytes for lithium metal batteries. The questions span fundamental concepts and multi-step problems requiring more advanced reasoning, covering Li-ion solvation, solid–electrolyte interphase behavior, electrolyte composition, and performance metrics such as Coulombic efficiency. Incorrect options reflect plausible alternatives or common misunderstandings within the field.

\paragraph{LAMMPS and VASP Q\&A (33)}
These Q\&A pairs were manually authored by a domain expert and formatted as multiple-choice questions focusing on atomistic simulation workflows using LAMMPS and VASP. The questions span fundamental software capabilities, simulation setup, and methodological considerations, as well as multi-step reasoning related to interpreting simulation behavior and selecting appropriate computational approaches. Incorrect options reflect plausible alternatives or common misunderstandings in the use of molecular dynamics and electronic-structure tools.

\paragraph{Corrosion prediction of metals against organic solvents (60)} 
Questions for corrosion prediction were constructed using data from the corrosion-survey-database published by NACE International (The Corrosion Society). The dataset comprises experimentally reported corrosion outcomes of organic solvents against a wide range of metallic materials. Organic compounds were represented using SMILES strings, and each question was formulated as a binary classification task indicating whether a given compound exhibits corrosive behavior toward any of 87 metals. Ground-truth labels were assigned based on documented corrosion observations, with positive labels indicating the presence of corrosion and negative labels indicating no observed corrosion. Questions were framed in a true/false format to assess a model’s ability to reason about chemical structure–corrosion relationships across diverse organic solvents and metal types.

\paragraph{Polymer glass-transition prediction (24)}
Questions for polymer glass-transition temperature (Tg) prediction were constructed using data from the NeurIPS Open Polymer Prediction 2025 Kaggle competition\cite{liu_neurips_polymer_2025}. The dataset comprises experimentally reported Tg values for a diverse set of polymer systems. Polymers were represented using structured molecular descriptors provided in the dataset, and each question was formulated as a property prediction task requiring estimation of Tg from the given polymer representation. Ground-truth labels were assigned directly from the reported experimental measurements. Questions were framed to assess a model’s ability to reason about polymer structure–property relationships relevant to thermomechanical behavior.

\paragraph{Lattice prediction from PXRD (60)}
We evaluated lattice parameter predictions using a dataset of ionic compounds from the Materials Project\cite{jain2013commentary} and metal-organic frameworks (MOFs) from the CoRE-2019 database\cite{chung2019advances}. Each model input was represented by a simulated powder X-ray diffraction (PXRD) pattern, its corresponding Miller indices, and the identified crystal system. To derive these features, we utilized Pymatgen’s\cite{ong2013python} SpacegroupAnalyzer - a symmetry-determination tool built upon the Spglib\cite{togo2024spglib} library - extracting the crystal system of each structure. The PXRD patterns and associated Miller indices were either simulated using Pymatgen’s XRDCalculator\cite{de2012structure} under a $CuK\alpha$ radiation source, or already computed in previous works\cite{khan2025connecting}. Ground truth lattice parameters were extracted directly from the source crystallographic information files (CIFs) to serve as the target for our evaluations.

\paragraph{Crystal system prediction from PXRD (60)}
In this task, we aimed to classify a material’s crystal system directly from its observed PXRD peaks. The dataset composition and the underlying methodologies for symmetry determination and PXRD simulation remain consistent with the workflow described in the PXRD Lattice Prediction section. For this classification model, the input features consist of the peak positions (defined by $2\theta$) and their corresponding relative intensities. We extracted these features using a peak-finding algorithm from the SciPy\cite{virtanen2020scipy} library, applying a minimum intensity threshold of 5 to ensure the inclusion of only significant diffraction signals.

\paragraph{PXRD denoising (30)}
This task focuses on denoising a corrupted PXRD pattern to recover its original, "clean" state. The target data consists of the simulated PXRD patterns for ionic compounds, as defined in the PXRD Lattice Prediction section. To prepare the inputs, the original patterns were smoothed using a Gaussian function and discretized into 1D arrays of 200 data points, where each element represents a specific diffraction intensity. As these arrays are standardized to a constant size and uniform $2\theta$ range, explicit $2\theta$ values were omitted to avoid redundancy. Finally, to simulate experimental artifacts, we introduced random Gaussian noise ($\mu = 0, \sigma = 1$) to the normalized intensities. This methodology can be fully described in this work\cite{khan2025connecting}.

\paragraph{Materials safety prediction (140)}
Safety-related questions were constructed using data sampled from the PubChem database, focusing on two hazard categories: inhalation toxicity (80 questions) and flammability (60 questions). Molecules were represented using SMILES strings, and each question was framed as a binary classification task based on Globally Harmonized System (GHS) hazard annotations. Ground-truth labels were assigned according to whether a given molecule was reported to carry the corresponding GHS hazard code, with positive labels indicating the presence of the hazard and negative labels indicating its absence. Questions were presented in a true/false format (e.g., “Does this molecule classify as a flammable liquid according to GHS classification?”) to evaluate a model’s ability to infer safety-relevant properties directly from molecular structure.

\paragraph{Metal-organic framework water stability text-mining (20)}
Questions were expertly curated using the experimentally measured water-stability data from the MOF-ChemUnity database \cite{pruyn2025mof}. Each question is based on unstructured textual descriptions of reported stability behavior, requiring extraction of relevant evidence and, where necessary, inference of stability outcomes from implicitly stated or qualitative information. 

\section{Representative example questions}
\label{example_questions}
This section provides representative example questions for selected tasks described in previous sections, illustrating the diversity of question formats, reasoning requirements, and evaluation protocols used in SDE.
\subsection{Chemistry domain}
\noindent\fbox{%
  \parbox{\linewidth}{%
  \textbf{Retrosynthesis.} 
      You are an expert in organic chemistry. The following are multiple choice questions about Retrosynthesis. Let's think step by step. Please wrap the final answer in XML tags: <answer>X</answer> (where X is A, B, C, or D) \\
      Question:\\
      For the retrosynthetic analysis of Nc1cc(Cl)c(Oc2ccc(Cl)c3ccccc23)c(Cl)c1, which transformation would be the poorly strategic choice? \\
     A. ([NH2;D1;+0:1]-[c:2])>>(C-C(-C)(-C)-O-C(=O)-[NH;D2;+0:1]-[c:2]) \\
     B. ([NH2;D1;+0:1]-[c:2]1:[c:3]:[c:4]:[c:5](-[\#8:6]-[c:7]):[c:8]:[c:9]:1)>>(O=[N+;H0;D3:1](-[O-])-[c:2]1:[c:3]:[c:4]:[c\\:5](-[\#8:6]-[c:7]):[c:8]:[c:9]:1)\\
     C. ([NH2;D1;+0:1]-[c:2])>>(C-C(=O)-[NH;D2;+0:1]-[c:2])\\
     D. ([C:2]-[CH2;D2;+0:1]-[NH;D2;+0:4]-[C:3])>>(C-S(=O)(=O)-O-[CH2;D2;+0:1]-[C:2]).([C:3]-[NH2;D1;+0:4])\\
     Answer:
  }
}
\noindent\fbox{%
  \parbox{\linewidth}{%
  \textbf{Molecular property estimation.} You are an expert in chemistry. The following are multiple choice questions about Molecular Property. Let's think step by step. Please wrap the final answer in XML tags: <answer>X</answer> (where X is A, B, C, or D) \\
  Question:\\
  Based on number of rotatable bonds, which arrangement represents these molecules in ascending order? \\
1. CC1CC(C(=O)O1)NC(=O)C2=CC=CC=C2 \\
2. CCOC(=O)C1=NOC(=C1C(=O)OC)C \\
3. CCC1=CC2=C(S1)C(=O)N3CCOC3=N2 \\
4. methyl 1,3-dimethyl-7,8,9,10,11,12-hexahydrocycloocta[a]indolizine-6-carboxylate \\
A. Molecule 2 < Molecule 1 < Molecule 3 = Molecule 4 \\
B. Molecule 1 < Molecule 2 < Molecule 4 = Molecule 3 \\
C. Molecule 1 < Molecule 4 < Molecule 2 < Molecule 3 \\
D. Molecule 3 = Molecule 4 < Molecule 1 < Molecule 2 \\
Answer:
  }
}
\noindent\fbox{%
  \parbox{\linewidth}{%
  \textbf{Experimental techniques.} You are an expert in experimental chemistry. The following are multiple choice questions about Laboratory Techniques. Let's think step by step. Please wrap the final answer in XML tags: <answer>X</answer> (where X is A, B, C, or D) \\
  Question:\\
  During a methylation reaction using MeI (1.2 equiv.) under ice bath conditions (~0°C), you observe poor regioselectivity between competing nucleophilic sites. Which approach would most likely improve the selectivity? \\
A. Reduce MeI to 1 equivalents to minimize over-alkylation \\
B. Remove the ice bath and run at room temperature \\
C. Use an ice-salt bath to reach -15°C \\
D. Switch to a dry ice-acetone bath (-78°C) \\
Answer:
  }
}
\noindent\fbox{%
  \parbox{\linewidth}{%
  \textbf{NMR-based structure elucidation.} You are a chemist assistant with expertise in molecular structure elucidation. Given the following spectroscopic data for an unknown compound: Molecular Formula: C13H12OSe. 1H NMR (CDCl3, 400 MHz): $\delta$ 7.50 (d, J = 8.8 Hz, 2H); 7.33-7.31 (m, 2H); 7.21-7.16 (m, 3H); 6.84 (d, J = 8.4 Hz, 2H); 3.79 (s, 3H). 13C NMR (CDCl3 100 MHz); $\delta$ (ppm): 159.7, 136.5, 133.2, 130.9, 129.1, 126.4, 119.9, 115.1, 55.2. MS (relative intensity) m/z: 264 (65), 262 (34), 184 (100), 153 (32), 65 (14). Determine the complete molecular structure.  \\
Requirements:  \\
    - Provide the structure in SMILES notation  \\
    - The answer should be exact and canonical  \\
    - Include only the SMILES string in your answer, wrapped in <SMILES></SMILES> tags  \\
  Example answer format:  \\
  <SMILES>CCN(CC)CC</SMILES>  \\
  Answer: \\
  }
}
\subsection{Biology Domain}
\noindent\fbox{%
  \parbox{\linewidth}{%
      \textbf{Lipinski rule assessment.} You are an expert biochemist specializing in drug-like property analysis. Your task is to analyze a molecule's SMILES representation and determine whether it satisfies Lipinski's Rule of Five.\\
      Given Information: \\
      - Input SMILES: CC(C)C[C@@H]1NC(=O)[C@@H](CC(C)C)NC(=O)[C@@H](C)NC(=O)[C@@H]2CCCN2C(=O\\
      )[C@H](Cc2ccccc2)NC(=O)[C@@H](CC(C)C)N(C)C(=O)[C@@H]2CCCN2C1=O \\
      Lipinski's Rule of Five Criteria: \\
      A molecule satisfies the rule if ALL of the following conditions are met: \\
      - Hydrogen Bond Donors (HBD) $\leq$ 5 \\
      - Hydrogen Bond Acceptors (HBA) $\leq$ 10 \\
      - Molecular Weight (MW) < 500 Da \\
      - LogP < 5\\
      Task: Calculate the four properties and determine if the molecule satisfies all criteria.\\
      Options: \\
      A: Satisfies Lipinski's Rule of Five (meets all criteria) \\
      B: Violates Lipinski's Rule of Five (fails one or more criteria)\\
      Requirements: \\
      - Calculate all four molecular properties from the SMILES \\
      - Check if ALL criteria are satisfied - Provide only the letter of your answer \\
      - Do not include explanations or reasoning \\
      - Wrap your answer in <ANSWER></ANSWER> tags\\
      Output Format: <ANSWER>[Single letter: A or B]</ANSWER>
  }
}
\noindent\fbox{%
  \parbox{\linewidth}{%
  \textbf{GWAS causal gene identification.} You are an expert in genetics and GWAS analysis. Your task is to identify the most likely causal gene within a genomic locus for a given GWAS phenotype. Given Information:\\
  GWAS Phenotype: Multi-trait sex score Genes in Locus: APOC1, APOC2, APOC4, APOC4-APOC2, APOE, BCAM, BCL3, BLOC1S3, CBLC, CEACAM16, CEACAM19, CEACAM20, CKM, CLASRP, CLPTM1, ENSG00000267173, ERCC2, EXOC3L2, GEMIN7, IGSF23, KLC3, MARK4, NECTIN2, NKPD1, PPP1R13L, PPP1R37, PVR, RELB, TOMM40, TRAPPC6A, ZNF180, ZNF229, ZNF285, ZNF296\\
  Requirements:\\
  Analyze the biological relationship between the phenotype and each gene in the locus Consider known gene functions, pathways, and disease associations Select the gene most likely to be causally related to the phenotype Choose only from the genes provided in the list Provide only the gene symbol as your answer\\
  Output Format: <ANSWER>[gene symbol from the provided list]</ANSWER>
  }
}
\noindent\fbox{%
  \parbox{\linewidth}{%
  \textbf{CRISPR delivery strategy selection.} You are an expert in CRISPR gene editing technologies. Your task is to identify the two most relevant CRISPR delivery methods based on the user's requirements.\\
  Given Information: \\
  - User Requirements: I hope to edit primary T-cell\\
  Available Delivery Methods: \\
  a. Plasmid Transfection \\
  b. Lentivirus/Retrovirus \\
  c. RNP/mRNA electroporation \\
  d. RNP/mRNA microinjection \\
  e. mRNA LNP \\
  f. AAV\\
  Requirements: \\
  - Analyze the user's needs and experimental context \\
  - Select the most appropriate delivery method as Answer1 \\
  - Select the second most appropriate delivery method as Answer2 \\
  - Provide only single letters (a-f) as answers \\
  - Wrap your answer in <ANSWER></ANSWER> tags\\
  Output Format: <ANSWER>[Answer 1: single letter: a, b, c, d, e, or f], [Answer 2: single letter: a, b, c, d, e, or f]</ANSWER>
  }
}

\subsection{Physics domain}
\noindent\fbox{%
  \parbox{\linewidth}{%
  \textbf{Astrophysics and cosmology.} You are an expert in Astrophysics and Cosmology. The following are multiple choice questions about astrophysics and cosmology. Let's think step by step. Please wrap the final answer in XML tags: <answer>X</answer> (where X can be A, B, C, D, or combinations like AB, ABC, etc.) \\
  Question: \\
  If we are living in a universe with primordial black holes as the only dark matter candidate (i.e. $\Omega_{DM} = \Omega_{PBH}$), and all of them are centered around some mass $M_{PBH}$, with number density given by $n_{PBH}$. What is the dominant change (Poisson fluctuation) induced by PBHs to the linear power spectrum $P(k)$ related to the number density of PBHs $n_{PBH}$ (only show the proportional relationship)? Derive this proportionality from the cosmic over-density field \\
  Answer:
  }
}
\noindent\fbox{%
  \parbox{\linewidth}{%
  \textbf{Quantum information science.} You are an expert in Quantum Information Theory. Please solve the following question step by step.

\paragraph{Question.}
Consider a tripartite state $\boxed{\rho_{ABC}}$ consisting of $d$-dimensional qudits.
Let $A$, $B$, $C$ contain $N_A$, $N_B$, and $N_C$ qubits, respectively.
Suppose there exist a channel $R$ acting on the Hilbert space and outputing to the Hilbert space of $AB$ such that
\[
\left\| R[\rho_{BC}] - \rho_{ABC} \right\|_1 = \epsilon .
\]
One can show that the conditional mutual information
\[
\boxed{I(A:C \mid B) \le c \sqrt{\epsilon}} .
\]
How does the prefactor $c$ depend on $d$, $N_A$, $N_B$, and $N_C$?
You can throw away all the constants. Provide your reasoning first, then conclude with your final answer.\\
IMPORTANT: You MUST put your final mathematical answer in this exact format:
\[
\text{Final Answer: } \boxed{\text{your mathematical expression}}
\]
Do not use any other format for your final answer. \\
  Answer:
  }
}
\noindent\fbox{%
  \parbox{\linewidth}{%
  \textbf{Probability and statistics.} You are an expert in Probability and Statistics. The following are multiple choice questions about probability and statistics. Let's think step by step. Please wrap the final answer in XML tags: <answer>X</answer> (where X can be A, B, C, D, or combinations like AB, ABC, etc.) Question: \\
  Which of the concentration inequalities below are based on moment generating function? \\
  A: Chebyshev \\
  B: Chernoff \\
  C: Azuma-Hoeffding  \\
  D: Bernstein \\
  Answer:
  }
}

\subsection{Materials Domain}
\noindent\fbox{%
  \parbox{\linewidth}{%
  \textbf{MOF synthesis Q\&A.}
  You are an expert in MOF synthesis. Select the correct answer from the choices below.\\
  
  Wrap your final answer in <answer> tags, do not include any additional text or explanations in your response.
  Only answer with the letter corresponding to the correct option, for example: <answer>A</answer>\\
  
  A research team is synthesizing an aluminum-based MOF using a 4,4',4''-Benzene-1,3,5-triyl-tris-(benzoate) linker, denoted as BTB, and explores different parameters to maximize both yield and crystal size. They find that at temperatures below 120°C, yield is < 15\%, while at 140°C, yield is 80\%. A BTB:Al ratio of 3:4 increases yield by 10\% over 2:1. Using a modulator of formic acid:water at 1:1 doubles the crystal size without changing yield, while a 4:1 ratio decreases crystal size by 30\%. Reaction times > 72h do not affect yield or size. Which of the following recipes maximizes both yield and crystal size?\\

 A. 2:1 BTB:Al, pure formic acid, 140°C, 72h \\
 B. 3:4 BTB:Al, formic acid:water 1:1, 140°C, 72h\\
 C. 3:4 BTB:Al, formic acid:water 4:1, 140°C, 72h\\
 D. 4:1 BTB:Al, pure formic acid, 130°C, 72h\\
 E. 3:4 BTB:Al, pure formic acid, 120°C, 96h\\
 Answer:
  }
}
\noindent\fbox{%
  \parbox{\linewidth}{%
  \textbf{Crystal system prediction from PXRD} 
  You are an expert crystallographer. Your task is to classify the crystal system of the following material based on its PXRD data.\\

  Material type: Ionic Compound\\
  Peak positions: 25.235, 41.792, 49.462, 60.609, 66.758, 76.329, 76.393\\
  Peak intensities: 100.0, 48.161, 31.473, 15.599, 6.261, 10.34, 7.737\\

Answer:
  }
}
\noindent\fbox{%
  \parbox{\linewidth}{%
  \textbf{Metal-organic framework water stability text-mining} 
    You are an expert in metal-organic frameworks. You will be given a passage of text from a paper, and must use it to decide the MOF's water stability.
  Note, the passage may not contain enough information, or contain relevant information, needed to answer the question. Wrap the final answer in <answer>X</answer> tags (X = A, B, or C).\\

 Based off this passage, identify if the MOF called "MOF-805" is stable or unstable in water. Note, there may not be enough information to provide an answer:\\
 
 "The cycle performance results show that, for MOF-805, MOF-806, MOF-808, and Basolite A100, A300, and C300, the uptake constantly drops in every cycle. The surface area of these MOFs was redetermined after the water cycle tests, showing a significant decrease. This observation suggests that the loss of water uptake capacity is related to the loss of porosity."\\

  A: Stable\\
  B: Unstable\\
  C: Not enough information\\
  Answer:

  }
}
\section{Detailed discussion on research project experiments}
\label{Supp:projects_section}
\subsection{Retrosynthesis pathway design}
The workflow follows an \texttt{Initialization} and \texttt{Mutation} phase which we detail below. The original implementation is from Wang et al.\cite{wang2025llm}

\textbf{Initialization.} Each experiment began by computing the Tanimoto similarity between the target molecule to a list of reference molecules using Morgan fingerprints with radius 2. These reference molecules are from the training and validation sets of Chen et al.\cite{chen2020retro}, which in turn are extracted from the United States Patent and Trademark Office (USPTO)\cite{lowe_2012}. Three reference synthesis routes are retrieved with probability proportional to the Tanimoto similarity. The \texttt{Initialization} phase tasks the LLM with proposing an initial synthesis route to the target molecule, given the reference routes as context. The proposed routes are assessed based on \textit{validity} which is comprised of the following: 

\begin{enumerate}
    \item \textbf{Molecule validity}: all molecules are RDKit parsable and the final precursors are commercially available (here, the eMolecules commercially available building blocks stock from Chen et al.\cite{chen2020retro} is used).
    \item \textbf{Reaction validity}: all LLM-proposed reaction templates have an exact match in the reaction database (here, the reaction database is derived from \texttt{USPTO-Full} extracted from Yu et al.\cite{yu2024double}).
    \item \textbf{Route validity}: The LLM proposes a route following the desired data format. If the first step of the proposed route fails, the LLM is prompted to propose another route.
\end{enumerate}

In the case that an LLM-proposed reaction template does not exist in the reaction database, a difference fingerprint (RDKit's CreateDifferenceFingerprintForReaction) is created for the proposed reaction and similar reactions are retrieved from the reference USPTO database. These retrieved reactions are applied and the first one that results in valid reactants is selected. If no valid reaction is found, then a reaction template is selected based on Tanimoto similarity to the target molecules present in the USPTO database. Following the original work, 10 total initial valid routes are generated.

\textbf{Mutation.} The \texttt{Mutation} phase refines the population of initial routes. One parent route is selected and the non-purchasable intermediate molecules are identified (e.g. molecules that are a result of applying a reaction template but are not commercially available so must be decomposed further). Using the same Tanimoto similarity comparison to the USPTO reference database, routes of similar targets are retrieved. The LLM is then tasked to propose a modified route given these reference routes and feedback on current problems in the parent route (which is one of the initial routes generated in the initialization phase). If the LLM successfully proposes valid steps in the synthesis route, these steps are appended to the parent route and the full route (so far). This "full" route may still have problems which can then be iterated on in the next mutation trial. All routes in the population are scored with a combination of the synthetic complexity (SC)\cite{coley2018scscore} and synthetic accessibility (SA)\cite{ertl2009estimation} scores for the non-purchasable intermediates. The 10-best scoring routes are kept before beginning a new mutation round. The entire search process terminates when either a valid synthesis pathway is found or the LLM budget is exhausted (in this work, we allow 100 LLM queries), whichever occurs first. The evaluation runs the LLM framework on prescribed sets of target molecules and reports the number of molecules that return a valid synthesis route, within 100 LLM queries (i.e. the solve rate).

\textbf{Results.} Using the \texttt{Pistachio Hard}\cite{pistachio} benchmark set of 100 molecule targets, we evaluated the solve rate using multiple LLMs and compared to common methods spanning MCTS\cite{segler2018planning} and Retro*\cite{chen2020retro} search \Cref{tab:llm_synplanner_results}. Overall, the LLMs' performance is competitive with baselines with an explicit search algorithm. The results show the feasibility of using the LLM itself for retrosynthetic planning, replacing the search algorithm. We note however, that the results were only run for one replicate due to computational cost (newer LLMs are notably more expensive than its predecessors). Interestingly, GPT-4o outperformed all newer models which empirically struggled more to generate valid routes. Specific failure modes include outputting routes that do not conform to the desired route data format and/or violating the molecule or reaction validity checks.

\begin{table}[!h]
\centering
\caption{Retrosynthesis solve rate comparison on \texttt{Pistachio Hard} (100 targets). Following the original work, \texttt{temperature = 0.7}, if permitted. For \texttt{gpt-5} and \texttt{gpt-5-chat}, only \texttt{temperature = 1.0} is permitted. For \texttt{gpt-4o} and \texttt{gpt-5-chat}, \texttt{max\_tokens = 16384}. Otherwise, \texttt{max\_tokens = 32768}}
\label{tab:llm_synplanner_results}
\small
\vspace{0.5em}
\begin{tabular}{lr}
\hline
\textbf{Method} & \textbf{Pistachio Hard Solve Rate (\%)} \\
\hline
Graph2Edits\cite{graph2edits} (MCTS) & 26.0 \\
RootAligned\cite{rootaligned} (MCTS) & 83.0 \\
LocalRetro\cite{localretro} (MCTS) & 52.0 \\
Graph2Edits (Retro*) & 71.0 \\
RootAligned (Retro*) & 78.0 \\
LocalRetro (Retro*) & 63.0 \\
\hline
\texttt{gpt-4o} & 60.0 \\
\texttt{gpt-5-chat} & 49.0 \\
\texttt{gpt-5} & 53.0 \\
\texttt{claude-sonnet-4.5} & 53.0 \\
\texttt{deepseek-R1} & 42.0 \\
\hline
\end{tabular}
\end{table}

\clearpage
\textbf{gpt-4o Outperforms Newer LLMs.} Table \ref{tab:llm_synplanner_results} shows that \texttt{gpt-4o} consistently outperforms newer LLMs, particularly its successors, \texttt{gpt-5} and \texttt{gpt-5-chat}, on the \texttt{Pistachio Hard} set of targets. We investigate the failure modes of all compared LLMs by analyzing the frequency in which an LLM-proposed reaction step encounters specific errors (Table \ref{tab:llm_synplanner_failure_analysis}). We make the following observations for the most apparent differences amongst the LLMs: (1) \texttt{gpt-5} and \texttt{claude-sonnet-4-5} propose routes that have more steps, before either solving the molecule or encountering an error (\textit{Steps Per Route Attempt}). (2) A notable failure mode for \texttt{gpt-5}, \texttt{gpt-5-chat}, and \texttt{deepseek-reasoner} is generating invalid SMILES. Note that this does not necessarily mean that these models are worse than \texttt{gpt-4o} and \texttt{claude-sonnet-4-5} at generating valid SMILES, but rather, worse at generating valid SMILES in the context of the synthesis route data structure. The two types of invalid SMILES errors occur in the \textit{Updated Set}: After applying a reaction template, the LLM adds invalid reactant SMILES to the molecule set and in the \textit{Mol Set}: The LLM tries to decompose a SMILES that is invalid (this could have been inherited from the previous step). (3) Reasoning can improve SMILES validity in the route context. The most relevant comparison is \texttt{gpt-5} and \texttt{gpt-5-chat} where the former is a reasoning model and has notably lower invalid SMILES rates. We additionally highlight that \texttt{claude-sonnet-4-5} has a particularly low invalid SMILES rate, but has a notably higher \textit{Updated Set Mismatch} error. This error means that after applying a reaction template, the LLM incorrectly generated the \textit{updated molecule set}, which should equal the original set plus the new reactants, minus the decomposed product.

\begin{table}[!h]
\centering
\caption{LLM failure modes on \texttt{Pistachio Hard} (100 targets).}
\label{tab:llm_synplanner_failure_analysis}
\small
\vspace{0.5em}
\scalebox{0.98}{
\begin{tabular}{lrrrrr}
\hline
\textbf{Metric} & \texttt{gpt-4o} & \texttt{gpt-5} & \texttt{gpt-5-chat} & \texttt{claude-sonnet-4-5} & \texttt{deepseek-reasoner} \\
\hline
\multicolumn{6}{l}{\textbf{Absolute Count Metrics}} \\
\hline
Total Solved Routes & 60 & 53 & 49 & 53 & 42 \\
Total Route Attempts & 4630 & 4701 & 3729 & 3392 & 4824 \\
Total Steps & 12674 & 17283 & 10815 & 11825 & 15343 \\
Steps Per Route Attempt & 2.74 & 3.68 & 2.90 & 3.49 & 3.18 \\
\\[-0.6em]
\multicolumn{6}{l}{\textbf{Percentage (\%) of Total Steps Metrics}} \\
\hline
Valid Steps & 36.77 & 28.78 & 34.83 & 31.97 & 32.46 \\
\\[-0.6em]
\multicolumn{6}{l}{\textbf{Invalid SMILES Errors}} \\
In the Updated Set & 16.47 & 21.53 & 41.85 & 4.96 & 24.27 \\
In the Mol Set& 6.75 & 13.98 & 19.57 & 3.32 & 13.22 \\
\\[-0.6em]
\multicolumn{6}{l}{\textbf{Invalid Reaction Errors}} \\
Reaction not in Database & 54.23 & 58.99 & 59.57 & 48.68 & 56.16 \\
Reaction Exists but Template Fails & 0.43 & 0.55 & 0.42 & 0.82 & 0.68 \\
\\[-0.6em]
\multicolumn{6}{l}{\textbf{Set Errors}} \\
Updated Set Mismatch & 8.57 & 11.68 & 5.18 & 18.53 & 10.70 \\
Selected Mol to Decompose not in Set & 36.63 & 28.06 & 31.79 & 31.09 & 30.52 \\
\hline
\end{tabular}
}
\end{table}

\clearpage
\subsection{Molecule optimization}
We evaluated the molecule optimization task targeting two objectives, \textit{jnk3} and \textit{gsk3$\beta$}. Each experiment began with an initial population of $120$ molecules sampled from the ZINC dataset\cite{zinc2012}. In each generation, two parent molecules were drawn from the current population with probability proportional to their fitness. For LLM-based methods, mutation and crossover were implemented by prompting the model with one or two parent molecules and asking it to propose a new molecule, either by mutating a single parent or recombining both. This procedure was repeated until $70$ offspring were generated. All offspring were evaluated by the oracle, and the union of parents and offspring was re-ranked by fitness; the top-$120$ molecules were retained as the population for the next generation. We capped the total number of oracle calls at $10{,}000$ and applied early stopping: if the mean fitness of the top-$100$ molecules failed to improve by at least $10^{-3}$ over $5$ consecutive generations, the run was terminated. Methods were compared using the area under the curve of the top-$k$ average objective versus the number of oracle calls (AUC$_{\text{top-}k}$) with $k = 10$, which jointly captures optimization quality and sample efficiency.

The quantitative results for both objectives are reported in Table~\ref{tab:molleo_results}. For non-LLM baselines, Graph GA is consistently weaker than learning-based approaches, especially on \textit{jnk3}, while REINVENT provides a strong and stable reference across tasks. This highlights the advantage of specialized molecular generative frameworks for property-driven optimization.

Among LLM-based methods, GPT-4o performs competitively on both objectives, matching REINVENT on \textit{jnk3} and remaining close on \textit{gsk3$\beta$}. Claude Sonnet 4.5 achieves the best overall performance across both tasks, with the highest AUC Top 10 on \textit{jnk3} and a particularly strong result on \textit{gsk3$\beta$}, suggesting that both optimization efficiency and final top-$k$ quality benefit from strong generative priors and effective exploration. DeepSeek-R1 remains competitive but shows a larger gap on \textit{gsk3$\beta$} in terms of Avg Top 10, indicating that maintaining high-quality top candidates may be more sensitive to model-specific generation behavior for this objective.

In contrast, GPT-5 and GPT-5-chat underperform on \textit{jnk3}, showing noticeable drops in both AUC Top 10 and Avg Top 10. On \textit{gsk3$\beta$}, they improve substantially (e.g., GPT-5 Avg Top 10 $0.942$), but still lag behind the best-performing models in sample efficiency. Consistent with our empirical observations, GPT-5 tends to propose molecules with a higher duplication rate, which reduces effective exploration of chemical space and can disproportionately hurt AUC-based metrics, even when the final top-$k$ set is reasonably strong. This comparison is also affected by an unavoidable decoding mismatch: GPT-5 and GPT-5-chat can only be run with \texttt{temperature = 1.0}, whereas all other LLMs are evaluated with \texttt{temperature = 0.8}. As a result, the observed gaps may reflect a combination of model behavior and sampling differences, rather than the reasoning-oriented design of GPT-5 alone.

\begin{table}[h]
\centering
\caption{Performance comparison of different models on molecule optimization tasks (jnk3, gsk3$\beta$, sitagliptin\_mpo). All LLM models were tested with temperature = 0.8 and max\_tokens = 8192 except for GPT-5 and GPT-5-chat.}
\label{tab:molleo_results}
\begin{tabular}{lrrrrrr}
\midrule
\textbf{Method} & \multicolumn{2}{c}{\textbf{jnk3}} & \multicolumn{2}{c}{\textbf{gsk3$\beta$}} & \multicolumn{2}{c}{\textbf{sitagliptin\_mpo}} \\
\cmidrule(lr){2-3}\cmidrule(lr){4-5}\cmidrule(lr){6-7}
 & \textbf{AUC Top 10} & \textbf{Avg Top 10} & \textbf{AUC Top 10} & \textbf{Avg Top 10} & \textbf{AUC Top 10} & \textbf{Avg Top 10} \\
\midrule
Graph GA & 0.548 & 0.890 & 0.779 & 0.945 & 0.425 & 0.475 \\
REINVENT & 0.794 & 0.912 & 0.868 & 0.978 & 0.434 & 0.508 \\
\midrule
\texttt{gpt-4o} & 0.796 & 0.932 & 0.857 & 0.972 & 0.572 & 0.633 \\
\texttt{gpt-5-chat} & 0.717 & 0.803 & 0.832 & 0.887 & 0.453 & 0.478 \\
\texttt{gpt-5} & 0.695 & 0.752 & 0.822 & 0.942 & 0.553 & 0.765 \\
\texttt{claude-sonnet-4.5} & 0.865 & 0.935 & 0.981 & 1.00 & 0.422 & 0.456 \\
\texttt{deepseek-R1} & 0.749 & 0.932 & 0.849 & 0.893 & 0.575 & 0.608 \\
\midrule
\end{tabular}
\end{table}

\clearpage
\subsection{Transition metal complex optimization}
\label{Supp:tmc_section}
In each iteration, a prompt is meticulously crafted, comprising generic instructions alongside specific information, constraints, and objectives, and is then presented to an LLM to be evaluated. 
The reference data as the input into the model through carefully constructed prompts containing: (1) a pool of 50 ligands represented by their SMILES strings, IDs, charges, and connecting atom information; (2) 20 randomly sampled initial TMCs from a space of 1.37M possible Pd(II) square planar complexes, along with their pre-calculated properties (HOMO-LUMO gap and polarisability); and (3) natural language descriptions of the design objectives (e.g., maximizing HOMO-LUMO gap). 
The LLM subsequently proposes a new set of TMCs, which undergo a rigorous validation process including charge constraints (-1, 0, or +1), structure generation using molSimplify\cite{molsimplify}, geometry optimization with GFN2-xTB, and connectivity validation to ensure no unintended bond rearrangements occur. 
Valid TMCs and their calculated properties are integrated into the prompt for the subsequent iteration, effectively completing the scientific discovery loop. 
Specifically, we start with 20 initial TMCs, with the LLM proposing 10 new TMCs at each iteration until a maximum iteration of 20 is reached. 
All TMCs explored during the optimization process, regardless of their fitness values, are included in the prompt for the next iteration. 
This mimics the human learning experience where one can learn from both good and bad examples. 
To prevent the LLM from overemphasizing specific records, the historical TMC data within the prompt is randomly shuffled before each iteration.
To minimize bias from the initial TMC sampling, five random seeds are used to sample the initial known TMCs.

\begin{figure*}[!htbp]
    \centering
    \includegraphics[width=0.98\textwidth]{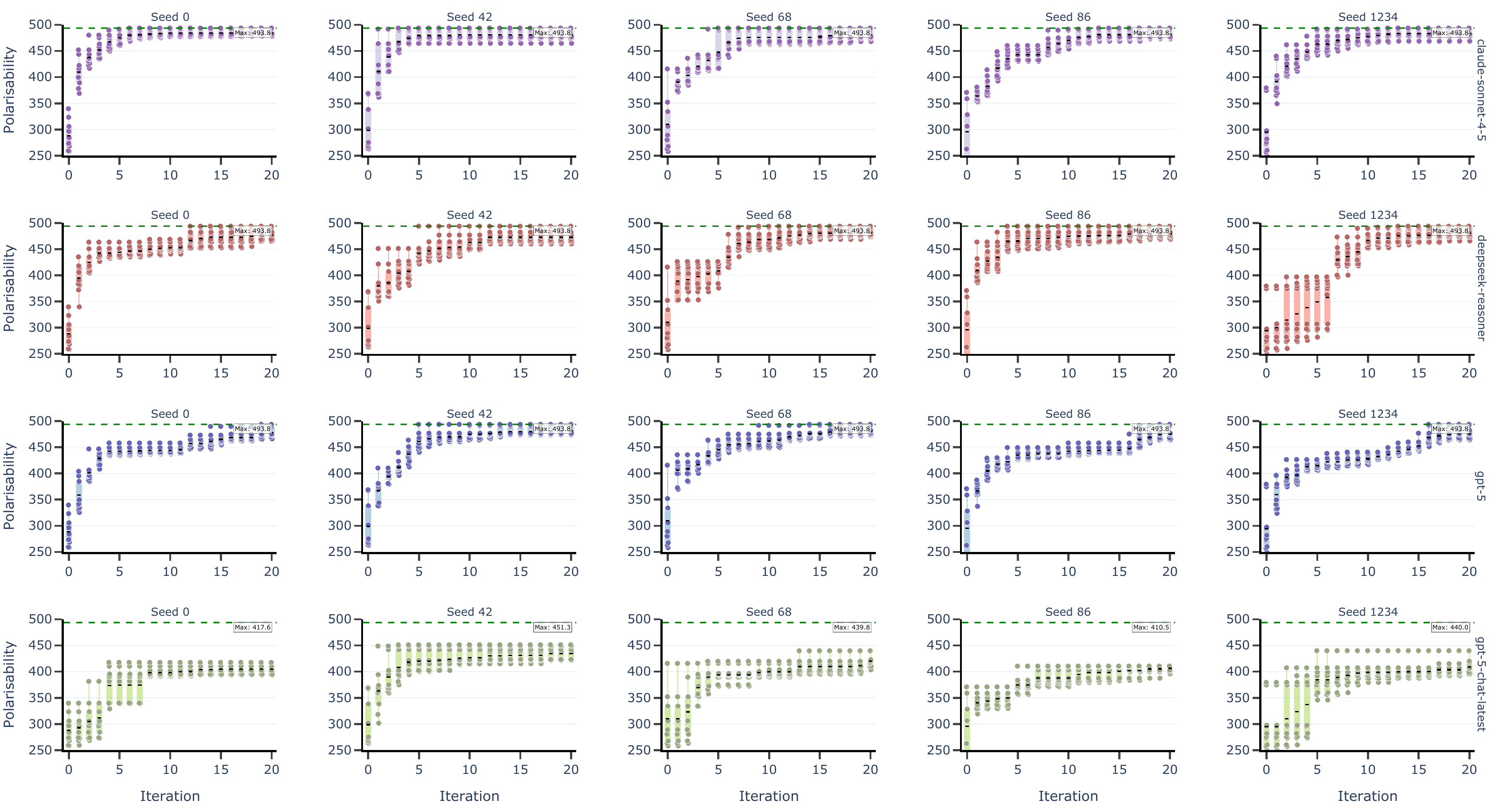}
    \caption{
    \textbf{Distribution of top-10 unique TMCs by iterations on maximizing polarisability.} Each model is evaluated on five different random seeds  and shown in different columns. Results from different models are displayed at different rows. \texttt{gpt-5-chat-latest} is colored in green, \texttt{gpt-5} in blue, \texttt{deepseek-R1} in red, and \texttt{claude-sonnet-4.5} in purple.}
    \label{Supp:tmc_sp}
\end{figure*}

\qquad In the first task of proposing TMCs with maximized polarisability, \texttt{gpt-5}, \texttt{deepseek-R1}, and \texttt{claude-sonnet-4.5} successfully finds the optimal solution in the space of 1.37M TMCs at all five random seeds (Fig. \ref{Supp:tmc_sp}).
On the contrary, \texttt{gpt-5-chat-latest} fails to do so in all five random seeds, showing the significance of reasoning capability in TMC optimization.
Across three reasoning models, \texttt{claude-sonnet-4.5} demonstrates quicker convergence through iteration compared to \texttt{gpt-5} and \texttt{deepseek-R1} consistently.
A similar trend is observed when models are asked to expand the Pareto frontiers by proposing TMCs (Fig. \ref{Supp:tmc_pf}).
There, \texttt{gpt-5-chat-latest} still finds the most limited Pareto frontiers, hardly identifying TMCs with polarsability > 400 a.u. and HOMO-LUMO gap > 4 eV.
Out of the three reasoning models, \texttt{deepseek-R1} gives the most expanded and balanced Pareto frontiers.
It should be noted that the influence of random seed is significant, which leads to different sets of TMCs found as Pareto frontiers.
\texttt{claude-sonnet-4.5} is impacted the most by random seeds, and only explore TMCs locally at seed 68, 86, and 1234.
\texttt{gpt-5}, despite not yielding the best combined Pareto frontiers across all models, follows the instruction clearly and attempts a balanced exploration at all random seeds.

\begin{figure*}[!htbp]
    \centering
    \includegraphics[width=0.98\textwidth]{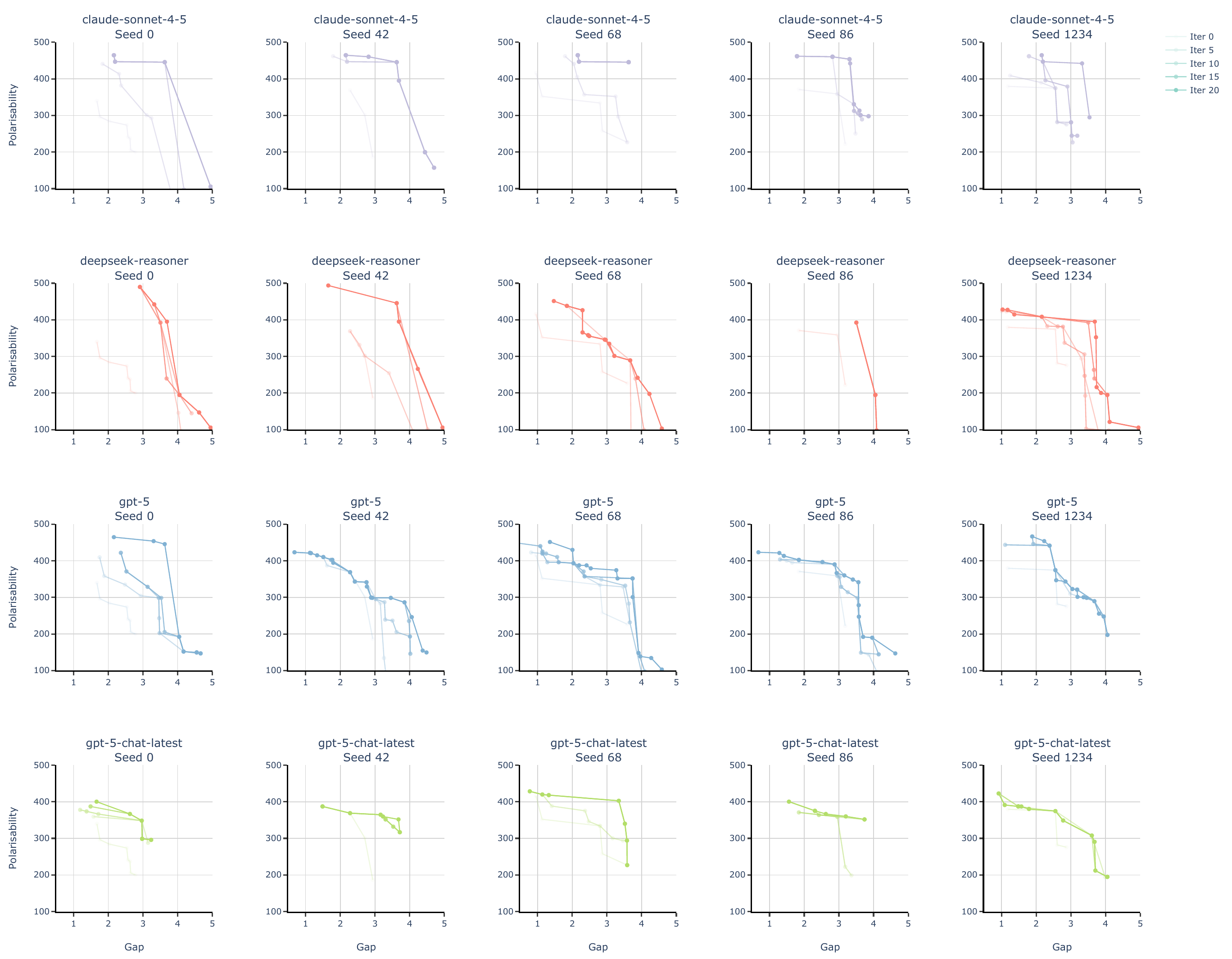}
    \caption{
    \textbf{Pareto frontiers of proposed TMCs by iterations.} Each model is evaluated on five different random seeds and shown in different columns. Results from different models are displayed at different rows. \texttt{gpt-5-chat-latest} is colored in green, \texttt{gpt-5} in blue, \texttt{deepseek-R1} in red, and \texttt{claude-sonnet-4.5} in purple, with reduced transparency as iteration number increases.}
    \label{Supp:tmc_pf}
\end{figure*}

\clearpage
\subsection{Crystal structure discovery}
Each experiment began with an initial population of $100$ groups of parents ($100 \times 2 = 200$ parent structures), randomly seeded from the reference pool, which is composed with 5,000 known stable structures from MatBench-bandgapt~\cite{dunn2020benchmarking} dataset with lowest deformation energy evaluated by CHGNet~\cite{deng2023chgnet}. The mutation and crossover operations for LLMs were implemented by prompting the LLMs with two sampled parent structures based on their fitness values (minimizing $E_\text{d}$) and querying them to propose $5$ new structures either through mutation of one structure or crossover of both structures. After generating new offspring in each generation, we evaluated the new offspring and merged their evaluations with the parent evaluations from the previous iteration. The merged pool of parents and children were then ranked by their fitness values (minimizing $E_\text{d}$), and the top-$100 \times 2$ candidates were kept in the population as the pool for the next iteration. We evaluate generated structures through metrics that assess validity, diversity, novelty, and stability. Structural validity checks three-dimensional periodicity, positive lattice volume, and valid atomic positions. Composition validity verifies positive element counts and reasonable number of elements ($\leq 10$). Structural diversity is computed by deduplicating the generated set using pymatgen's StructureMatcher algorithm, then calculating the ratio of unique structures to total generated. Composition diversity measures the fraction of distinct chemical compositions. For novelty assessment, we compare generated structures against the initial reference pool. Composition novelty identifies structures whose reduced formulas are absent from the reference set. Structural novelty is determined by grouping reference structures by formula, then for each generated structure with a matching formula, using StructureMatcher to check if it matches any reference structure with the same composition; unmatched structures are considered structurally novel. Stability evaluation uses CHGNet to relax structures and compute formation energy, then calculates energy above the convex hull ($E_\text{d}$) via a pre-computed patched phase diagram database. We report metastability rates at three thresholds: $E_\text{d} < 0.0$ eV/atom (thermodynamically stable), $E_\text{d} < 0.03$ eV/atom (highly metastable), and $E_\text{d} < 0.10$ eV/atom (M3GNet metastability criterion). The integrated SUN (Structures Unique and Novel) score combines stability and novelty: (1) filter to structures with $E_\text{d} < 0.0$ eV/atom; (2) identify unique structures within this stable subset using pymatgen's Structure.matches with scaling enabled; (3) check novelty against the reference pool; (4) compute SUN score as the number of structures simultaneously stable, unique, and novel, divided by the total number of generated structures.

We evaluated multiple LLMs on the crystal structure generation task, comparing their performance against established baseline methods CDVAE~\cite{xie2022crystal} and DiffCSP~\cite{jiao2024crystal}. \Cref{tab:matllm_results} summarizes the results across key metrics when considering both parent and child structures to form the next generation. Overall, LLM-based search achieves superior validity and metastability rates compared to traditional generative models CDVAE and DiffCSP. S.U.N. rate for LLM-based search is evaluated against the entire MatBench-bandgap dataset. GPT-5 achieved the highest overall performance in generating novel structures that are both compositionally novel and thermodynamically metastable. DeepSeek Reasoner and Grok-4 showed competitive metastability rates (88.90\% and 87.13\% for $E_d < 0.1$ eV/atom, respectively). Comparison among the GPT-5 family suggests that model scale and training objectives significantly impact crystal structure generation quality. Claude Sonnet 4.5 achieves comparable metastability rates to GPT-5-mini but a lower S.U.N. rate (38.99\%), indicating less exploration and more exploitation in the search space. In addition, reasoning models like DeepSeek Reasoner require more tokens for internal reasoning chains. With the 8,000 token limit used in these experiments, such models might overflow the context window when generating complex crystal structures with detailed chemical reasoning. Experiments with extended context windows can further improve reasoning model performance on this task.

\begin{table}[h]
\centering
\caption{Performance comparison of different models on MatLLMSearch crystal structure generation task. All LLM models were tested with \texttt{temperature: 1.0} and \texttt{max\_tokens: 8000}.}
\label{tab:matllm_results}
\scriptsize
\begin{tabular}{lrrrrr}
\hline
\textbf{Method} & \textbf{Structural} & \textbf{Comp} & \textbf{Metastability} & \textbf{Metastability} & \textbf{S.U.N. Rate} \\
 & \textbf{Validity (\%)} & \textbf{Validity (\%)} & \textbf{($E_d < 0.1$ eV/atom, \%)} & \textbf{($E_d < 0.0$ eV/atom, \%)} & \textbf{(\%)} \\
\hline
CDVAE & 100 & 86.70 & 28.8 & -- & -- \\
DiffCSP & 100 & 83.25 & -- & 5.06 & 3.34 \\
\hline
\texttt{gpt-5-mini} & 100 & 100 & 74.60 & 50.05 & 46.24 \\
\texttt{gpt-5-chat} & 100 & 100 & 64.36 & 46.93 & 44.37 \\
\texttt{gpt-5} & 100 & 100 & 88.33 & 63.22 & 55.31 \\
\texttt{claude-sonnet-4.5} & 100 & 100 & 78.71 & 50.21 & 38.99 \\
\texttt{deepseek-R1} & 100 & 100 & 88.90 & 61.22 & 48.25 \\
\texttt{grok-4} & 100 & 100 & 87.13 & 60.29 & 49.80 \\
\hline
\end{tabular}
\end{table}

\clearpage
\subsection{Protein sequence optimization}
Each experiment began with an initial population of $200$ sequences, seeded with one experimentally defined wild-type sequence and $199$ additional variants generated by single-site random mutations of the wild type sequence. For GB1 and TrpB, this pool contains all measured 4-site variants, so initialization is not limited to single mutants. The mutation and crossover for LLMs were implemented by prompting the LLMs with two sampled parent sequences based on their fitness values and querying it to propose a new sequence either through mutation of one sequence or crossover of both sequences. If a proposed sequence was not present in the benchmark’s provided pool of scored variants (e.g., for GB1 and TrpB), it was treated as invalid and discarded, and the model was prompted again until a valid sequence with a defined fitness value was obtained. When we turned fitness values into probabilities to sample parents, we first normalized the fitness values to ensure non-negativity and then divided by the sum to obtain probabilities. We also applied a small shift to ensure the pool was not dominated by one candidate. After generating $100$ new and valid offspring in each generation, the new pool including the offspring and the parents were re-ranked by the fitness values and the top-$200$ candidates were kept in the population as the pool for next iteration. This process was repeated for $8$ iterations. We reported the fitness values of the top 1 candidates, normalized with their $(\min,\max)$ score range specific to each dataset. For GB1\cite{wu2016adaptation} and TrpB\cite{johnston2024combinatorially}, the range was computed directly from the experimentally measured fitness values in their respective benchmark datasets. For the ML-based oracles (GFP\cite{sarkisyan2016local} and AAV\cite{bryant2021deep}), the $(\min,\max)$ values were taken from the corresponding oracle’s training data to maintain consistency with its predictive scale\cite{kirjner2023improving}. For the synthetic Potts-model landscape (Syn-3bfo), the range was fixed to the most frequent region $(\min,\max)=(-3,\,3)$. We validated and compared the performance of LLMs against a simple evolutionary algorithm with identical initial populations and hyperparameters. The mutation operator in the baseline was a single-site mutation with a fixed probability of $0.3$. Once triggered, a mutation site was chosen at random and flipped to another random amino acid. The crossover operator in the baseline was implemented by selecting two parents based on their fitness values and a crossover site at random, then swapping the prefix and suffix of the two sequences. 

\begin{figure*}[!htbp]
    \centering
    \begin{tabular}{c}
        (a) \\
        \includegraphics[width=0.68\textwidth]{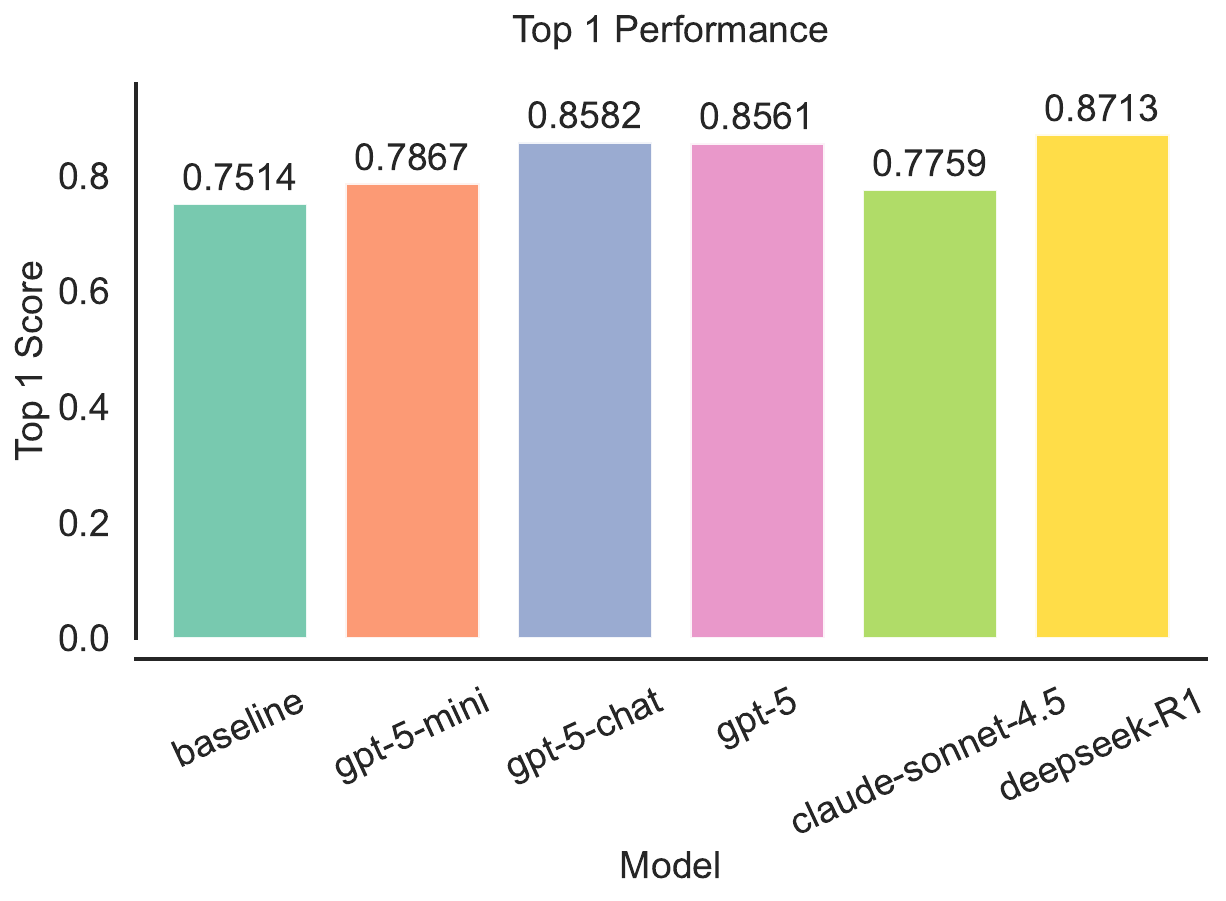} \\
        
        (b) \\
        \includegraphics[width=0.68\textwidth]{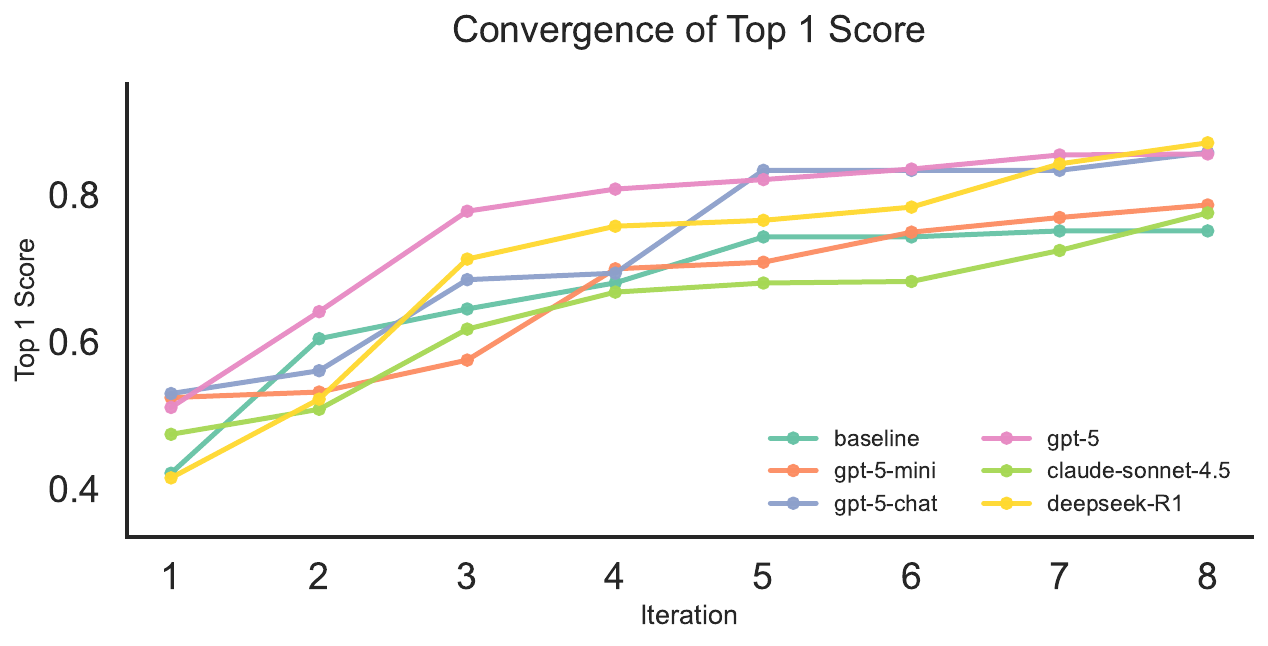} \\
        (c) \\
        \includegraphics[width=0.68\textwidth]{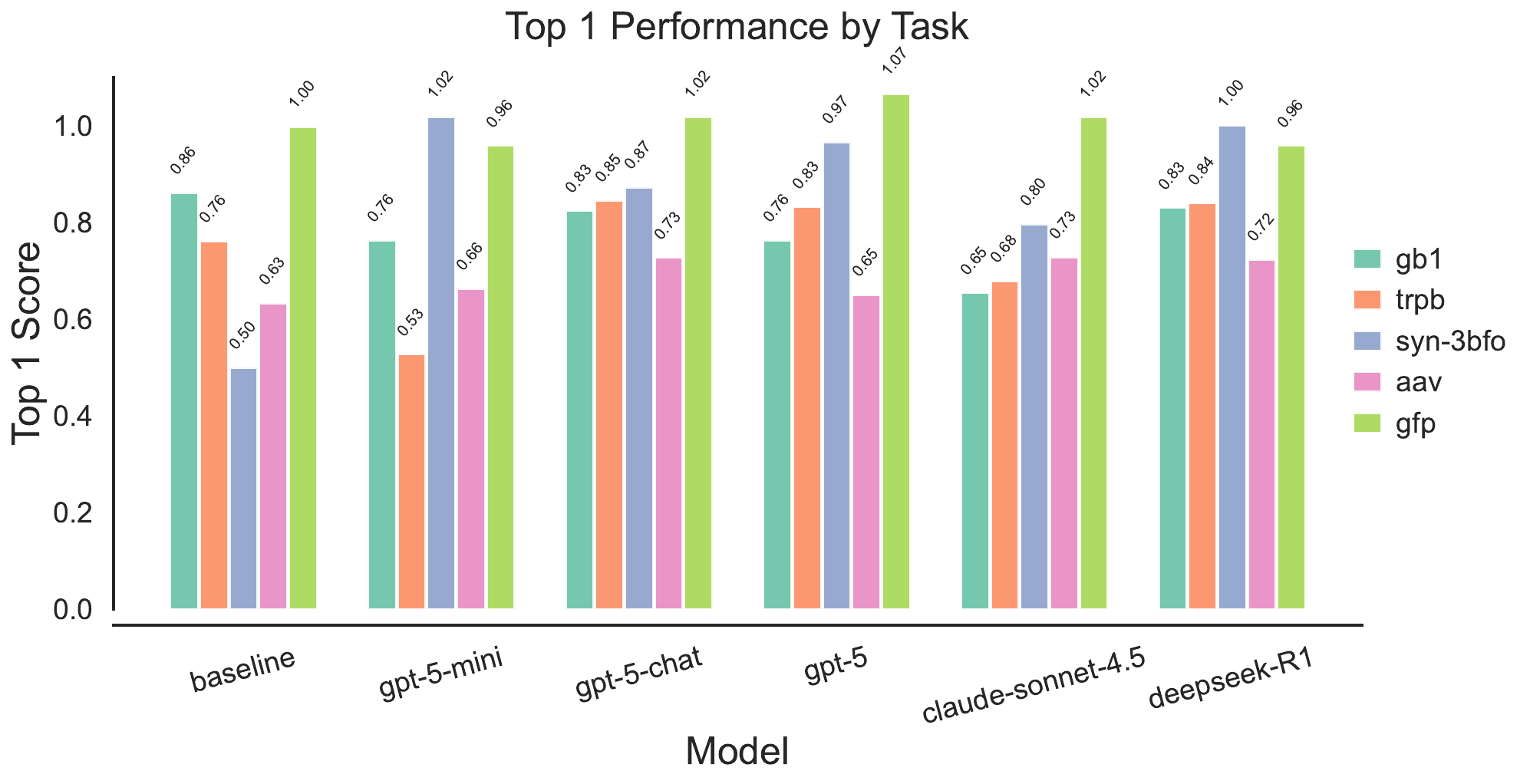} \\
    \end{tabular}
    \caption{(a) Bar chart of Top-1 performance across models. Bars report the Top-1 score for each model, where higher is better. \texttt{deepSeek-R1} achieves the highest Top-1 score at 0.8713, followed by \texttt{gpt-5-chat} at 0.8582 and \texttt{gpt-5} at 0.8561, while the baseline attains 0.7514. (b) Convergence of Top-1 score over evolutionary iterations for each model. Curves report the mean Top-1 score at each iteration averaged across the five protein optimization tasks, where higher indicates better solutions. (c) Top-1 performance by task and model. For each model cluster on the x-axis, five bars report the final Top-1 score achieved on GB1, TrpB, Syn-3bfo, AAV, and GFP.}
    \label{fig:POresult}
\end{figure*}

\Cref{fig:POresult} summarizes Top-1 performance for LLM-guided search, where each bar reflects the best score achieved per model and then averaged across the five tasks. \texttt{deepseek-R1} attains the highest average Top-1 score of $0.8713$, improving about $16.0\%$ over the baseline. \texttt{gpt-5-chat} and \texttt{gpt-5} follow closely at $0.8582$ and $0.8561$, indicating leading positions in the protein sequence optimization problem.

\quad Beyond the aggregated ranking, the gains are concentrated on the more challenging landscape: on Syn-3bfo, all LLM-guided variants substantially exceed the baseline, with improvements ranging from $59.5\%$ to $103.8\%$, suggesting that LLM-driven mutation and crossover better navigate fitness landscapes with strong interaction effects between mutation sites, leading to rugged regions where simple local operators struggle. In contrast, results on GB1 and GFP show limited and sometimes mixed gains. For GB1, the effective search space is small and the baseline already performs strongly, leaving little headroom and making the outcome sensitive to whether proposals preserve the few critical sites. For GFP, performance is near saturation for several methods, so differences tend to be smaller and can vary by model. Finally, the mid-tier models show reduced robustness across tasks, most notably \texttt{gpt-5-mini}, which performs competitively on some benchmarks yet drops sharply on TrpB, whereas \texttt{gpt-5-chat} exhibits the most consistent performance across tasks. Overall, these results suggest that LLM-based search is most valuable for harder search regimes, while robustness across diverse fitness landscapes remains an important consideration.

\quad The convergence curves in \Cref{fig:POresult} show a common pattern across methods: rapid gains in the first few iterations followed by slower, diminishing improvements as the search concentrates around strong candidates. \texttt{gpt-5} makes the largest early jump within the first three iterations, suggesting it proposes high-quality mutations or crossovers with fewer oracle evaluations. \texttt{deepSeek-R1} improves more gradually but continues to climb in later iterations, which points to stronger stability during refinement rather than relying on a single early leap. \texttt{gpt-5-chat} displays a clear improvement around iterations 4 to 5, consistent with moving from initial exploration into a more effective search regime, and it ultimately approaches the top-performing methods. By contrast, \texttt{gpt-5-mini} and \texttt{claude-sonnet-4.5} increase more slowly and level off earlier at lower scores, indicating less effective candidate proposals under the same evaluation budget. Overall, the curves suggest that differences arise from both final performance and optimization dynamics, including how quickly a model finds good regions and how reliably it improves thereafter\cite{wang2025large}.

\clearpage
\subsection{Gene editing}
We assess model performance using data from past genetic perturbation experiments. 
We simulate the perturbation of a gene g by retrieving the relevant observation of the perturbation-induced phenotype f(g) from this dataset. 
In every experimental round we perturb 128 genes at a single please, representing a reasonably sized small-scale biological screen.
Five rounds of experiment are performed with all historical observation fed into the prompts for the next round.
We use Interferon-$\gamma$ (IFNG) dataset, which measures the changes in the production of a key cytokine involved in immune signaling in primary human T-cells.

\qquad Across all models, \texttt{gpt-5} and \texttt{claude-sonnet-4.5} perform similarly well, reaching the same number of total hits with the only difference that \texttt{gpt-5} sometimes generates invalid perturbations.
Meanwhile, \texttt{deepseek-R1} generates fewer hits compared to the two best models.
\texttt{gpt-5-chat}, despite having much fewer total hits, reaches the best final efficiency as many of genes generated there are invalid and thus would not be tested explicitly in labs.

\begin{table}[ht]
\centering
\caption{Comparison of 4 LLMs on IFNG gene discovery task across 5 rounds (128 genes per round).}
\label{Supp:gene}
\small
\begin{tabular}{lccccc}
\toprule
Model & Total hits & Mean hit rate (HR) & Final efficiency & Best round HR & Unique genes tested \\
\midrule
\texttt{claude-sonnet-4.5} & \textbf{83} & \textbf{14.65\%} & \textbf{14.09\%} & \textbf{30.33\%} &  \textbf{640} \\
\texttt{gpt-5}& \textbf{83} & 13.84\% & 13.88\% & 26.02\%  & 598 \\
\texttt{deepseek-R1} & 74 & 12.67\% & 12.65\% & 21.67\%  & 585 \\
\texttt{gpt-5-chat} & 50 & 8.94\% & 14.71\% & 15.83\%  &  340 \\
\bottomrule
\end{tabular}
\end{table}

\clearpage
\subsection{Symbolic regression}
\label{sec:sr}
The symbolic regression task aims to discover closed-form mathematical expressions that govern the underlying dynamics of observed systems directly from data. Given a dataset of input--output pairs $\{(x_i, y_i)\}_{i=1}^{N}$, where $x_i \in \mathbb{R}^d$ denotes system states (or derived features) and $y_i \in \mathbb{R}$ denotes target values, the objective is to recover an interpretable symbolic expression $\hat{f}(x)$ that accurately approximates the true governing equation and generalizes beyond the training regime.
Unlike purely numerical regression, symbolic regression or equation discovery requires structured exploration over a discrete program space of possible mathematical relations, balancing expressivity, numerical accuracy, and simplicity. This setting naturally lends itself to iterative hypothesis generation and refinement in symbolic spaces, making it well suited for evaluating LLM-guided scientific discovery techniques.
At initialization, the LLM-based framework (employed from~\cite{shojaee2025llmsr}) constructs a prompt containing a concise description of the target scientific equation discovery task, variable definitions, and a small number of simple in-context examples (e.g., linear or low-degree polynomial relations). These examples seed a multi-island experience buffer that serves as the initial population for evolutionary refinement. At each iteration, the LLM samples candidate equation hypotheses as Python program skeletons of the form
\begin{equation}
\texttt{def f(x\_1, ..., x\_d, params): return y},
\end{equation}
where the symbolic structure and logic of the function is proposed by the LLM and the numeric parameters are left unspecified as placeholder parameters. 
For each proposed skeleton, these continuous placeholder parameters are then optimized against the observed data using off-the-shelf numeric optimizers (e.g., BFGS via \texttt{scipy}), yielding a fitted candidate equation. The optimized hypothesis is then evaluated by an oracle that executes the program and computes its fitness as the negative mean squared error (MSE) on the training data. Invalid programs (e.g., numerical instability, execution errors, or degenerate outputs) are discarded. High-scoring equation hypotheses are retained in the external experience buffer, which is organized as a multi-island population to maintain diversity and reduce premature convergence.
The evolutionary loop proceeds by repeatedly sampling from this experience buffer to condition subsequent LLM prompts, enabling the model to refine promising structures while continuing to explore novel functional forms. This process is repeated for a fixed number of iterations (up to $1000$ in our experiments), after which the highest-scoring equation is selected as the discovered scientific law.

We evaluate symbolic regression on a collection of physics nonlinear dynamical systems from recent benchmark~\cite{shojaee2025llmsrbench}. For each system, trajectories are provided numerically with split into \emph{in-distribution (ID)} and \emph{out-of-distribution (OOD)} regimes, where OOD data corresponds to extrapolation beyond the training range for held-out test purpose of discovered equations.
All datasets are normalized following standard practice, 
using the existing train--test splits across all methods to ensure fair comparison.
We compare multiple large language model (LLM) backbones as hypothesis generators within the same symbolic discovery framework, including \texttt{claude-sonnet-4.5}, \texttt{gpt-5}, \texttt{deepseek-R1}, and \texttt{gpt-5-chat-latest}. At each iteration, the LLM proposes candidate symbolic programs expressed in a restricted grammar of mathematical operators. These candidates are evaluated numerically against the dataset, ranked by error, and the best-performing programs are retained to guide subsequent proposals.

All LLM-based methods share the same prompt structure, grammar constraints, evaluation pipeline, and iteration budget; only the underlying language model backbone differs. Each experiment is repeated with multiple random seeds, and results are aggregated across datasets and runs.
As a non-LLM baseline, we include PySR, a widely used state-of-the-art symbolic regression method based on evolutionary search. PySR is run with recommended default hyperparameters and comparable computational budgets over all datasets.
We report both accuracy-based and error-based metrics to capture complementary aspects of symbolic discovery. Specifically, we report accuracy at a relative error threshold $\tau = 0.1$ as a more strict metric of fitness with respect to data:
\begin{equation}
\mathrm{Acc}_{\tau} = \mathbb{I}\left(
\max_{1 \le i \le N_{\text{test}}}
\frac{|\hat{y}_i - y_i|}{|y_i|}
\le \tau
\right),
\end{equation}
where $\hat{y}_i$ denotes model predictions and $\mathbb{I}(\cdot)$ is the indicator function. We additionally report the normalized mean squared error (NMSE): $\mathrm{NMSE} =
\frac{\sum_{i=1}^{N_{\text{test}}} (\hat{y}_i - y_i)^2}
{\sum_{i=1}^{N_{\text{test}}} (y_i - \bar{y})^2},$ where $\bar{y}$ is the mean of the ground-truth outputs. Metrics are computed separately for ID and OOD splits of each dataset.

Table~\ref{tab:sr} summarizes the quantitative performance of all methods on symbolic regression. Among LLM-based approaches, \texttt{deepseek-R1} achieves the highest in-distribution accuracy and the lowest NMSE, while \texttt{gpt-5} attains comparable performance with stronger out-of-distribution generalization. \texttt{claude-sonnet-4.5} performs competitively in-distribution but exhibits slower convergence and higher residual error, particularly in OOD settings. In contrast, \texttt{gpt-5-chat-latest} shows substantially degraded performance.
Compared to PySR, all LLM methods achieve markedly higher accuracy and mostly lower error. This suggests that LLM-guided discovery can surpass leading evolutionary symbolic regression by leveraging global structural priors and iterative hypothesis refinement rather than relying solely on pure local search operators.
Figure~\ref{fig:symbolic_discovery_curve} further illustrates these trends through discovery curves, which track the best normalized data-driven error achieved as a function of iteration, averaged across datasets. 

Beyond quantitative metrics, we examine the final symbolic programs discovered by each method. Qualitative inspection reveals that \texttt{deepseek-R1} and \texttt{gpt-5} show robustness in finding interpretable expressions faster in the process of discovery that closely match the true governing equations, often with minimal extraneous terms. \texttt{claude-sonnet-4.5} also frequently identifies partially correct structures but mostly with redundant components that are more sensitive to problems and initial populations.
Example of representative discovered programs for each method are provided below (example problem PO37), illustrating qualitative differences in symbolic structure, compactness, and physical interpretability.

\begin{figure*}[!htbp]
\centering
\includegraphics[width=0.6\textwidth]{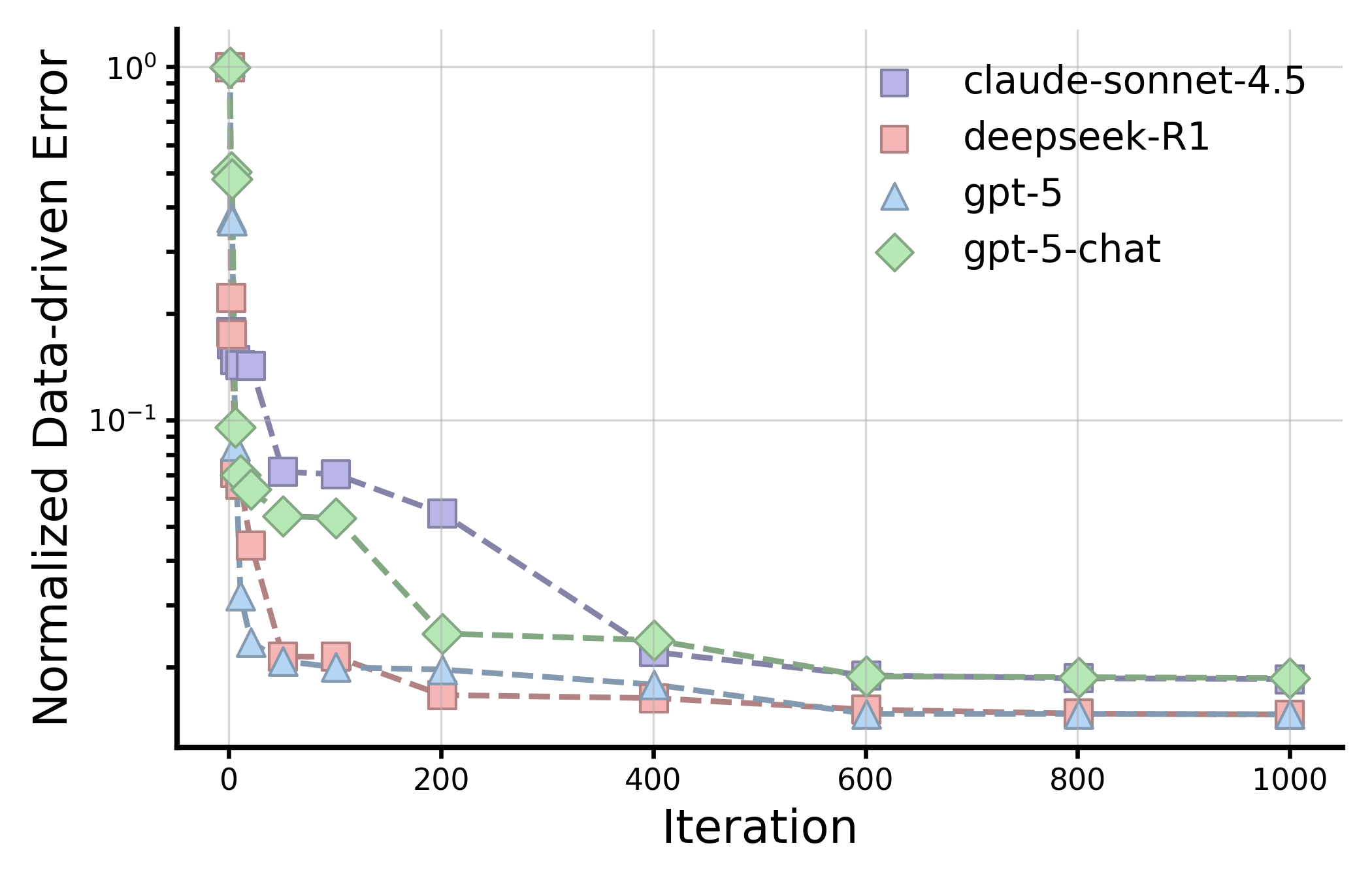}
\caption{
\looseness=-1 Discovery curves for symbolic regression across LLM backbones.
Normalized data-driven error (lower is better) of the best-discovered equation as a function of LLM-guided iterations, averaged across all the benchmark datasets. 
}
\label{Supp:tmc_pf}
\end{figure*}

\begin{table}[h]
\centering
\caption{Quantitative comparison results on symbolic regression.
In-distribution (ID) and out-of-distribution (OOD) performance for symbolic regression, reporting accuracy to threshold 0.1 ($Acc_{0.1}$, higher is better) and normalized mean squared error (NMSE, lower is better). Results compare LLM-based methods and PySR, a non-LLM state-of-the-art baseline, highlighting differences in accuracy, error, and generalization to OOD data. Best values are highlighted in \textbf{bold}.}
\resizebox{0.8\textwidth}{!}{
    \begin{tabular}{lrrrrr}
    \midrule
    \textbf{Model} & \textbf{$Acc_{0.1}$ ID (\%)$\uparrow$} & \textbf{$Acc_{0.1}$ OOD (\%)$\uparrow$} & \textbf{NMSE ID$\downarrow$} & \textbf{NMSE OOD$\downarrow$} \\
    \midrule
    PySR & 13.79 & 3.44 & 1.0786 & 475.7\\
    \midrule
    \texttt{claude-sonnet-4.5}  & 55.17 & 48.28 & 0.03896 & 274.5 \\
    \texttt{gpt-5}  & 55.17 & \textbf{51.72} & 0.00611 & \textbf{266.3} \\
    \texttt{deepseek-R1} & \textbf{58.62} & \textbf{51.72} & \textbf{0.00426} & 1009.1 \\
    \texttt{gpt-5-chat} & 48.28 & 34.48 & 0.03823 & 91118.4 \\
    \midrule
    \end{tabular}
}
\label{tab:sr}
\end{table}

\begin{figure}[t]
  \centering
  \includegraphics[width=0.5\linewidth]{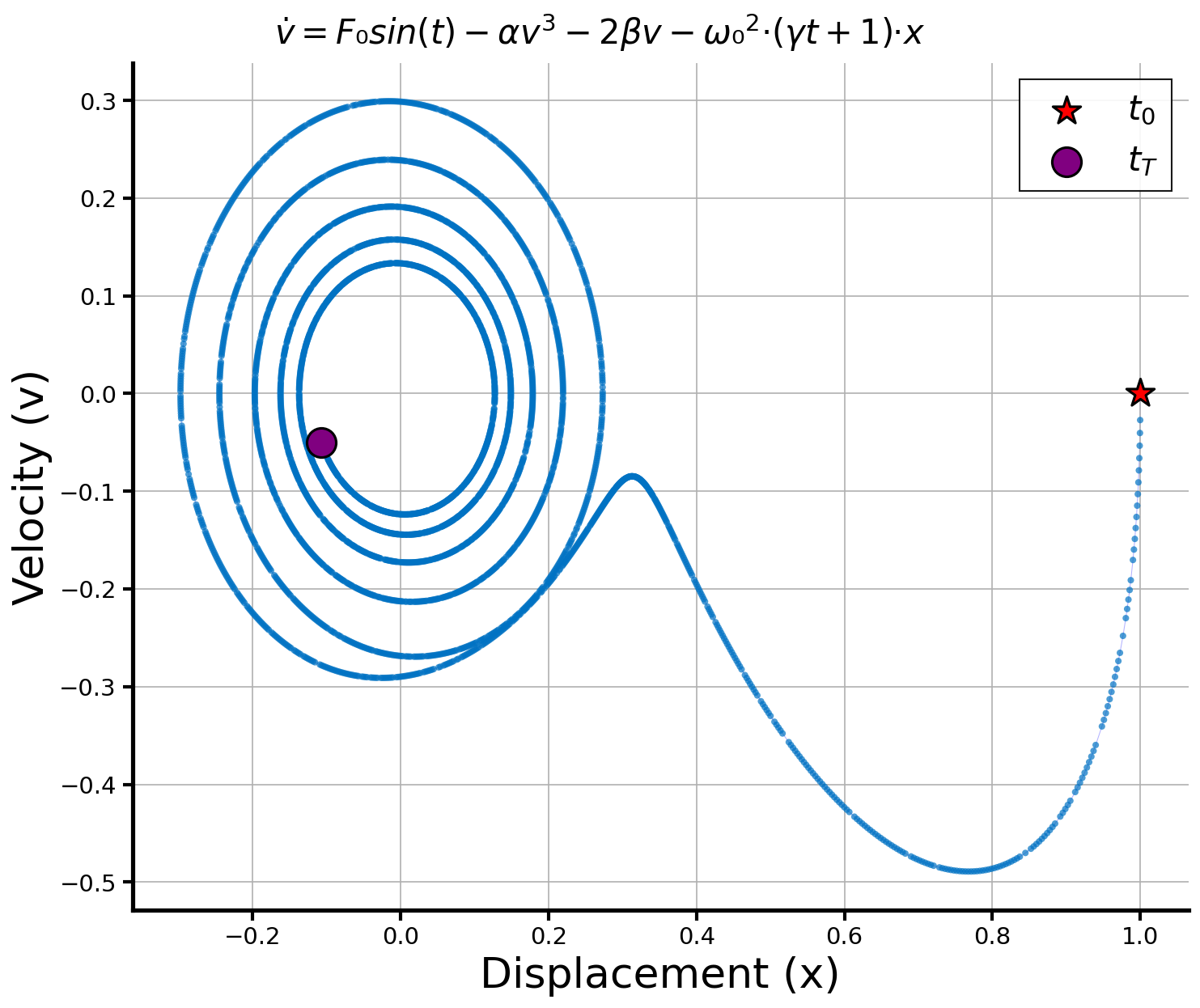}
  \caption{Phase-space trajectory ($x$ vs.\ $v$) visualization of example non-linear oscillator problem PO37~\cite{shojaee2025llmsrbench} The plot annotates the ground-truth equation skeleton and indicates the initial
  state $t_{0}$ and final state $t_{T}$.}
  \label{fig:po37_phase}
\end{figure}

\paragraph{Qualitative example for final discovered equation programs.}
We present a qualitative example for benchmark problem PO37~\cite{shojaee2025llmsrbench}. Figure~\ref{fig:po37_phase}
shows the phase-space trajectory of the ground truth non-linear oscillator generated dynamics, alongside its ground-truth equation skeleton.

\newpage

{\small
\begin{lstlisting}
########## PySR #########
params[0]**exp(t)*x*t - params[1] - x/params[2] - x*params[3]
\end{lstlisting}
}

{\small
\begin{lstlisting}
########## GPT-5 #########
def equation(x, t, v, params):
    # Unpack parameters
    k1, k2, k3, c1, c2, c3, F1, w1, phi1, bias = params
    # Stiffness terms (linear + time-varying)
    stiffness = -(k1 + k2 * t) * x
    # Nonlinear stiffness (Duffing)
    cubic_stiffness = -k3 * x**3
    # Damping (linear + position-dependent + cubic velocity)
    damping = -(c1 + c2 * x**2) * v - c3 * v**3
    # External periodic driving
    drive = F1 * np.sin(w1 * t + phi1)
    # Combine all terms
    acceleration = stiffness + cubic_stiffness + damping + drive + bias
    
    return acceleration
\end{lstlisting}
}

{\small
\begin{lstlisting}
########## DeepSeek-R1 #########
def equation(x, t, v, params):
    # Unpack parameters
    w_sq, beta, delta1, delta2, alpha, F, omega, gamma, epsilon, lambda_ = params
    # Restoring forces
    linear_restoring = -w_sq * x
    cubic_restoring = -beta * x**3
    quintic_restoring = -epsilon * x * np.abs(x)**4
    # Damping forces
    linear_damping = -delta1 * v
    quadratic_damping = -delta2 * v * np.abs(v)
    position_dependent_damping = -alpha * x**2 * v
    # Position-velocity coupling
    position_velocity_coupling = -gamma * x * np.abs(v)
    # External forcing with modulation
    sinusoidal_forcing = F * np.sin(omega * t)
    envelope = np.exp(-lambda_ * np.abs(t))
    modulated_forcing = sinusoidal_forcing * envelope
    # Mixed cubic term
    cubic_mixed = -0.1 * (v**3) * np.sign(x) * np.exp(-0.01 * x**2)
    # Combine all terms
    acceleration = (
        linear_restoring +
        cubic_restoring +
        quintic_restoring +
        linear_damping +
        quadratic_damping +
        position_dependent_damping +
        position_velocity_coupling +
        modulated_forcing +
        cubic_mixed
    )
    
    return acceleration
\end{lstlisting}
}

{\small
\begin{lstlisting}
########## Claude-Sonnet-4.5 #########
def equation(x, t, v, params):
    # Restoring forces (odd powers preserve symmetry)
    linear_restoring = params[0] * x
    cubic_restoring = params[1] * x**3
    quintic_restoring = params[2] * x**5
    # Linear damping
    linear_damping = params[3] * v
    # Non-linear damping mechanisms
    van_der_pol_damping = params[4] * x**2 * v
    cubic_velocity_damping = params[5] * v**3
    # Cross-coupling term
    velocity_position_coupling = params[6] * x * v**2
    # Asymmetric potential (breaks x -> -x symmetry)
    asymmetric_force = params[7] * x**2
    # Parametric forcing
    time_dependent_forcing = params[8] * np.cos(params[9] * t) * x
    # Combine all terms
    acceleration = (
        linear_restoring +
        cubic_restoring +
        quintic_restoring +
        linear_damping +
        van_der_pol_damping +
        cubic_velocity_damping +
        velocity_position_coupling +
        asymmetric_force +
        time_dependent_forcing
    )
    
    return acceleration
\end{lstlisting}
}

{\small
\begin{lstlisting}
########## GPT-5-Chat #########
def equation(x, t, v, params):
    # Extract parameters
    inv_mass = params[0]  # 1/m (inverse mass)
    k = params[1]         # linear stiffness
    c = params[2]         # linear damping
    beta = params[3]      # cubic stiffness
    delta = params[4]     # cubic damping
    F1 = params[5]        # forcing amplitude 1
    omega1 = params[6]    # forcing frequency 1
    phi1 = params[7]      # forcing phase 1
    F2 = params[8]        # forcing amplitude 2
    omega2 = params[9]    # forcing frequency 2
    # Calculate force components
    restoring = -k * x - beta * (x ** 3)
    damping = -c * v - delta * (v ** 3)
    forcing = F1 * np.cos(omega1 * t + phi1) + F2 * np.sin(omega2 * t)
    # Calculate acceleration
    a = inv_mass * (restoring + damping + forcing)
    
    return a
\end{lstlisting}
}

\clearpage
\subsection{Solving Ising model}

In this project, each experiment targets the discovery of a low-energy configuration of a classical Ising model with unknown optimal bitstring structure, given a fixed but hidden set of pairwise couplings. The system is defined by a random all-to-all connected $N$-spin Ising Hamiltonian
\begin{equation}
\label{eq:ham}
H(\mathbf{s}) = \sum_{i<j} J_{ij} s_i s_j ,
\end{equation}
where $s_i \in \{+1,-1\}$ are spin variables and $J_{ij}$ are symmetric couplings drawn randomly at the beginning of each experiment and held fixed throughout the run. The objective is to recover a bitstring $\mathbf{b} \in \{0,1\}^N$, mapped to spins via $s_i = +1$ if $b_i=1$ and $s_i=-1$ otherwise, that minimizes the Hamiltonian energy.

Unlike traditional combinatorial optimization methods that rely on local update rules or handcrafted heuristics, we formulate the problem as a LLM-guided evolutionary search. Candidate solutions are generated by a large language model (LLM), evaluated by a physics-based oracle, and iteratively refined through an experience-driven prompt-conditioning mechanism.

At initialization, the framework constructs a prompt describing the Ising optimization task, including the Hamiltonian definition, the spin--bit mapping, and the system size $N$. A small set of simple candidate bitstrings (e.g., all zeros, all ones, and alternating patterns) is provided as in-context examples to seed the search. These initial candidates define the starting population for the evolutionary process. At each iteration, the LLM is queried to propose exactly one new candidate bitstring of length $N$. The output is constrained to a raw binary string without explanation, ensuring that each proposal corresponds to a discrete global configuration of the Ising system. In this formulation, each LLM output represents a symbolic hypothesis over the full configuration space rather than a local modification rule.

Each candidate bitstring is evaluated by a deterministic oracle that computes its energy under the fixed Hamiltonian in Eq.~\ref{eq:ham}.
For benchmarking and convergence monitoring, the exact ground-state energy is precomputed via brute-force enumeration for the system sizes considered. The primary fitness signal is defined as the normalized energy gap above the ground state. 
Rather than employing explicit genetic operators such as mutation or crossover, evolutionary refinement is implemented implicitly through prompt conditioning and experience accumulation. The framework maintains an external experience buffer that stores previously generated bitstrings together with their associated fitness scores. At each iteration, information derived from this buffer, including previously explored configurations, the current best-so-far energy, and qualitative feedback on search progress, is incorporated into the LLM prompt. When the energy gap remains large, the prompt encourages exploration of qualitatively different bit patterns; as the gap narrows, the prompt biases the LLM toward local refinements around promising configurations. This mechanism induces selection pressure while maintaining diversity in the search process. 

The experiment is repeated for 40 iterations per round and 4 rounds in total are conducted. Throughout the optimization, the best-so-far bitstring and its corresponding energy are tracked at each iteration, yielding a complete optimization trajectory.
The proposed framework is evaluated on Ising models with system size $N=12$ and a randomly generated coupling matrix. Performance is assessed using the final energy gap relative to the exact ground state and the convergence behavior across iterations. As a classical baseline, simulated annealing is implemented using single-spin-flip Metropolis updates, where at each step one randomly selected spin is flipped and the move is accepted according to the Boltzmann criterion. The temperature follows an exponential cooling schedule \(T_k = T_0 \alpha^k\) with \(T_0 = 1.0\) and \(\alpha = 0.9\), and the total number of annealing steps $k$ is matched to the number of LLM iterations to ensure a controlled comparison. By comparing the LLM-generated trajectories to the exact solution and the baseline simulated annealing, we quantify both optimization effectiveness and the ability of the method to navigate rugged, high-dimensional energy landscapes.

Overall, this framework demonstrates how LLMs can be repurposed as global, structure-aware proposal mechanisms for discrete physics optimization problems. By combining LLM-based hypothesis generation with exact physical evaluation and an implicit evolutionary memory, the method enables systematic exploration of exponentially large configuration spaces. We observe substantial variability in performance across different LLMs, and at the present stage no consistent advantage is found over conventional baselines such as simulated annealing. Future work aimed at understanding and improving the synergy between language-guided reasoning and physics-based evaluation may establish a general and effective paradigm for language-assisted scientific optimization.

\begin{figure}[h]
    \centering
    \includegraphics[width=0.75\linewidth]{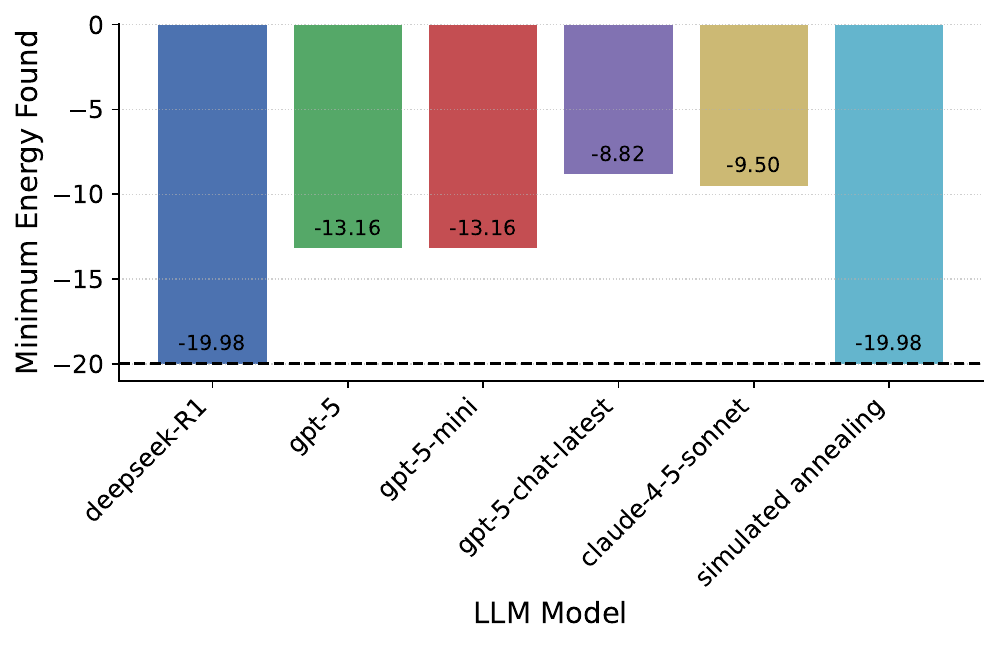}
    \caption{Bar chart of Ising model energy minimization across models. Bars report the score for each model, where lower is better. DeepSeek-R1 achieves the best score at -19.98, followed by GPT-5 at -13.16 and GPT-5-mini at -13.16, while the baseline (simulated annealing) attains -19.98. The ground truth energy is -19.98.}
    \label{Supp:ising}
\end{figure}

\clearpage

\begin{table}[!htbp]
\centering
\caption{Correspondence between projects and their top-3 scenarios in this work.}
\label{Supp:correspondence}
\begin{tabular}{lc}
\toprule
Project & Scenario \\
\midrule
Protein sequence optimization & protein localization, CRISPR delivery, property matching \\
Gene editing & gwas causal gene, gene editing, molecular pair \\
Retrosynthesis pathway design & retrosynthesis, reaction mechanism, forward reaction prediction\\
Molecule optimization & descriptor prediction, fragment completion, molecular property\\
TMC optimization & TMC properties, redox potential, molecular property\\
Crystal structure discovery & general materials, PXRD lattice prediction, composite materials\\
Symbolic regression & computation, core knowledge, statistics\\
Solving Ising model  & computation, condensed matter, quantum information\\
\bottomrule
\end{tabular}
\end{table}

\printbibliography[
  title={Supplementary References},
  segment=2,
  notcategory=maincited
]
\end{refsegment}

\end{refsection}

\end{document}